\documentclass[twoside]{article}

\usepackage[accepted]{aistats2026}

\usepackage[dvipsnames]{xcolor}

\usepackage{amsmath,amsfonts,bm}

\def\eqref#1{equation~\ref{#1}}

\def\1{\bm{1}}

\def\vx{{\bm{x}}}

\def\vz{{\bm{z}}}

\DeclareMathAlphabet{\mathsfit}{\encodingdefault}{\sfdefault}{m}{sl}
\SetMathAlphabet{\mathsfit}{bold}{\encodingdefault}{\sfdefault}{bx}{n}

\newcommand\iidsim{\stackrel{\text{i.i.d.}}{\sim}}

\newcommand{\dis}{{$\Delta\text{IS}$\xspace}}
\newcommand{\dex}{{$\Delta\text{EX}$\xspace}}

\newcommand{\lz}[1]{\textbf{\textcolor{WildStrawberry}{{lz: #1}}}} 
\newcommand{\nb}[1]{\textbf{\textcolor{blue}{{nb: #1}}}} 
\newcommand{\av}[1]{\textbf{\textcolor{orange}{{av: #1}}}} 
\newcommand{\lds}[1]{\textbf{\textcolor{green}{{lds: #1}}}} 
\newcommand{\kolya}[1]{\textbf{\textcolor{violet}{{nm: #1}}}}
\newcommand{\victor}[1]{\textbf{\textcolor{blue}{{ve: #1}}}}

\renewcommand{\lz}[1]{\relax}
\renewcommand{\nb}[1]{\relax}
\renewcommand{\av}[1]{\relax}
\renewcommand{\lds}[1]{\relax}
\renewcommand{\kolya}[1]{\relax}
\renewcommand{\victor}[1]{\relax}

\makeatletter
\newcommand{\removelatexerror}{\let\@latex@error\@gobble}
\makeatother

\usepackage{graphicx} %
\usepackage{amsmath}
\usepackage{amssymb}
\usepackage{amsfonts}
\usepackage{xspace}
\usepackage{multirow}
\usepackage{booktabs}
\usepackage{url}
\usepackage[ruled,noend]{algorithm2e}
\usepackage[parfill]{parskip}
\usepackage{hyperref}
\usepackage[capitalize]{cleveref}
\usepackage[inline]{enumitem}

\hypersetup{
colorlinks=true,linkcolor=pink!50!brown,citecolor=teal,urlcolor=orange
}

\newcommand{\method}{\Delta\text{IS}}

\newtheorem{theorem}{Theorem}

\newtheorem{proposition}{Proposition}

\newtheorem{definition}{\textbf{Definition}}

\makeatletter
\crefname{section}{\S\@gobble}{\S\@gobble}
\Crefname{section}{\S\@gobble}{\S\@gobble}
\makeatother
\crefname{figure}{Fig.}{Figs.}
\Crefname{figure}{Fig.}{Figs.}
\crefname{equation}{Eq.}{Eqs.}
\Crefname{equation}{Eq.}{Eqs.}
\crefname{table}{Tab.}{Tabs.}
\Crefname{table}{Tab.}{Tabs.}
\crefname{appendix}{App.}{Apps.}
\Crefname{appendix}{App.}{Apps.}
\crefname{algorithm}{Alg.}{Algs.}
\Crefname{algorithm}{Alg.}{Algs.}

\makeatletter
\crefname{section}{\S\@gobble}{\S\@gobble}
\crefname{subsection}{\S\@gobble}{\S\@gobble}
\crefname{proposition}{Prop.}{Props.}
\crefname{figure}{Fig.}{Figs.}
\crefname{table}{Table}{Tables}

\newcommand{\zerodisplayskips}{%
  \setlength{\abovedisplayskip}{2mm}%
  \setlength{\belowdisplayskip}{2mm}%
  \setlength{\abovedisplayshortskip}{1mm}%
  \setlength{\belowdisplayshortskip}{1mm}}
\appto{\normalsize}{\zerodisplayskips}
\appto{\small}{\zerodisplayskips}
\appto{\footnotesize}{\zerodisplayskips}

\makeatother

\newcommand{\std}[1]{\text{\small $\pm #1$}}

\usepackage[round]{natbib}

\begin{document}
\runningauthor{Zellinger, Branchini, De Smet, Elvira, Malkin, Vergari}
\twocolumn[

\aistatstitle{How to Approximate Inference with Subtractive Mixture Models}

\aistatsauthor{Lena Zellinger\\University of Edinburgh 
\And Nicola Branchini\\University of Warwick 
\AND Lennert De Smet\\KU Leuven 
\And V\'{i}ctor Elvira\\University of Edinburgh
\And Nikolay Malkin\\University of Edinburgh  
\And Antonio Vergari\\University of Edinburgh
}
\aistatsaddress{}
]

\begin{abstract}
Classical mixture models (MMs) are widely used tractable proposals for approximate inference settings such as variational inference (VI) and importance sampling (IS).
Recently, mixture models with negative coefficients, called subtractive mixture models (SMMs), have been proposed as a potentially more expressive alternative.
However, how to effectively use SMMs for VI and IS is still an open question as they do not provide latent variable semantics and therefore cannot use sampling schemes for classical MMs. 
In this work, we study how to
circumvent this issue by designing several
expectation estimators for IS and learning schemes for VI with SMMs,
and we empirically evaluate them for 
distribution approximation.
Finally, we discuss the additional challenges in estimation stability and learning efficiency that they carry and propose ways to overcome them.
Code is available at \url{https://github.com/april-tools/delta-vi}.
\end{abstract}

\section{INTRODUCTION}
Mixture models (MMs) are a staple in probabilistic modeling since they can represent complex multi-modal distributions via  
a 
\textit{convex combination of simple probability density functions} (PDFs) \citep{mclachlan2019finite}. A classical MM is defined as
\begin{equation}
q_{\text{MM}}(\boldsymbol{x}) = 
\sum\limits_{k=1}^{K} \alpha_{k} q_{k}(\boldsymbol{x}),\quad \alpha_{k}  \geq 0,\quad \sum\limits_{k=1}^{K}\alpha_{k} = 1
    \label{eq:mm}
\end{equation}
for every input $\boldsymbol{x}\in \mathbb{R}^D$,
where $q_k$ are the mixture components and  $\alpha_{k}$ are the mixture weights.
The coefficients $\alpha_k$ can be interpreted as the parameters of a categorical prior in a discrete latent variable model \citep{bishop2006pattern,peharz2016latent}, which enables efficient ancestral sampling from $q_{\text{MM}}$. 
As a result, MMs have been extensively used in approximate inference settings, where the goal is to estimate intractable quantities w.r.t. a target density by sampling from a \textit{tractable surrogate}.
In particular, MMs are common surrogates in variational inference (VI) \citep{jaakkola1998improving,guo2016boosting,morningstar2021automatic,kviman2022multiple} and well-studied proposals for (multiple) importance sampling (IS) \citep{veach1995optimally,owen2000safe,elvira2019generalized,sbert2018multiple}.

\begin{figure}
\begin{center}
\resizebox{0.49\textwidth}{!}{
\begin{tabular}{ccc}
\Huge{\textbf{Target}} & \Huge{\textbf{GMM (K = 2)}} & \Huge{\textbf{SMM (K = 2)}}\\
\includegraphics[scale=0.5]{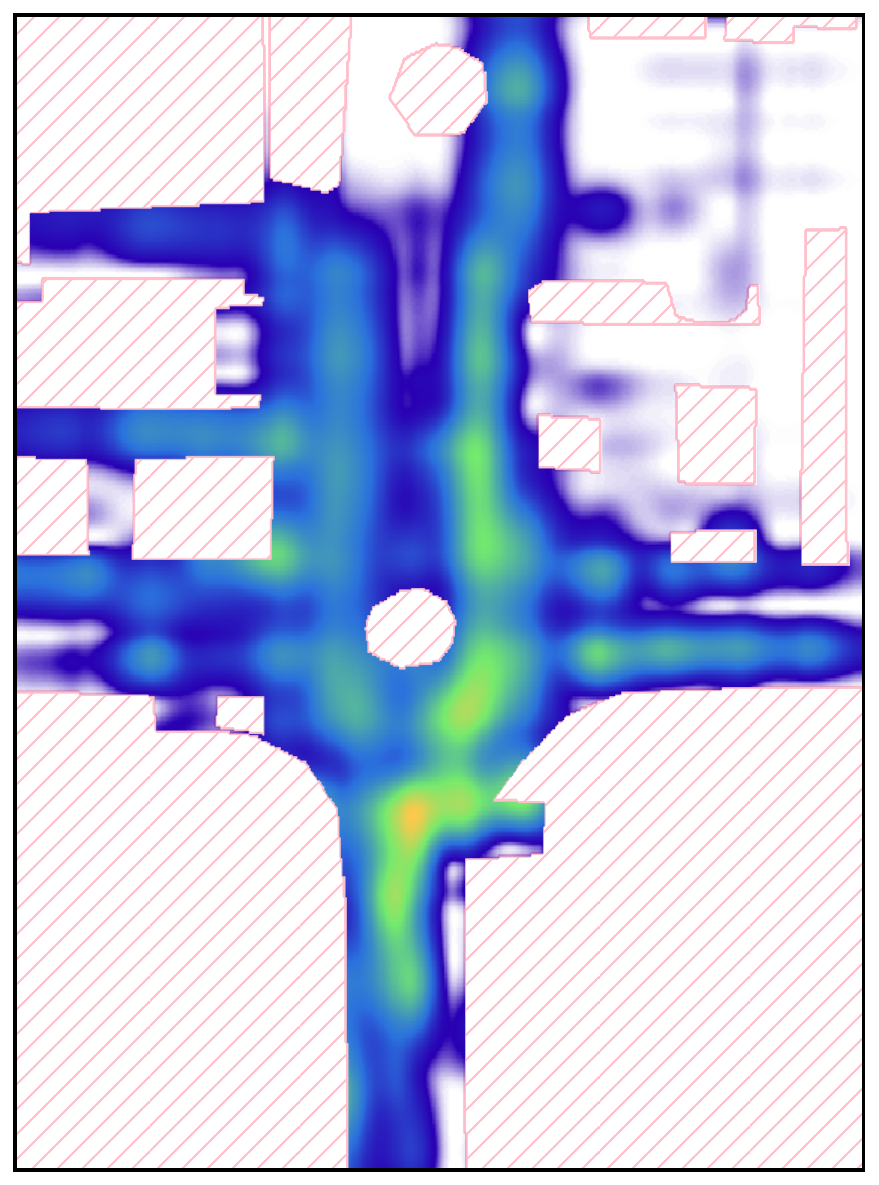}
&
\includegraphics[scale=0.5]{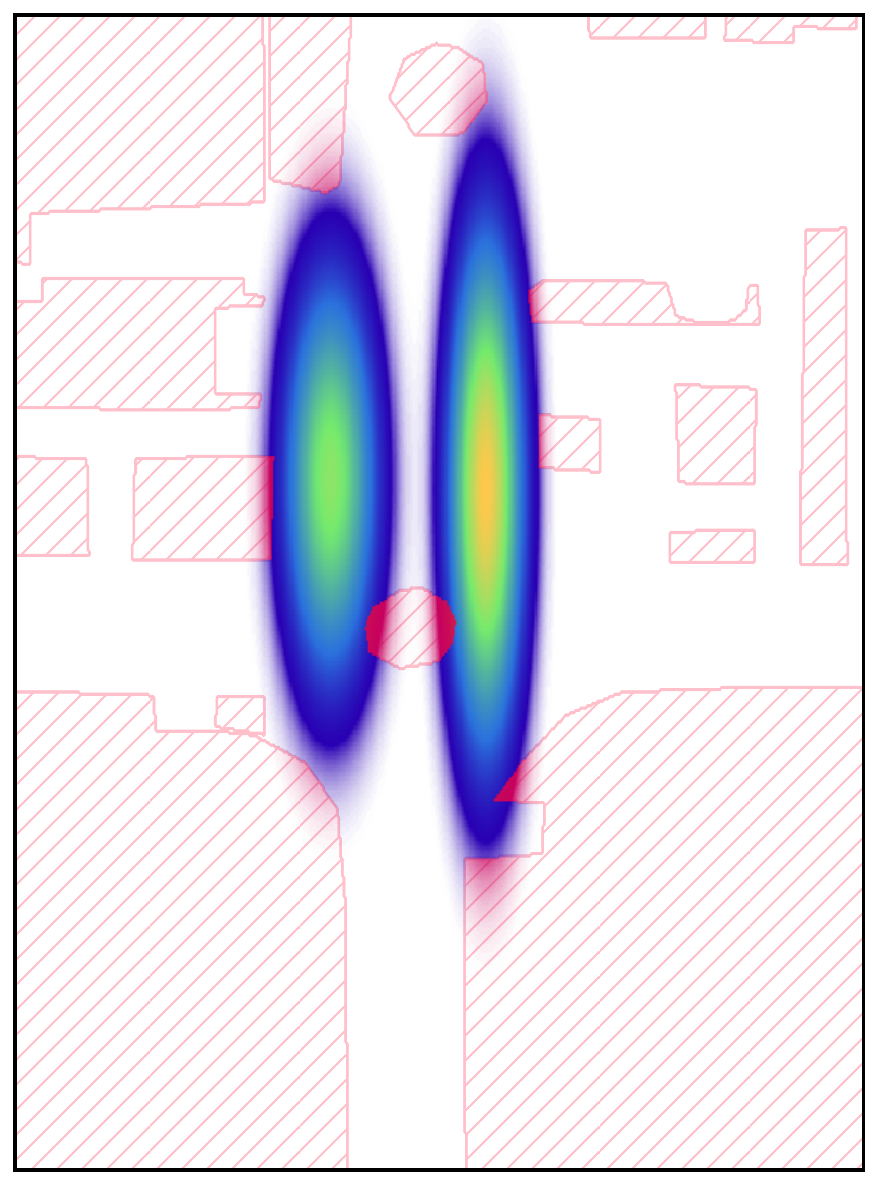}
&
\includegraphics[scale=0.5]{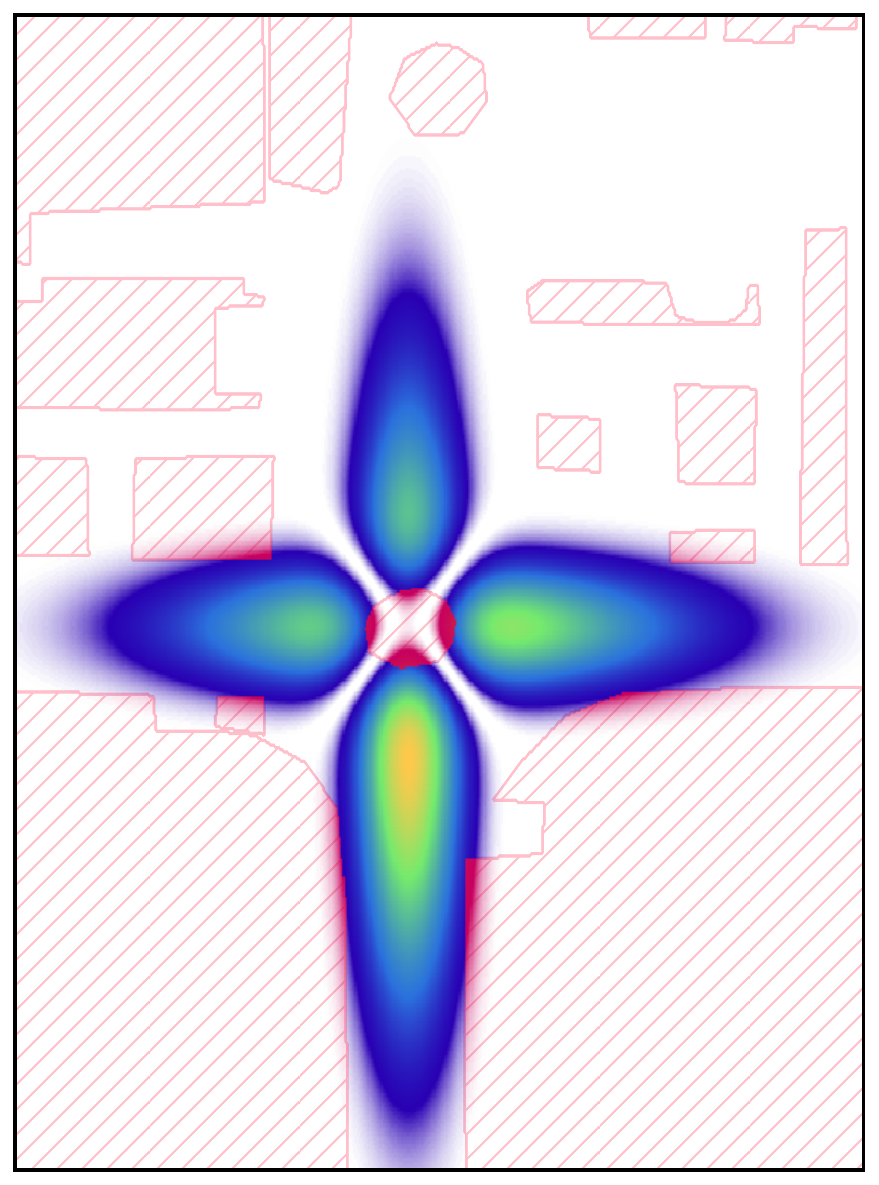}
\end{tabular}
}
\end{center}
\caption{\textbf{SMMs are effective variational families for targets with disconnected support} such as trajectories over the walkable area from the \emph{Stanford drone dataset} \citep{robicquet2016learning}. With just $K=2$ components, the SMM \emph{learns the absence of density} at the central roundabout while a GMM requires $K>2$.
We lay the foundation to use SMMs for VI.
}\label{fig:pal_fig_1}
\end{figure}

Recently, \textit{subtractive mixture models} (SMMs) \citep{marteauferey2020nonparametric,rudi2021-psd,loconte2024subtractive,cai2024eigenvi} have been introduced in ML
as a generalization of classical additive MMs that relaxes the convexity constraint on the mixture coefficients in \cref{eq:mm} by allowing them to take negative values.
The resulting \emph{subtraction of density} can be particularly beneficial when modeling targets that have deep valleys or disconnected support (see \cref{fig:pal_fig_1}). 
\emph{Squaring SMMs} \citep{loconte2024subtractive} has enabled unconstrained gradient-based learning via maximum likelihood, and squared SMMs have shown good results in data-driven settings \citep{loconte2023turn,loconte2024sos,wang2024relationship}. 
Using SMMs also for approximate inference -- without access to target samples -- can open up a wider class of tractable surrogates that can be more \emph{expressive efficient} \citep{choi2020probabilistic}: they allow to model complex distributions with potentially exponentially fewer parameters than classical MMs. 
However, this additional flexibility comes at the cost of \emph{losing the latent variable interpretation} of the mixture weights. 
Negative coefficients cannot be interpreted as probabilities, and hence SMMs do not allow for ancestral sampling, raising the question:

\begin{quote}
\emph{How can we make SMMs viable tractable surrogates for IS and VI despite the lack of a latent variable semantics?}
\end{quote}
A first step in that direction comes from \citet{cai2024eigenvi} who devise a specialized closed-form VI scheme for squared SMMs with orthogonal polynomials as components, optimizing the Fisher divergence \citep{hyvarinen2005estimation}.
In this work, we take a broader perspective and study \emph{general expectation estimation over SMMs}, which is central to (1) learning SMMs via (black-box) VI and (2) using the resulting models as proposals for IS to estimate quantities of interest. We detail our contributions below.

\paragraph{Contributions.} In \Cref{sec:approximate_inference} we first provide a \textbf{(C1)} comprehensive discussion of sampling schemes for expectation estimation over SMMs. We cover the standard techniques auto-regressive inverse transform sampling \citep[ARITS,][]{loconte2024subtractive} and rejection sampling \citep{bignami1971note}, but also discuss \textbf{(C2)} $\Delta\text{IS}$ -- a new estimator recently hinted at by \citet{robert2025simulating}, which relies on a decomposition of the SMM into two additive MMs, recovering ancestral sampling. We further develop a \emph{safe variant} of $\Delta\text{IS}$, inspired by the safe adaptive IS literature \citep{delyon2021safe,korba2022adaptive}, to stabilize it in practice. %
In \Cref{sec:VI_SMMs}, we then \textbf{(C3)} show how each of these techniques can be used for black-box VI.
Notably, we discuss how $\Delta\text{IS}$ opens up the use of the 
reparameterization trick \citep{morningstar2021automatic}. %
 Finally, in \Cref{sec:experiments},  we \textbf{(C4)} extensively evaluate our pipelines for VI and IS on targets of varying dimensionality, providing a first empirical comparison to additive MMs and highlighting open challenges and future opportunities for approximate inference with SMMs.

\section{SUBTRACTIVE MIXTURES}\label{sec:smms}
Subtractive mixture models (SMMs) generalize classical MMs (\cref{eq:mm}) by allowing for negative mixture coefficients, leading to a potential subtraction of density.
A SMM over $\vx \in \mathbb{R}^D$ is defined as
\begin{equation}\tag{SMM}
    \label{eq:smm}
    q_{\text{SMM}}(\boldsymbol{x}) = Z^{-1} \cdot \sum\limits_{k=1}^{K} \alpha_{k} q_{k}(\boldsymbol{x}), ~ \text{where}
    ~ \alpha_{k}  \in \mathbb{R},
\end{equation}
where each $q_k$ is a (possibly unnormalized) PDF
and $Z=\sum\nolimits_{k=1}^{K} \alpha_{k} \int q_{k}(\vx)d\vx$ is the normalizing constant of the SMM.
A key challenge when constructing and \emph{learning} SMMs is ensuring $q_{\text{SMM}}(\boldsymbol{x}) \geq 0$ for all inputs $\boldsymbol{x}$ in order to retain a valid PDF. 
While it is possible to derive closed-form constraints for simple parametric distributions, such as Gaussian, Gamma, and Weibull \citep{jiang1999weibull,zhang2005finite,rabusseau2014learning}, this is non-trivial in general.
To this end, 
\cite{loconte2024subtractive} learned \textit{squared SMMs}, ensuring non-negativity by squaring \Cref{eq:smm}:
\begin{align}\label{eq:squared_smm}\tag{$\text{SMM}^2$}
    q_{\text{SMM}^2}(\boldsymbol{x}) 
    &= Z^{-1} \cdot \left(\sum\nolimits_{k=1}^{K}\alpha_{k} q_{k}(\boldsymbol{x})\right)^2\\
    &= Z^{-1} \cdot \sum\nolimits_{k=1}^{K}\sum\nolimits_{k^\prime=1}^{K^\prime} \alpha_{k}\alpha_{k^\prime} q_{k}(\boldsymbol{x}) q_{k^\prime}(\boldsymbol{x}) \nonumber   ,
\end{align}
where $Z =  \sum_{k=1}^{K}\sum_{k^\prime=1}^{K^\prime} \alpha_{k}\alpha_{k^\prime} \int q_{k}(\boldsymbol{x}) q_{k^\prime}(\boldsymbol{x}) d \vx $. 
\cref{fig:smms-pcs}  
shows the computational graph of a squared SMM.
Note that \cref{eq:squared_smm} is still a SMM since negative coefficients $\alpha_{k}\alpha_{k^\prime} < 0$ are possible.%
While a squared SMM has ${K+1}\choose{2}$ components after squaring, the number of learnable parameters is still $\mathcal{O}(K)$.
To exactly compute $Z$, and retain tractability, we need an analytical form for $\int q_{k}(\vx)q_{k^\prime}(\vx)d\vx$, which is the case for exponential families and other functions, such as polynomials on bounded intervals \citep{loconte2024subtractive}.
If the components $q_k$ form an orthonormal basis, and $\sum_i\alpha_i^2=1$, the SMM will be normalized by design \citep{cai2024eigenvi,loconte2026square}. 
In terms of \emph{expressive efficiency} \citep{choi2020probabilistic} squared SMMs and additive MMs are \emph{incomparable}: A single squared SMM can require exponentially fewer parameters than an additive MM to represent certain distributions \citep{loconte2024subtractive}, but the opposite can also be true \citep{loconte2024sos,wang2024relationship}.
However, a \emph{sum of squared (SOS) SMMs} can be more expressive efficient than both classical MMs and a squared SMM \citep{loconte2024sos}.
A simple SOS model can be built by using \emph{complex mixture weights} $\alpha_k:=a_k + b_ki$ and multiplying it with its complex conjugate \citep{loconte2024sos} resulting in
\begin{align*}\label{eq:sos_app}\tag{SOS} 
    Z^{-1} \cdot \bigg(&\sum\nolimits_{k=1}^{K}\sum\nolimits_{k^\prime=1}^{K^\prime} a_ka_{k'} q_{k}(\boldsymbol{x}) q_{k^\prime}(\boldsymbol{x}) \\
    +&\sum\nolimits_{k=1}^{K}\sum\nolimits_{k^\prime=1}^{K^\prime} b_kb_{k'} q_{k}(\boldsymbol{x}) q_{k^\prime}(\boldsymbol{x})\bigg).
\end{align*}
We use complex mixture weights in our experiments (\cref{sec:experiments}), as they have been shown to facilitate learning \citep{loconte2024sos}. However, our methodological results apply to general SMMs as expressed in \cref{eq:smm}. %
In the following section, we study how to estimate expectations over SMMs which is central to their application to both IS and VI.

\begin{figure*}[t!]
\begin{center}
\begin{tabular}{cccc}
\textbf{ARITS} & \textbf{Rejection} & \textbf{$\boldsymbol{q_+}$} & \textbf{$\boldsymbol{q_-}$}\\
\includegraphics[scale=0.25]{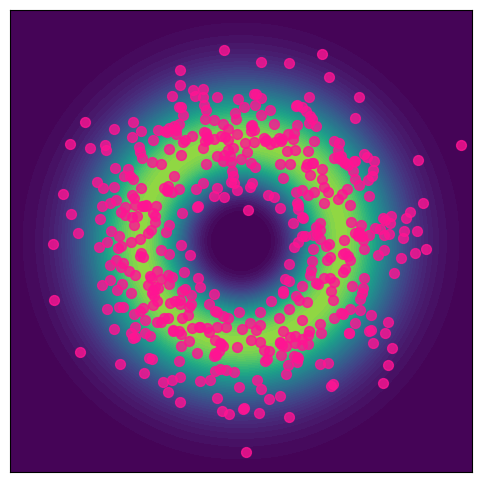} & 
\includegraphics[scale=0.25]{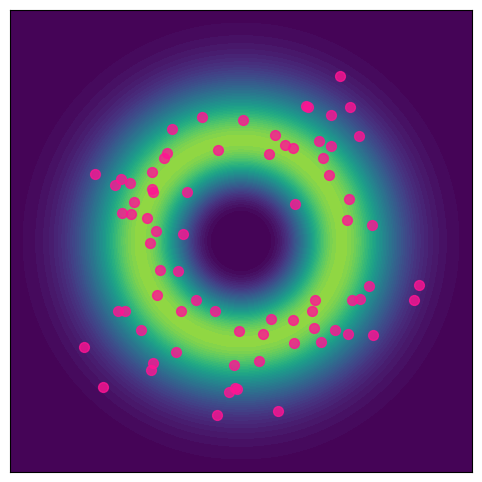} &
\includegraphics[scale=0.25]{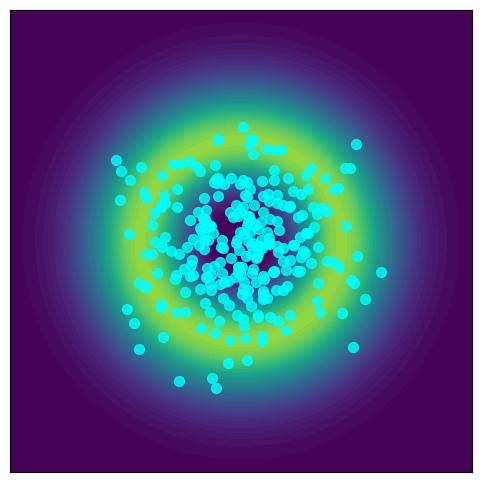} &
\includegraphics[scale=0.25]{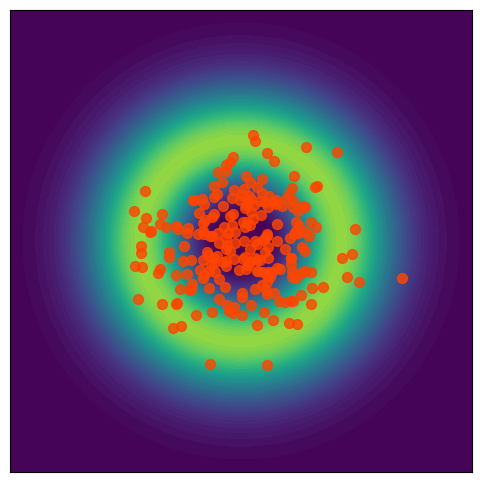}
\end{tabular}
\end{center}
\caption{\textbf{Visual comparison of sampling strategies on a 2D SMM.} ARITS directly simulates samples from the ring. Rejection sampling discards many samples since the average acceptance probability in this example is only around 0.137. $\Delta\text{IS}$ uses samples from both positively and negatively weighted components (depicted in blue and red respectively) to estimate a difference of expectations. All methods are depicted with $S=500$.}\label{fig:sample_comparison}
\end{figure*}

\section{EXPECTATION ESTIMATION WITH SMMs}\label{sec:approximate_inference}

    \begin{algorithm}[!t]
    \small
                    \SetKwInOut{Input}{Input}
                    \SetKwInOut{Output}{Output}
                    \textbf{Input:} {a SMM $q$ (\cref{eq:smm}) and 
        sample budget $S$}\;
                    {\textbf{Output:} $S$ i.i.d. samples from $q$}\;
                    $\mathcal{X}\leftarrow\{\}$\; 
                    \For{$s \in \{1, ..., S\}$}{
                    $\boldsymbol{x}^{(s)}\leftarrow\{\}$\;
                    \For{$d \in \{1, ..., D\}$}{
                        $u \iidsim \mathsf{Uniform}(0,1)$\;
                    $x_d^{(s)}\leftarrow\mathsf{CDF}^{-1}(u\mid \boldsymbol{x}_{<d}^{(s)})$\;
                    }
                $       \mathcal{X}\leftarrow\mathcal{X}\cup\{\boldsymbol{x}^{(s)}\}$\;
        }  
        \Return{$\mathcal{X}$}
    \caption{{\small ARITS($q, S$)}}\label{alg:arits}
\end{algorithm}

We focus on the task of estimating an intractable integral $I$ of the form 
\begin{equation}
    \label{eq:expectation}
    I=\mathbb{E}_{p(\vx)}[f(\vx)]=\int f(\vx) p(\vx) d\vx
\end{equation}
where we have access to an unnormalized density $\widetilde{p}(\vx)$ and $f(\vx)$ is a quantity of interest for the distribution $p(\vx) = \widetilde{p}(\vx)/Z_{p}$.
We next study how to use a given SMM to estimate $I$ via importance sampling using different sampling techniques. 
In \cref{sec:VI_SMMs}, we then discuss how we can \emph{learn} a proposal via black-box VI (\cref{sec:VI_SMMs}), which relies on sampling as a subroutine.

\subsection{How to sample SMMs?}
\label{sec:sampling}
To sample an additive MM (\cref{eq:mm}), 
one can exploit its latent variable semantics and first sample a component $q_k$ with probability proportional to $\alpha_k$, %
then a full instantiation from the corresponding component, i.e., $\boldsymbol{x} \sim q_k$. 
\cref{alg:ancestral-sampling} details this process, called \emph{ancestral sampling} \citep{bishop2006pattern},
whose complexity is linear in the number of components $K$ and dimensions $D$ if we assume each component factorizes into independent marginals
(see \cref{app:complexity}).
To further reduce the variance of the resulting estimator for \cref{eq:expectation}, one can use \textit{stratified sampling} \citep[\cref{alg:stratified}]{mcbook}, which still leverages the latent variable interpretation.
As the mixture coefficients in SMMs can no longer be interpreted as probabilities, neither \cref{alg:ancestral-sampling} nor \cref{alg:stratified} can be directly used, and hence sampling from SMMs requires specialized techniques.

\textbf{Auto-regressive inverse transform sampling (ARITS).} 
For (squared) SMMs, whose components allow for tractable marginalization, one can use ARITS
\citep{loconte2024subtractive, cai2024eigenvi}: Given a variable ordering, we can decompose the joint as $q_{\text{SMM}}(\boldsymbol{x})=\prod_{i=1}^{D}q_{\text{SMM}}(x_i\mid\boldsymbol{x}_{<i})$ and 
 sample each variable \emph{sequentially} by inverting the corresponding conditional CDF (\cref{alg:arits}).
This can be done up to a certain numerical precision $\epsilon$, e.g., via binary search (\Cref{alg:ARITS}).
Due to its sequential nature, ARITS incurs an additional cost that is at least $\mathcal{O}(D)$ times greater than ancestral sampling for classical MMs (\cref{app:complexity}). 
Furthermore, a non-negligible cost can come from inverting the CDF numerically.
In our experiments (\cref{sec:experiments}), we find ARITS to be well-behaved for $\epsilon=10^{-6}$, but we can hardly scale it beyond $D=32$.
This prompts us to look for more scalable ways to sample $q_{\text{SMM}}$ or \textit{avoid sampling it directly}.
    \begin{algorithm}[!t]
    \small
                \SetKwInOut{Input}{Input}
                \SetKwInOut{Output}{Output}
                \textbf{Input:} {a decomposed SMM $q$ (\cref{eq:diffrep}) and 
        sample budget $S$}\;
                \textbf{Output:} {$K \leq S$ i.i.d. samples from $q$}\;
                $M \leftarrow Z / Z_{+},\:\:$ $\mathcal{X} \leftarrow \{ \}$\; 
    \For{$s \in \{1, ..., S\}$}{

        $\boldsymbol{x}^{(s)} \iidsim q_{+}$\;
        $u \iidsim \mathsf{Uniform}(0,1)$\;
        \textbf{if} {$u \leq q(\boldsymbol{x}^{(s)})/(M \cdot q_+(\boldsymbol{x}^{(s)}))$}:
        \:\:$\mathcal{X} \leftarrow \mathcal{X}\cup \{\boldsymbol{x}^{(s)}\}$\;
    }
    
    \Return{$\mathcal{X}$}
    \caption{{\small rejectionSampling($q, S$)}}\label{alg:rejection}
\end{algorithm}

We start by noting that the general form of a SMM (\Cref{eq:smm}) can be rewritten as a  \textbf{\emph{difference of two additive MMs}} \citep{bignami1971note,robert2025simulating}, that is,
\begin{equation}
\tag{$\Delta\text{SMM}$}
\label{eq:diffrep}
    q_{\text{SMM}}(\vx) = {Z^{-1}}\Big( Z_{+} \cdot q_{+}(\vx) - Z_{-} \cdot q_{-}(\vx)\Big) ,
\end{equation}
where $q_{+}(\vx) = \widetilde{q}_{+}(\vx)/Z_{+}$, $q_{-}(\vx) = \widetilde{q}_{-}(\vx)/Z_{-}$ are \emph{additive} mixture PDFs composed of the positively and negatively weighted components of $q_{\text{SMM}}$ respectively.
\Cref{fig:smms-pcs} shows this decomposition for a squared SMM.
$Z_{+} = \int \widetilde{q}_+(\vx)d\vx$ and $Z_{-}=\int \widetilde{q}_-(\vx)d\vx$ are their normalizing constants, and 
$Z = Z_+ - Z_-$ is the normalizing constant of $q_{\text{SMM}}$, as before.
Since $q_+$ and $q_-$ are additive MMs, they are amenable to ancestral sampling (\cref{alg:ancestral-sampling}). This enables the design of more scalable approximate inference routines for SMMs.

\textbf{Rejection sampling with SMMs.} \Cref{eq:diffrep} yields the bound $q_{\text{SMM}}(\vx) \leq (Z_{+}/Z)\, q_{+}(\vx)$, enabling rejection sampling (RS) from $q_{+}$ \citep{bignami1971note} (\Cref{alg:rejection}).
A proposed sample $\vx \sim q_{+}$ is accepted with probability $q_{\text{SMM}}(\vx)/(M\,q_{+}(\vx))$, where $M = Z_{+}/Z$. The expected acceptance probability is $a = Z/Z_{+}$. This avoids auto-regressive sampling, but can suffer when $a$ is small.
While cleverer acceptance schemes can be devised \citep{robert2025simulating}, %
we found that the vanilla rejection sampling scales well and delivers good estimations in our experiments (\cref{sec:experiments}).
For our theoretical analysis, we consider a fixed-budget variant of RS with $S$ proposed samples and denote the (random) number of acceptances by $K$. 
We analyze the variance when estimating $I = \mathbb{E}_{q_{\text{SMM}}}[f(\vx)]$ as $\widehat{I}_{\text{RS}} = \frac{1}{K} \sum_{k=1}^{K} h(\vx^{(k)})$ 
where $\vx^{(k)}$ are the accepted samples resulting from \Cref{alg:rejection}. 

\begin{proposition}[Variance of rejection sampling]\label{proposition:variance_rs}
    Assume $\int h(\vx)^2 q_{\mathrm{SMM}}(\vx)\,d\vx < \infty$ and $K$ is a zero-truncated Binomial RV to avoid zero acceptances, i.e., $K \sim \mathrm{TrBin}(S;a)$. Then,
    $$
    \mathbb{V}_{\substack{\vx \sim q_{\text{SMM}}\\ K \sim \mathrm{TrBin}(S;a)}}
    \left [ \widehat{I}_{\text{RS}} \right ] = \mathbb{V}_{\vx \sim q_{\mathrm{SMM}}}[h(\vx)] \cdot  \gamma(S, a)  \nonumber ,
    $$
    where 
    $\gamma(S, a) = 
    {\sum_{k=1}^{S}\frac{1}{k} \binom{S}{k} a^{k}(1-a)^{S-k} }/({1 - (1-a)^S})
    $. 
\end{proposition}
The proof is in \Cref{app:rejection}. 
Note that RS is unbiased and consistent  \citep{robert1999monte} and that direct i.i.d. sampling (no rejections) %
from $q_{\text{SMM}}$ would have a scaling of $1/S$ deterministically and $\gamma(a,S) \geq 1/S$. As $a=Z/Z_{+}$ goes from $0$ to $1$, $\gamma(a,S)$ goes from $1$ to $1/S$, impacting the MC convergence rate. \Cref{fig:rejection_inflation} (Appendix) illustrates this. We next study an estimation scheme, that \emph{avoids the rejection step}.

\subsection{Importance Sampling with SMMs}\label{sec:IS_SMMs}

It follows from \Cref{eq:diffrep} that we can rewrite any expectation $\mathbb{E}_{q_{\text{SMM}}}[h(\vx)]$,
for an absolutely integrable $h : \mathbb{R}^d \rightarrow \mathbb{R}$, into a \textit{difference of expectations} as 
\begin{equation}\label{eq:dex}
\tag{\dex}
{Z_{+}}/{Z}\cdot \mathbb{E}_{q_{+}} \left [ h(\vx) \right ] - {Z_{-}}/{Z}\cdot \mathbb{E}_{q_{-}} \left [ h(\vx)\right ].
\end{equation}
When approximating $\mathbb{E}_{q_{\text{SMM}}}[h(\vx)]$, we can hence estimate a \emph{weighted difference of expectations} w.r.t. to the additive MMs $q_+$ and $q_-$ via (stratified) ancestral sampling, instead of sampling $q_{\text{SMM}}$ directly.
The \ref{eq:dex} representation has been recently noted in \citet{robert2025simulating} but, to the best of our knowledge, has not been studied for constructing scalable approximate inference schemes with SMMs, which we discuss next for \emph{unnormalized IS} (UIS) \citep{mcbook}:
Given i.i.d. samples from a \emph{proposal of choice} $q$, the UIS estimator approximates \cref{eq:expectation} as
\begin{equation}\label{eq:uis}\tag{UIS}
\widehat{I}_{\text{UIS}} = \sum\limits_{s=1}^{S} f(\vx^{(s)}) {p(\vx^{(s)})}/{q(\vx^{(s)})}, ~~\vx^{(s)} \iidsim q.
\end{equation}
The variance of the UIS estimator (\Cref{eq:uis}) is minimized when $q$ closely matches the integrand, $ |f|p/\int |f|p$. 
In our setting $q$ is chosen to be a (squared) SMM.
Both ARITS and rejection sampling can be used to realize \Cref{eq:uis} with an SMM proposal. %

\paragraph{The \dis{} estimator.} Alternatively, we can use \Cref{eq:dex} to derive a scalable IS estimator based on samples from $q_+$ and $q_-$, \emph{without the need for a rejection criterion}.
This leads to our \dis{} estimator:
\begin{align*}\label{eq:dis}
\tag{\dis}
\widehat{I}_{\Delta\text{IS}} &=  %
\frac{Z_+}{Z}  \frac{1}{S_{+}} \sum\limits_{s=1}^{S_{+}}  f(\vx_{+}^{(s)}) \frac{p(\vx_{+}^{(s)})}{q_{\text{SMM}}(\vx_{+}^{(s)})}\\&
- \frac{Z_-}{Z} \frac{1}{S_{-}} \sum\limits_{s=1}^{S_{-}} f(\vx_{-}^{(s)})\frac{p(\vx_{-}^{(s)})}{q_{\text{SMM}}(\vx_{-}^{(s)})},\\&
\text{with}\quad \vx_{+}^{(s)}\iidsim q_{+}(\vx_{+}), ~~~
\vx_{-}^{(s)} \iidsim q_{-}(\vx_{-}),
\end{align*}
where $S = S_{+} + S_{-}$ denotes the overall sampling budget and samples are drawn i.i.d. from $q_+$ and $q_-$.
With \dis, we can sample both $q_{+}$ and $q_{-}$ via ancestral sampling, without rejection, while still obtaining an unbiased and consistent estimator, as we show next.

\begin{theorem}[Properties of \dis]\label{proposition:properties} 
Under mild assumptions, the \dis{} estimator has the following properties. See \cref{proof:theorem_1} for proofs. \\[2pt]
\textbf{Unbiasedness and strong consistency:} $\Delta\text{IS}$ is unbiased, i.e., $\mathbb{E}_{\substack{\vx_+ \sim q_+\\\vx_- \sim q_-}}[\widehat{I}_{\Delta\text{IS}}] = I$, and it is strongly consistent, $\mathbb{P}_{\substack{\vx_+ \sim q_+\\\vx_- \sim q_-}}(\lim_{\substack{S_+ \rightarrow +\infty\\S_- \rightarrow +\infty}} \widehat{I}_{\method} = I) = 1$. \newline
\textbf{Variance.} 
The variance of $\widehat{I}_{\method}$ is given by
\begin{equation}\label{eq:variance_dex}
    \resizebox{.95\hsize}{!}{$\mathbb{V}[\widehat{I}_{\method}] = \frac{Z_{+}^{2}}{Z^{2}}\frac{1}{S_{+}} \mathbb{V}_{q_{+}}[f(\vx) w(\vx)] + \frac{Z_{-}^{2}}{Z^2} \frac{1}{S_{-}} \mathbb{V}_{q_{-}}[f(\vx) w(\vx)],$} \nonumber
\end{equation} 
where $w(\vx) = \frac{p(\vx)}{q_{\text{SMM}}(\vx)}$.

\textbf{Optimal proposal.} The SMM proposal minimizing the variance is equivalent to the optimal UIS proposal,  
$$q^{\bigstar}(\vx) = \arg \min_{q} \mathbb{V}_{\substack{\vx_+ \sim q_+\\\vx_- \sim q_-}}[\widehat{I}_{\method}] = \frac{|f(\vx)|p(\vx)}{\int |f(\vx)|p(\vx) d \vx}.$$
Further, $\mathbb{V}_{\substack{\vx_+ \sim q_+\\\vx_- \sim q_-}}{\big[}\widehat{I}_{\method}{\big]} = 0$ if and only if $q = q^{\bigstar}$ and $f(\vx) \geq 0$ almost everywhere (or $f(\vx) \leq 0$).
\end{theorem}
    
\textbf{A safer \dis.} Although the optimal proposal of \dis{} matches UIS, their variances differ in general. This can lead to noticeable differences in estimation quality given the same proposal. To see why, consider that \dis{} samples from regions where $q_+$ and $q_-$ are large individually even when $q_{\text{SMM}} \approx 0$, which can induce very large importance weights and result in high variance (see \Cref{fig:sample_comparison}). 
To stabilize estimates, we propose to add a \emph{``safe component''} $q_{\text{safe}}$ to the proposal to guarantee non-negligible mass where $p$ has support:
\begin{equation}
q(\vx) = (1-\beta)\,q_{\text{SMM}}(\vx) + \beta\,q_{\text{safe}}(\vx), \quad \beta \in [0,1).
    \label{eq:safe}
\end{equation}
A ``flat'' $q_{\text{safe}}$ effectively fills low-density valleys of the SMM. This integrates naturally with \dis{} by treating $q_{\text{safe}}$ as part of $q_+$. In spirit, this mirrors the safe adaptive IS (SAIS) literature that mixes a heavy-tailed density into IS proposals \citep{owen2000safe,delyon2021safe,korba2022adaptive}. %

\textbf{A stratified \dis.} 
To further reduce the variance of \dis, we make use of stratified sampling (\cref{alg:stratified}) \citep{mcbook,elvira2019generalized}, as opposed to standard ancestral sampling (\cref{alg:ancestral-sampling}).
Moreover, for a sampling budget $S$, we heuristically fix the sample size for $q_+$ and $q_-$ as $S_+ = \big\lfloor\frac{Z_+}{Z_+ + Z_-} S\big\rfloor$ and $S_- = \big\lfloor\frac{Z_-}{Z_+ + Z_-} S\big\rfloor$ to sample from the mixtures in proportion to their relative contribution to the estimator. 
One could \emph{correlate} samples from $q_+$ and $q_-$ for additional variance reduction; we leave this technique for future work.

\section{BLACK-BOX VI WITH SMMs}\label{sec:VI_SMMs}
We now investigate how we can use the IS estimators we designed in \Cref{sec:IS_SMMs}---ARITS, rejection, and \dis---as subroutines for black-box VI (BBVI). 
From now on, we will simply refer to the SMM proposal as $q_{\theta}$ as a shorthand, unless confusing.
The parameter $\theta$ encompasses all learnable parameters of the SMM, i.e., the mixture weights as well as the parameters of the components.
The main objective of BBVI is to find the optimal surrogate $q_{\theta^{*}}$ that minimizes a given divergence $L(q_\theta, p)$, which often can be written as an expectation $\mathbb{E}_{q_{\theta}}[\ell(\vx;\theta)]$, for a given loss $\ell(\vx;\theta)$ 
During learning, \emph{gradient estimators} are used to approximate $\nabla_{\theta} \mathbb{E}_{q_{\theta}}[\ell(\vx;\theta)]$ via MC \citep{mohamed2020monte}.
We will focus on the commonly used reverse KL divergence (RKL) for which $\ell(\vx; \theta) = \log(q_{\theta}(\vx)/p(\vx))$, but note that our results can be extended %
to other objectives of the form $\mathbb{E}_{q_{\theta}}[\ell(\vx;\theta)]$.
Crucially, \emph{squaring} the SMM allows us to retain a valid PDF during optimization without introducing constraints on the model parameters.
We discuss next how well-studied gradient estimators for BBVI can be combined with the expectation estimators for SMMs defined in \cref{sec:approximate_inference}.

The \textbf{{REINFORCE}} (or \emph{score function}) estimator \citep{glynn1986stochastic,williams1992simple} relies on the log-derivative trick, i.e., $\nabla_{\theta}q_{\theta}(\vx) = q_{\theta}(\vx)\nabla_{\theta}\log q_{\theta}(\vx)$, to construct an unbiased estimator for $\nabla_{\theta} \mathbb{E}_{q_{\theta}}[\ell(\vx;\theta)]$. 
We use REINFORCE with a \emph{leave-one-out control variate} (RLOO) for variance reduction \citep{salimans2014using, kool2019buy}. For the RKL and $S$ samples $\vx^{(s)} \iidsim q_{\theta}$, this estimator is given as
\begin{align*}
&\widehat{\nabla}_{\theta}^{\text{RLOO}} \text{KL}(q_{\theta}||p) = \frac{1}{S}\sum_{s=1}^S {\Big[}{\log \Big(}\frac{q_{\theta}(\vx^{(s)})}{\widetilde{p}(\vx^{(s)})}{\Big)}
- 
\\
&\frac{1}{S-1}\sum_{l \neq s}\log {\Big(}\frac{q_{\theta}(\vx^{(l)})}{\widetilde{p}(\vx^{(l)})}{\Big)}{\Big]} \nabla_{\theta} \log q_{\theta}(\vx^{(s)}) ,\tag{RLOO}
\label{eq:rloo}
\end{align*}
Notably, his estimator does not require the unknown expectation to have a differentiable integrand.

\textbf{$\Delta\text{VI}$.} 
Another option is to obtain a \emph{pathwise gradient estimator} using the reparameterization trick, which might come with lower variance than naive REINFORCE \citep{kingma2013auto, rezende2014stochastic}. 
We show that, following the approach of \citet{morningstar2021automatic} for classical MMs, this is also possible for SMMs, once represented as \ref{eq:diffrep}.
In particular, since both $q_+$ and $q_-$ are additive mixtures, we can apply \emph{stratification} within each mixture, which results in a fully reparameterizable sampling scheme for $q_+$ and $q_-$ respectively \citep{morningstar2021automatic}. 
For the RKL, the objective is 
\begin{align*}
\tag{$\Delta\text{VI}$}
\text{KL}(q_{\theta}||p) = \sum_{k=1}^K \frac{\alpha_k Z_k}{Z} \mathbb{E}_{q_k}\left [ \log\left(\frac{q_{\theta}(\vx)}{\widetilde{p}(\vx)}\right)\right ],
\label{eq:deltavi}
\end{align*}
where $q_k$ denote the individual components of the SMM\footnote{For a squared SMM, $q_k$ denotes a product of two components (\cref{eq:squared_smm}).} (cf. \Cref{eq:smm}) and $Z_k = \int \widetilde{q}_k(\vx)d\vx$. See \Cref{app:delta_vi} for the full derivation.
We refer to the above objective as $\Delta\text{VI}$ to highlight the possibility of negative coefficients combining the individual expectations, which differentiates it from the stratified ELBO (SELBO) for additive mixture models described by \citet{morningstar2021automatic}.
While we apply this treatment to the RKL, we point out that similar estimators for SMMs are possible for other losses expressed as $\mathbb{E}_{q_{\theta}}[\ell(\vx;\theta)]$, such as the Fisher divergence~\citep{yang2019variational,cai2024batch}.
We provide an empirical comparison of the discussed VI strategies in \cref{sec:experiments}.

\section{RELATED WORK}\label{sec:related_work}
Classical MMs have been used extensively in (black-box) VI.
Additionally, there has been work on learning MMs with a variety of divergences beyond KL
\citep{ryu2016convex,el2020enhanced,lambert2022variational,daudel2023monotonic}.  
IS methods often employ mixture proposals, or equivalently a collection of proposals that can be interpreted as a mixture; this approach is known as multiple importance sampling \citep[MIS;][]{veach1995optimally,owen2000safe,elvira2019generalized,sbert2018multiple}. 
MIS methods have been widely applied in graphics~\citep{sbert2018multiple,kondapaneni2019optimal,muller2019neural}. Learning sampling mixtures is the basis of many adaptive importance sampling (AIS) algorithms, e.g., by resampling~\citep{cappe2004population,elvira2017improving}, via expectation-maximization~\citep{cappe2008adaptive}, with MCMC to adapt the proposals~\citep{martino2017layered}, or exploiting geometry of the target~\citep{fasiolo2018langevin,elvira2022optimized} (see \citet{bugallo2017adaptive} for a review). 
Our work thus contributes to many recent works connecting VI and AIS \citep{yao2018yes,domke2018importance,finke2019importance,jerfel2021variational,guilmeau2024adaptive}. 
Further, $\Delta\text{IS}$ can be seen as a linear combination of two MC estimators that allows for negative coefficients. 
Similar constructions have been studied to combine a set of given estimators in the context of multiple IS \citep{kondapaneni2019optimal} and multilevel MC \citep{giles2015multilevel}, but not used as subroutines for learning a SMM.
Concurrently to us, \cite{Martino2025NegativeWeights} studies sampling from SMMs, and derives different estimators, but not in a VI context.

MMs can further be extended to \emph{deep hierarchical mixtures}, also referred to as \emph{probabilistic circuits} \citep[PCs;][]{choi2020probabilistic}. 
Monotonic PCs, i.e., deep additive MMs, have been investigated for VI in specialized settings, such as inference in discrete graphical models \citep{shih2020probabilistic}, quantized continuous distributions \citep{sladek2025approximate} and hierarchical mixtures of VAEs \citep{tan2019hierarchical}.

\looseness=-1
Closer to our work, \citet{cai2024eigenvi} recently proposed \emph{EigenVI}, a VI method that learns squared SMMs with orthogonal basis functions as components. 
EigenVI optimizes the Fisher divergence \citep{hyvarinen2005estimation} between the surrogate and target by solving an eigenvalue problem, which sidesteps stochastic gradient-based optimization. 
Learning with EigenVI is fast, but the components are not learnable and the approach currently does not allow for alternative objectives. 
Moreover, inverting the CDF of the orthogonal basis is harder than doing so for Gaussian components, which we use in our experiments.

\section{EXPERIMENTS}\label{sec:experiments}
\begin{figure}[!t]
  \includegraphics[width=0.24\textwidth]{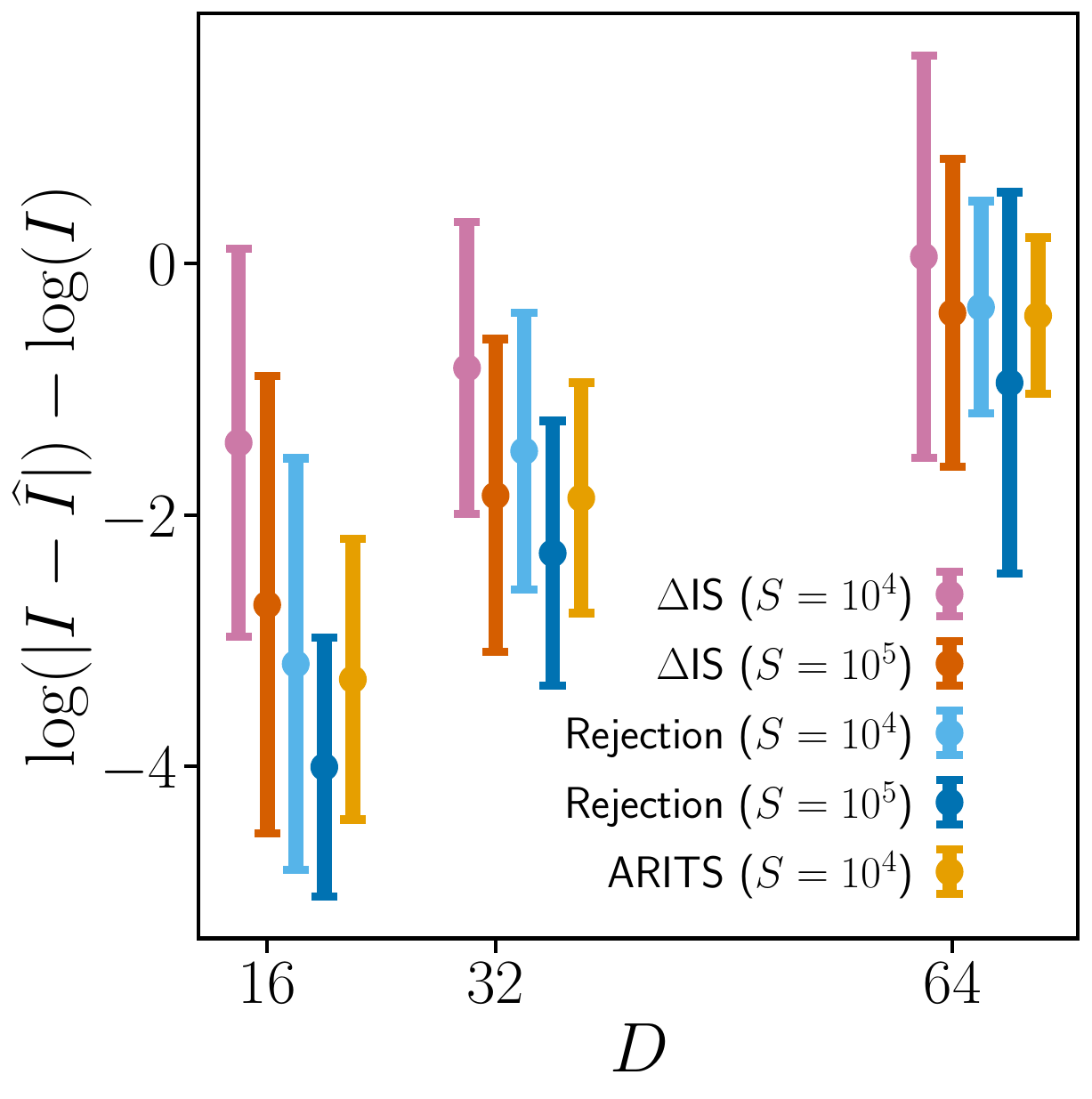}\includegraphics[width=0.24\textwidth]{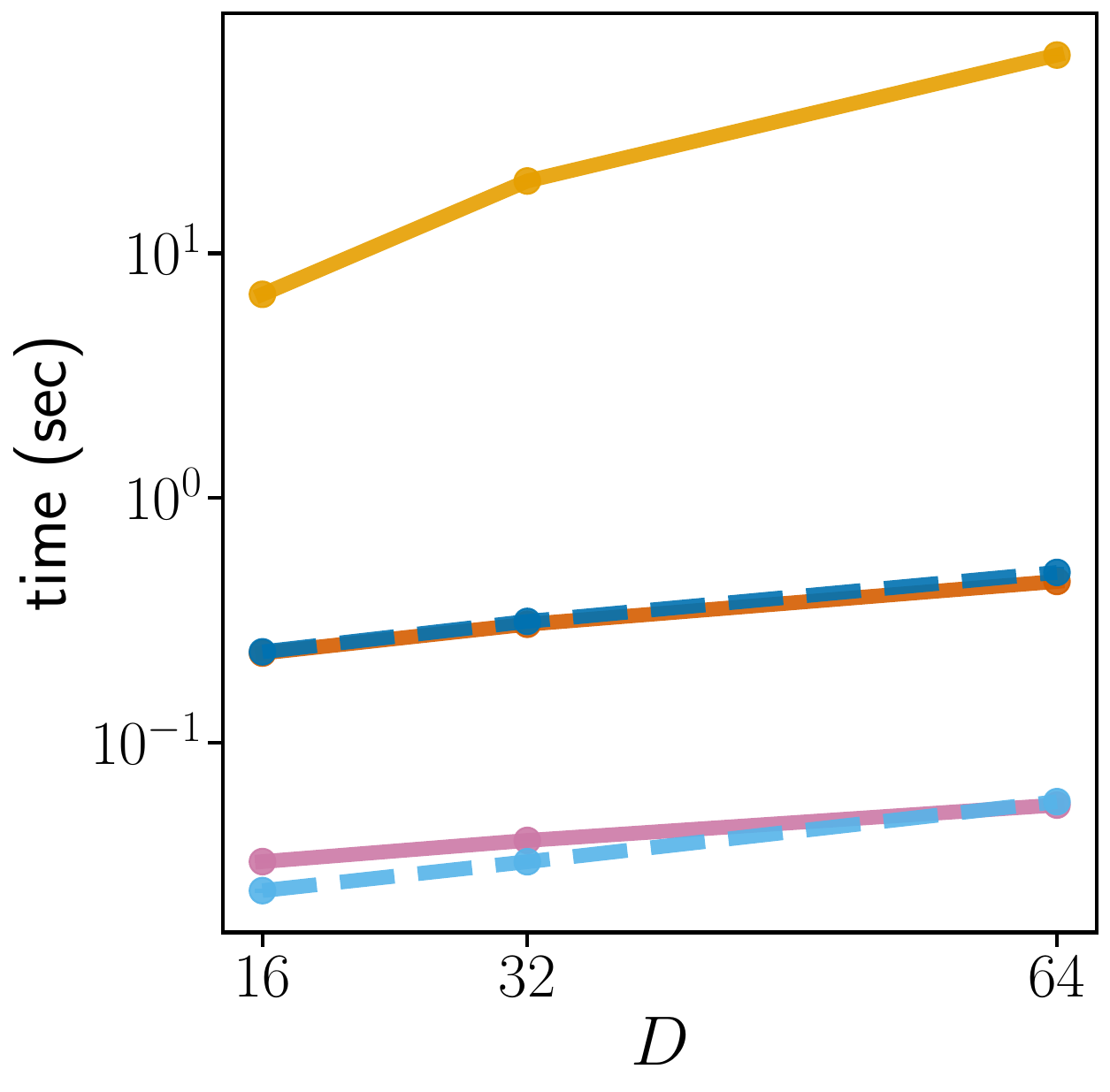}
    \caption{\textbf{Rejection sampling and $\Delta\text{IS}$ can achieve comparable estimation quality to ARITS when given sufficient sampling budget, but can be orders of magnitude faster} in high dimensions as shown for MC estimation. We depict (mean $\pm$ stddev) over $30$ instances. Details in \cref{app:rq_0}.}
\label{runtime_exp}
\end{figure}

\begin{table*}[ht!]
\begin{minipage}{0.32\textwidth}
    \caption{
    Our BBVI methods \ref{eq:deltavi} and \ref{eq:rloo} with rejection and ARITS can recover differently shaped 2D targets while being more parameter-efficient than EigenVI \citep{cai2024eigenvi}. \cref{tab:eigen-vi-quants} in \cref{app:rq21} reports the number of learnable parameters for each model and the corresponding FKL values (which can also be found in \cref{tab:high_dim} for our VI variants). We note that even when fitted densities look similar between SMMs and GMMs, the learned components can greatly differ, see \cref{fig:gmm4vis} for an example. 
    }\label{tab:rq_1}
\end{minipage}\hfill\begin{minipage}{0.65\textwidth}
\setlength{\tabcolsep}{2.5pt}
    \resizebox{\textwidth}{!}{
\begin{tabular}{ccccccc}\toprule
& & &  &\multicolumn{3}{c}{$\text{SMM}^2(\mathbb{C})$}\\ \cmidrule(lr){5-7}
Target & GMM & EigenVI (S) & EigenVI (L) & $\Delta\text{VI}$ &  RLOO (Rej.) & RLOO (ARITS)  \\ \midrule
\includegraphics[scale=0.25]{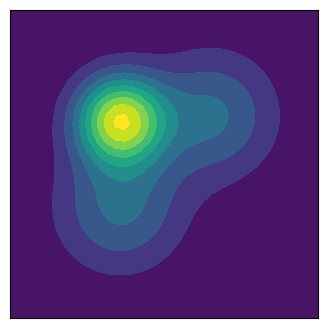} & 
\includegraphics[scale=0.25]{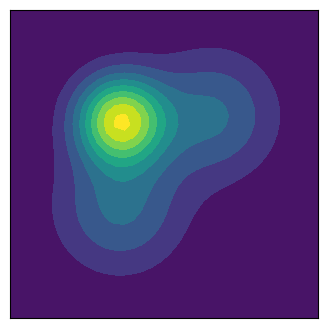} &
\includegraphics[scale=0.45]{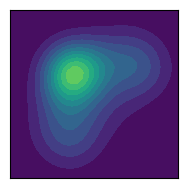}
&
\includegraphics[scale=0.45]{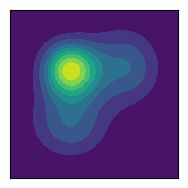}
&
\includegraphics[scale=0.25]{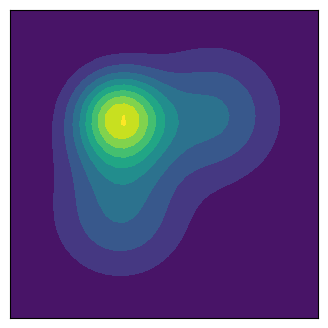} 
&
\includegraphics[scale=0.25]{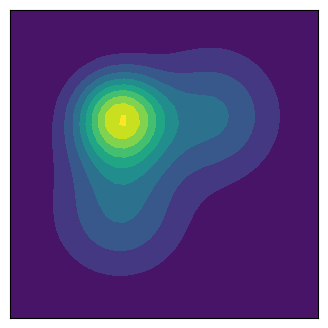} 
&
\includegraphics[scale=0.25]{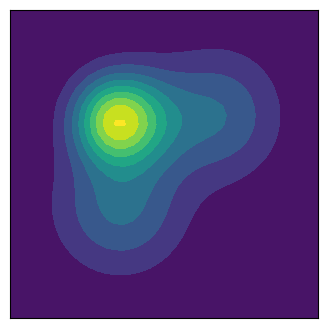}
\\
\includegraphics[scale=0.25]{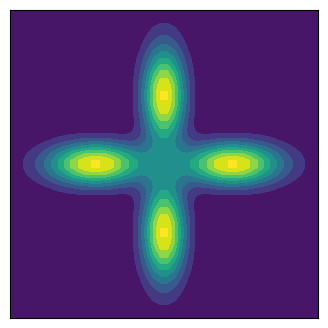} & 
\includegraphics[scale=0.25]{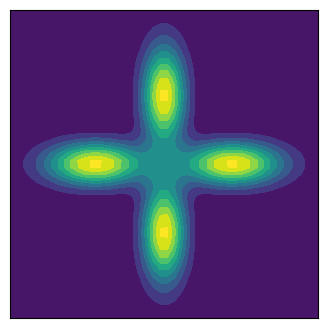} &
\includegraphics[scale=0.45]{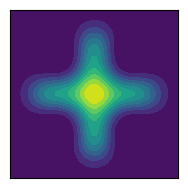}
&
\includegraphics[scale=0.45]{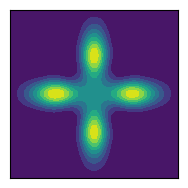}
&
\includegraphics[scale=0.25]{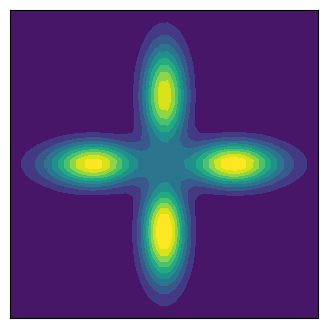} 
&
\includegraphics[scale=0.25]{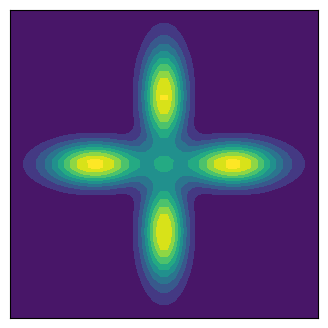} 
&
\includegraphics[scale=0.25]{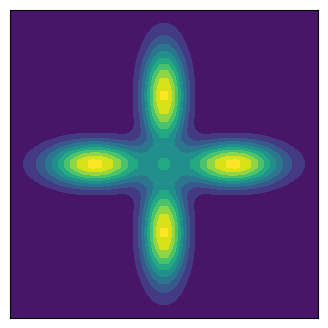}\\
\includegraphics[scale=0.25]{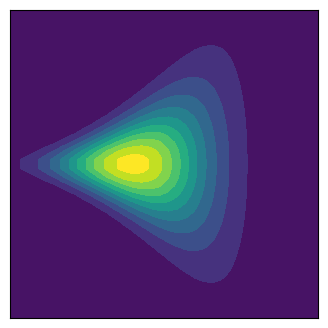} & 
\includegraphics[scale=0.25]{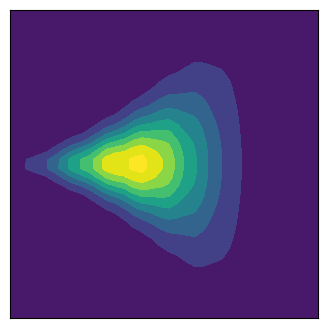} &
\includegraphics[scale=0.45]{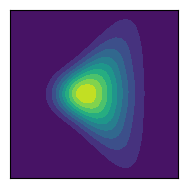}
&
\includegraphics[scale=0.45]{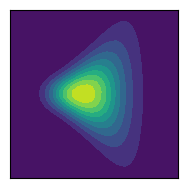}
&
\includegraphics[scale=0.25]{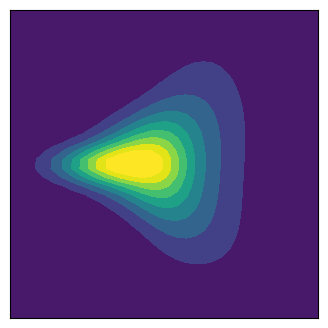} 
&
\includegraphics[scale=0.25]{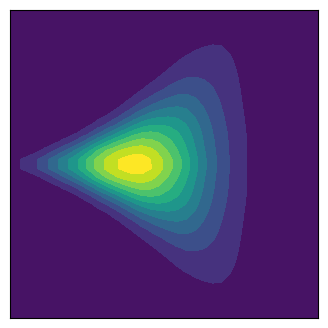} 
&
\includegraphics[scale=0.25]{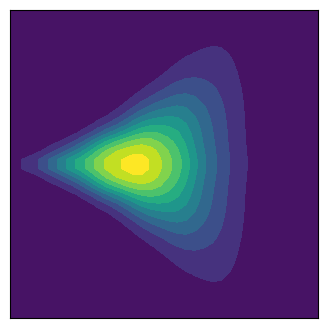}\\
\includegraphics[scale=0.25]{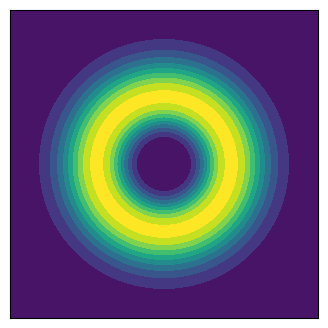} 
& 
\includegraphics[scale=0.25]{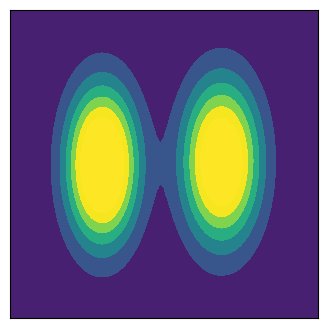}
&
\includegraphics[scale=0.45]{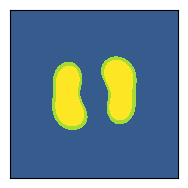}
&
\includegraphics[scale=0.45]{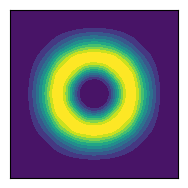}
&
\includegraphics[scale=0.25]{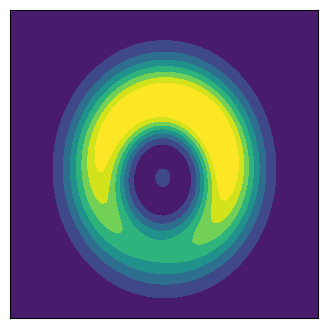} 
&
\includegraphics[scale=0.25]{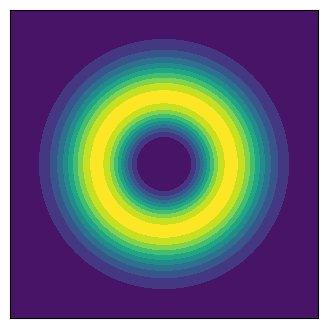} 
&
\includegraphics[scale=0.25]{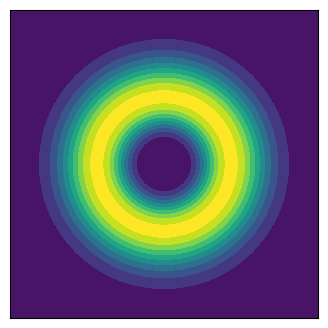}\\\bottomrule
\end{tabular}
}
\end{minipage}
\end{table*}

We now empirically assess how the estimators discussed in this paper perform in three settings: \emph{vanilla MC}, \emph{BBVI}, and \emph{IS with learned proposals}. 
We explore the following research questions: 
(\textbf{RQ 1}) How does the approximation quality and runtime of different expectation estimation strategies compare?
(\textbf{RQ 2}) How do our VI strategies for SMMs compare and what are prominent challenges in comparison to VI with classical MMs?
(\textbf{RQ 3}) How can we perform effective IS with SMMs?
And lastly, (\textbf{RQ 4}) Can we use SMMs to model challenging neuro-symbolic targets?

Throughout the experiments, we choose $q_{\text{SMM}}$ as a squared SMM with complex weights (\cref{sec:smms}) and zero-covariance Gaussian components. 
We use ARITS with tolerance $\epsilon=10^{-6}$ (\Cref{alg:arits}).
The target densities are defined in \Cref{app:targets}.

\paragraph{RQ1) Scaling sampling with SMMs.}
First, we illustrate how ARITS, rejection sampling, and $\Delta\text{IS}$ trade off estimation quality with execution time.
We estimate $I=\mathbb{E}_{q_{\text{SMM}}}[f(\vx)]$ via \emph{standard MC} where
we choose $f(\vx)$ as the density of an unnormalized GMM, 
which allows us to compute the ground-truth expectation in closed form. 
We generate $30$ combinations of $q_{\text{SMM}}$ and $f$, and average the results.
We measure the estimation error as $\log(|\widehat{I}-I|) - \log(I)$ and report the runtime in seconds. 
\Cref{app:rq_0} provides further details.
\Cref{runtime_exp} summarizes the results for SMMs with $6$ components (before squaring). 
 \Cref{tab:table_runtime} discusses experiments with other values of $K$.
Our main insight is the following:
Both $\Delta\text{IS}$ and rejection sampling can achieve similar performance to ARITS given sufficient sampling budget while being much faster in terms of runtime.
Having scalable alternatives to ARITS is crucial for learning in higher dimensional settings.

\paragraph{RQ2) Comparing VI strategies for SMMs.}
We start by comparing the the quality of variational approximations achieved with the three VI strategies discussed in \Cref{sec:approximate_inference} against EigenVI
(\textbf{RQ2.1}), adopting their two-dimensional densities \citep{cai2024eigenvi} to which we add a ring-shaped density, on which we also investigate the effect of sampling budget size for learning (\textbf{RQ2.2}).
Then, we use further synthetic SMM targets to test whether our proposed VI strategies manage to learn negative parameters in high-dimensional settings, given that the target has prominent holes (\textbf{RQ2.3}). 
Lastly, we test SMMs on standard Bayesian logistic regression targets, as used in recent benchmarks \citep{blessing2024beyond} (\textbf{RQ2.4}). %
We include a standard GMM baseline with zero-covariance components (matching the SMMs), learned with the SELBO \citep{morningstar2021automatic} in all experiments.
Further details on the targets and setup can be found in \Cref{app:targets} and \Cref{app:rq_1} respectively.

\begin{figure}[!t]
  \centering
  \includegraphics[width=0.38\linewidth,trim={70 70 70 70},clip]{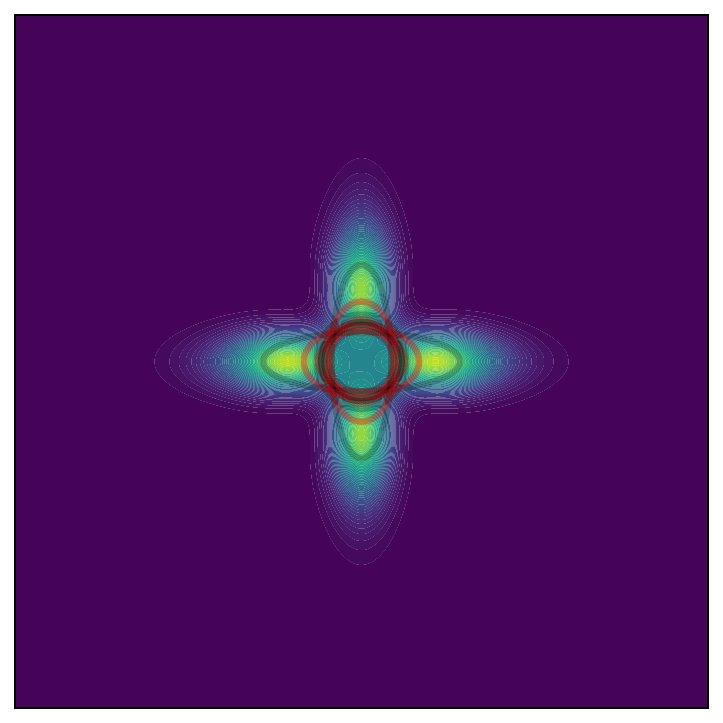}\hspace{20pt}
  \includegraphics[width=0.38\linewidth,trim={70 70 70 70},clip]{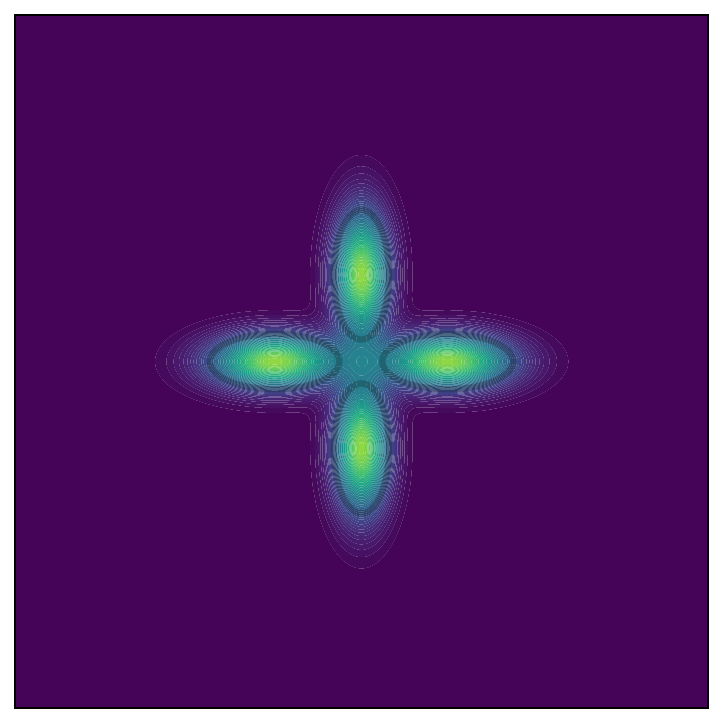}
  \caption{When fitting the same density (\emph{GMM4} target, \cref{tab:rq_1}), SMMs (left) and classical GMMs (right) behave differently as shown by how they place additive (gray) and subtractive (red) Gaussian components.}
  \label{fig:gmm4vis}
\end{figure}

\paragraph{RQ2.1) EigenVI.}
\Cref{tab:rq_1} visualizes our results and \cref{tab:eigen-vi-quants} in \cref{app:rq21} reports the the forward KL (FKL) between the model and the target. %
For EigenVI we use two model sizes: one comparable to the number of learnable parameters of our models (S) and the largest model used in \citet{cai2024eigenvi} (L).
Overall, for the same number of learnable parameters, our estimators for SMMs perform similarly or better than EigenVI.
On \emph{GMM4}, as well as the \emph{Funnel}, squared SMMs obtain comparable metrics to additive GMMs. %
The difference in fit is clearly visible for the \emph{Ring} density as an additive GMM with two components is not expressive enough (\Cref{tab:rq_1}). But also for the \emph{GMM4} target, which can be captured by both models, learned components can differ (see \cref{fig:gmm4vis}).
Using ARITS  with the RLOO gradient estimator is computationally feasible and gives consistently good results, and
RLOO with rejection sampling similarly works well on all targets, while $\Delta\text{VI}$ seems to lag behind.

\begin{figure}
\begin{tabular}{cc}
\small RKL ($\downarrow$) & \small FKL ($\downarrow$)\\
\includegraphics[scale=0.17]{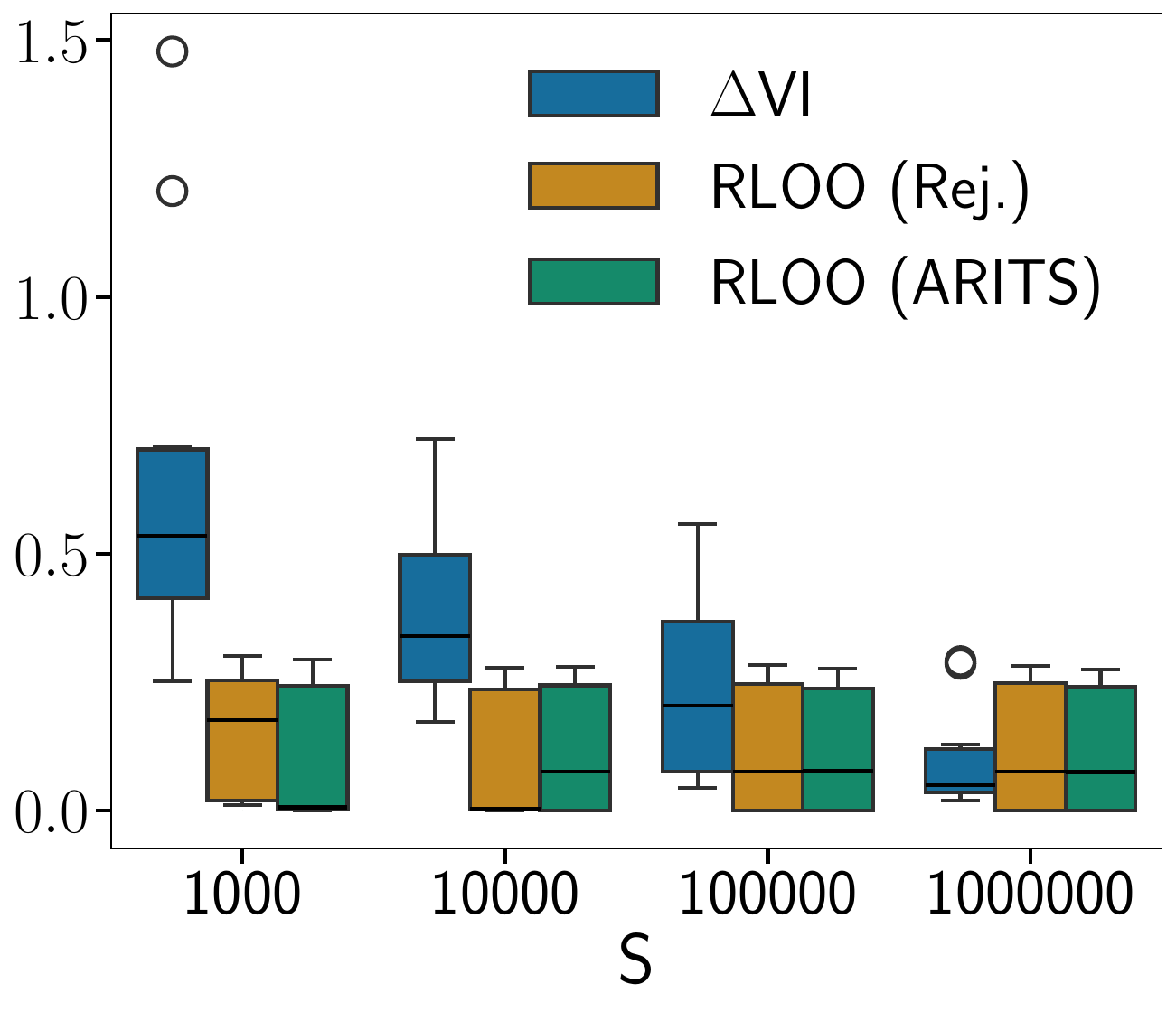} &
\includegraphics[scale=0.17]{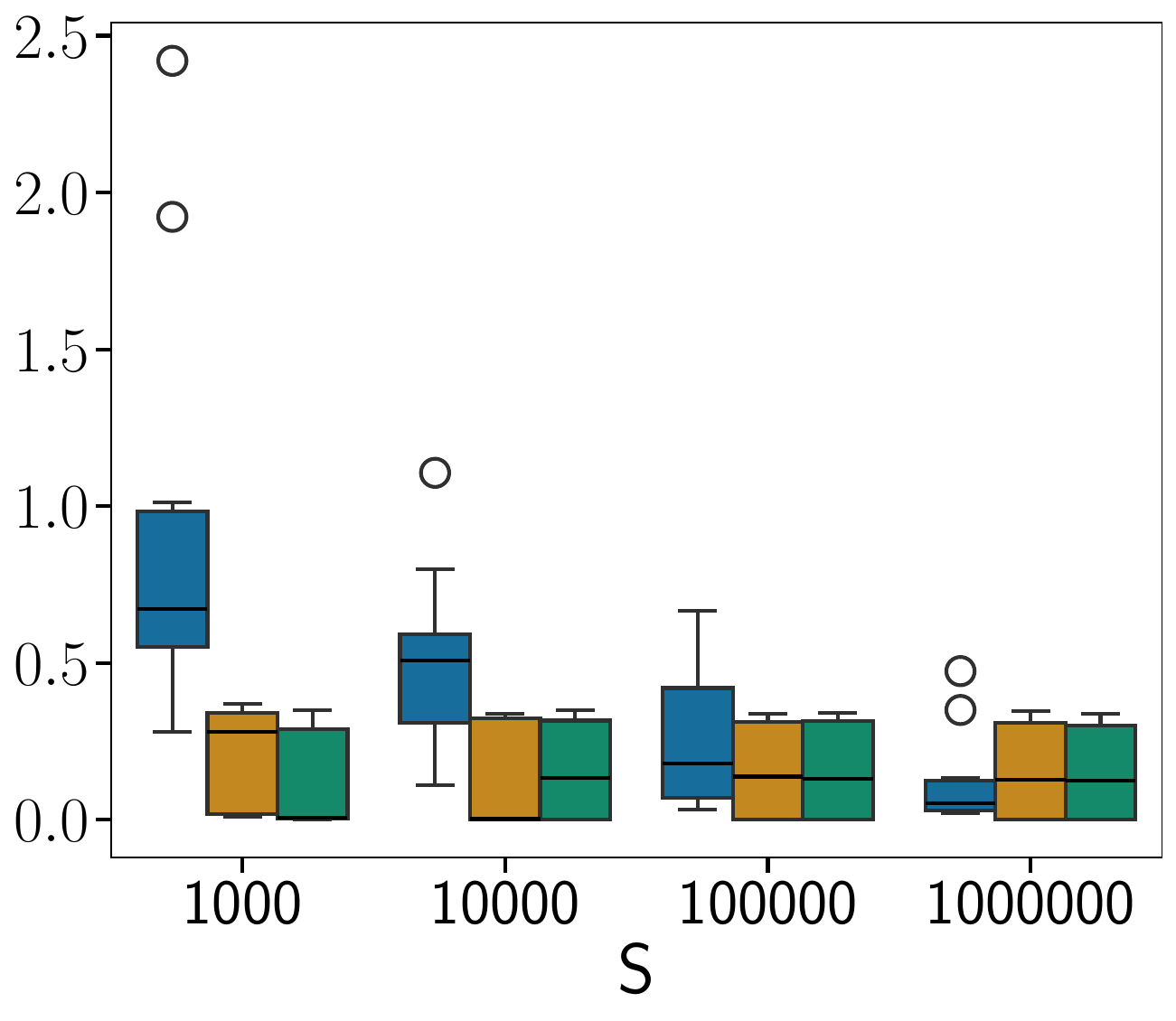}
\end{tabular}
\caption{$\Delta\text{VI}$ requires a higher number of samples than RLOO variants to achieve comparable RKL and FKL. The RKL and FKL values were collected from $10$ models learned with a budget of $S$ samples per step. 
}\label{fig:sample_size_boxplot}
\end{figure}

\paragraph{RQ2.2) Influence of sampling budget.}
We further analyze the behavior of our VI methods by varying the sampling budget $S$ during training.
\Cref{fig:sample_size_boxplot} shows how this impacts the RKL and FKL on the \emph{Ring} target.
We find (1) the RLOO variants perform well even for a substantially smaller sampling budget than the $10^5$ samples used for \Cref{tab:rq_1} and (2) $\Delta\text{VI}$ needs a higher sampling budget to perform similarly to RLOO. Interestingly, at $10^6$ samples per update step $\Delta\text{VI}$ manages to capture a ring-like shape more consistently than the RLOO variants. 
See \Cref{app:rq22} for a visualization of the models used for \Cref{fig:sample_size_boxplot}. 

\paragraph{RQ2.3) Higher-dimensional SMM targets.}
Next, we test our VI schemes on higher-dimensional targets that require learning negative mixture weights.
To this end, we generate \emph{Hollow} spheres, defined as squared SMMs of varying dimensionality with a substantial influence of negative components (see \cref{app:targets} for their definitions). 
Rejection sampling on the resulting targets yields a rejection rate between 80\% and 90\%.
The aim of this setup is to understand whether we can effectively learn to subtract density with our BBVI methods.
Our main insights match the two-dimensional targets: Both RLOO variants manage to recover the shape well while the $\Delta\text{VI}$ models are not up to par.
Interestingly, we noticed that optimization was very sensitive to the initialization across all VI variants.
See \cref{app:rq23} for examples of varying initializations and the resulting models. 
How to design robust initializations for BBVI with SMMs is an important open question.

\begin{table*}

\caption{
On densities with more prominent holes such as \emph{Ring} and \emph{Hollow}, our \ref{eq:deltavi} and \ref{eq:rloo} variants with SMMs deliver better performance than classical GMMs, while being comparable on other densities that do not necessarily require to subtract probability mass. See \cref{tab:rq_1} for visual examples of the two-dimensional fits.
}\label{tab:high_dim}
\resizebox{\textwidth}{!}{
\begin{tabular}{@{}lrrrrrrrr}
\toprule
 & \multicolumn{2}{c}{GMM} & \multicolumn{2}{c}{SMM + $\Delta$VI} & \multicolumn{2}{c}{SMM + RLOO (Rej.)} & \multicolumn{2}{c}{SMM + RLOO (ARITS)} \\
\cmidrule(lr){2-3}
 \cmidrule(lr){4-5}
 \cmidrule(lr){6-7}
 \cmidrule(lr){8-9}
 Target ($D$) & \multicolumn{1}{c}{RKL ($\downarrow$)} & \multicolumn{1}{c}{FKL ($\downarrow$)} & \multicolumn{1}{c}{RKL ($\downarrow$)} & \multicolumn{1}{c}{FKL ($\downarrow$)} & \multicolumn{1}{c}{RKL ($\downarrow$)} & \multicolumn{1}{c}{FKL ($\downarrow$)} & \multicolumn{1}{c}{RKL ($\downarrow$)} & \multicolumn{1}{c}{FKL ($\downarrow$)} \\\midrule
GMM3 ($2$) & $1.9 \cdot 10^{-6} \pm 1.3 \cdot 10^{-5}$ & $6.0 \cdot 10^{-6} \pm 1.6 \cdot 10^{-5}$ & $2.4 \cdot 10^{-4} \pm 6.8 \cdot 10^{-5}$ & $2.0 \cdot 10^{-4} \pm 8.4 \cdot 10^{-5}$ & $1.8 \cdot 10^{-4} \pm 6.7 \cdot 10^{-5}$ & $2.3 \cdot 10^{-4} \pm 5.4 \cdot 10^{-5}$ & $2.7 \cdot 10^{-4} \pm 6.8 \cdot 10^{-5}$ & $2.6 \cdot 10^{-4} \pm 9.1 \cdot 10^{-5}$ \\
GMM4 ($2$) & $1.4 \cdot 10^{-5} \pm 1.3 \cdot 10^{-5}$ & $1.9 \cdot 10^{-5} \pm 9.8 \cdot 10^{-6}$ & $5.2 \cdot 10^{-3} \pm 4.0 \cdot 10^{-4}$ & $5.5 \cdot 10^{-3} \pm 2.9 \cdot 10^{-4}$ & $1.1 \cdot 10^{-4} \pm 4.1 \cdot 10^{-5}$ & $1.1 \cdot 10^{-4} \pm 4.8 \cdot 10^{-5}$ & $5.8 \cdot 10^{-5} \pm 4.1 \cdot 10^{-5}$ & $7.4 \cdot 10^{-5} \pm 3.9 \cdot 10^{-5}$ \\
Funnel ($2$) & $3.5 \cdot 10^{-3} \pm 3.0 \cdot 10^{-4}$ & $3.7 \cdot 10^{-3} \pm 3.5 \cdot 10^{-4}$ & $2.7 \cdot 10^{-2} \pm 1.7 \cdot 10^{-3}$ & $4.1 \cdot 10^{-2} \pm 8.3 \cdot 10^{-4}$ & $8.4 \cdot 10^{-4} \pm 1.8 \cdot 10^{-4}$ & $8.2 \cdot 10^{-4} \pm 1.9 \cdot 10^{-4}$ & $1.1 \cdot 10^{-3} \pm 1.8 \cdot 10^{-4}$ & $1.3 \cdot 10^{-3} \pm 1.7 \cdot 10^{-4}$ \\
Ring ($2$) & $2.9 \cdot 10^{-1} \pm 2.3 \cdot 10^{-3}$ & $3.2 \cdot 10^{-1} \pm 2.2 \cdot 10^{-3}$ & $8.2 \cdot 10^{-2} \pm 1.3 \cdot 10^{-3}$ & $8.4 \cdot 10^{-2} \pm 1.2 \cdot 10^{-3}$ & $9.3 \cdot 10^{-6} \pm 1.6 \cdot 10^{-5}$ & $1.2 \cdot 10^{-5} \pm 1.7 \cdot 10^{-5}$ & $5.5 \cdot 10^{-6} \pm 1.6 \cdot 10^{-5}$ & $2.0 \cdot 10^{-6} \pm 1.0 \cdot 10^{-5}$ \\ \midrule
Hollow ($16$) & $2.8 \cdot 10^{-1} \pm 3.3 \cdot 10^{-3}$ & $1.8 \cdot 10^{-1} \pm 1.9 \cdot 10^{-3}$ & $2.3 \cdot 10^{-1} \pm 2.8 \cdot 10^{-3}$ & $2.0 \cdot 10^{-1} \pm 2.2 \cdot 10^{-3}$ & $3.7 \cdot 10^{-5} \pm 2.6 \cdot 10^{-5}$ & $2.6 \cdot 10^{-5} \pm 1.8 \cdot 10^{-5}$ & $2.2 \cdot 10^{-5} \pm 2.0 \cdot 10^{-5}$ & $3.1 \cdot 10^{-5} \pm 3.8 \cdot 10^{-5}$ \\
Hollow ($32$) &  $1.9 \cdot 10^{-1} \pm 2.2 \cdot 10^{-3}$ & $1.4 \cdot 10^{-1} \pm 1.8 \cdot 10^{-3}$ & $1.8 \cdot 10^{-1} \pm 1.9 \cdot 10^{-3}$ & $2.0 \cdot 10^{-1} \pm 1.7 \cdot 10^{-3}$ & $1.8 \cdot 10^{-5} \pm 2.3 \cdot 10^{-5}$ & $2.8 \cdot 10^{-5} \pm 2.4 \cdot 10^{-5}$ & $2.5 \cdot 10^{-5} \pm 1.7 \cdot 10^{-5}$ & $1.8 \cdot 10^{-5} \pm 1.8 \cdot 10^{-5}$ \\
Hollow ($64$) & $2.3 \cdot 10^{-1} \pm 3.3 \cdot 10^{-3}$ & $1.4 \cdot 10^{-1} \pm 1.7 \cdot 10^{-3}$ & $2.3 \cdot 10^{-1} \pm 2.7 \cdot 10^{-3}$ & $2.1 \cdot 10^{-1} \pm 1.1 \cdot 10^{-3}$ & $7.6 \cdot 10^{-5} \pm 2.7 \cdot 10^{-5}$ & $6.8 \cdot 10^{-5} \pm 2.9 \cdot 10^{-5}$ & / & / \\ \midrule
Funnel ($10$) & $7.9 \cdot 10^{-2} \pm 1.3 \cdot 10^{-3}$ & $1.0 \cdot 10^{0} \pm 3.5 \cdot 10^{-1}$ & $3.0 \cdot 10^{-1} \pm 1.5 \cdot 10^{-3}$ & $2.6 \cdot 10^{1} \pm 5.3 \cdot 10^{0}$ & $2.7 \cdot 10^{-1} \pm 1.4 \cdot 10^{-3}$ & $1.9 \cdot 10^{1} \pm 2.9 \cdot 10^{0}$ & / & / \\
\bottomrule
\end{tabular}
}

\end{table*}

\paragraph{RQ2.4) Standard high-dimensional targets.}
We consider a $10$-dimensional \emph{Funnel} \citep{neal2003slice}
and various Bayesian logistic regression (BLR) posteriors, which are very common benchmarks in statistics and VI. 
In particular, for BLR we use the datasets \emph{GermanCredit}, \emph{BreastCancer}, \emph{Ionosphere} and \emph{Sonar} from  \citet{blessing2024beyond}. 
In \Cref{tab:logreg_elbo}, we report estimated ELBOs for RLOO (with rejection), $\Delta$VI, and the GMM baseline.
We do not report RLOO (ARITS) due to it exceeding reasonable runtimes. 
On these targets, all methods perform similarly. 
Plots of bivariate conditionals on a grid of values suggest that these posteriors may be similar to Gaussians, which is also a fact that has been noted in the Bayesian statistics literature \citep{10.1214/16-STS581}. 
For these posteriors, we noticed that it was difficult, and very dependent on initialization, to obtain non-negligible negative contributions $Z_{-}/(Z_{+}-Z_{-})$. See \Cref{app:exp_setup} for details about hyperparameters and learned models.
For the $10$-dimensional \emph{Funnel}, we obtain better approximations with GMMs than with SMMs. 
Empirically, we found the SMM components to closely cluster together, resulting in fits that did not cover the spread-out funnel shape well.

\begin{table}[!t]
\caption{On BLR posteriors, GMMs and SMMs perform similarly in terms of ELBO. \label{tab:logreg_elbo}}
    \resizebox{\linewidth}{!}{
    \begin{tabular}{@{}lccc}
\toprule
Density & GMM & SMM + $\Delta$VI & SMM + RLOO (Rej.) \\
\midrule
credit 
    & $-1.72 \std{ 0.0231 }$
    & $-1.74 \std{ 0.0102 }$
    & $-1.71 \std{ 0.01 }$
    \\
ionosphere 
    & $-124 \std{ 0.0476 }$
    & $-124 \std{ 0.0382 }$
    & $-124 \std{ 0.0295 }$
    \\
breastcancer 
    & $-67.3 \std{ 0.0358 }$
    & $-67.6 \std{ 0.0311 }$
    & $-67.5 \std{ 0.0335 }$
    \\
sonar 
    & $-137 \std{ 0.0323}$ 
    & $-138 \std{ 0.0345}$
    & $-138 \std{ 0.0328}$
    \\
\bottomrule
\end{tabular}
}
\end{table}

\paragraph{RQ3) IS with SMMs.}
We test the expectation estimation strategies discussed in \cref{sec:approximate_inference} for \emph{normalizing constant estimation:}
Given the unnormalized target density $\widetilde{p}(\vx)$, we aim to estimate $I = \int\widetilde{p}(\vx)d\vx$ via an IS estimator based on a proposal $q_{\text{SMM}}$. %
In this setting, we first show with synthetic proposals how $\Delta\text{IS}$ can result in high variance compared to the standard UIS estimator and how a safe component can mitigate this issue in practice (\textbf{RQ3.1}). 
We then use (\textbf{RQ3.2}) \emph{learned} SMM proposals from from the previous RQ, and compare the resulting estimation quality with ARITS, rejection, $\Delta\text{IS}$, and IS estimators using a standard GMM proposal.

\textbf{RQ3.1) Safe $\Delta\text{IS}$.} As we show in \cref{app:rq_31}, $\Delta\text{IS}$ can result in high variance when used on targets with high negative contribution, such as the \emph{Ring} and \emph{Hollow} targets. We mitigate this by mixing the proposal with a safe component (\cref{eq:safe}). We choose $q_{\text{safe}}$ as a flat Gaussian $\mathcal{N}(0, \sigma_{\text{safe}}^2I^{D \times D})$, where $I^{D \times D}$ is the $D$-dimensional identity and $\sigma_{\text{safe}}$ is a hyperparameter that we set heuristically. Even with a small mixing coefficient $\beta$, a safe component can substantially reduce the variance of $\Delta\text{IS}$. How to automatically construct a safe component for a given problem is an interesting open question.

\textbf{RQ3.2) IS with learned proposals.}
Lastly, we complete a full SMM-based approximate inference pipeline and use learned SMM proposals for normalizing constant estimation.
We use proposals learned via rejection from \Cref{tab:high_dim} and compare against IS with learned GMMs.
For $\Delta\text{IS}$, we again choose the safe component as $\mathcal{N}(0, \sigma_{\text{safe}}^2I^{D \times D})$. We select $\sigma_{\text{safe}}$ and $\beta$ via grid search based on the empirical variance (see \cref{app:rq_32} for details).
\Cref{tab:rq_32} summarizes the results and 
\Cref{fig:rq32} provides boxplots based on different sample sizes.
The results mirror the previous RQs: when negative components are useful to represent a target density, ARITS performs the best and rejection sampling is a scalable alternative.
When targets do not require subtraction, GMMs give better estimates but SMMs do not lag too much behind.

\begin{figure}[!t]
\resizebox{0.49\textwidth}{!}{
\begin{tabular}{ccc}
\textbf{Target} & $\boldsymbol{K=2}$ & $\boldsymbol{K=8}$\\
\includegraphics[scale=0.15]{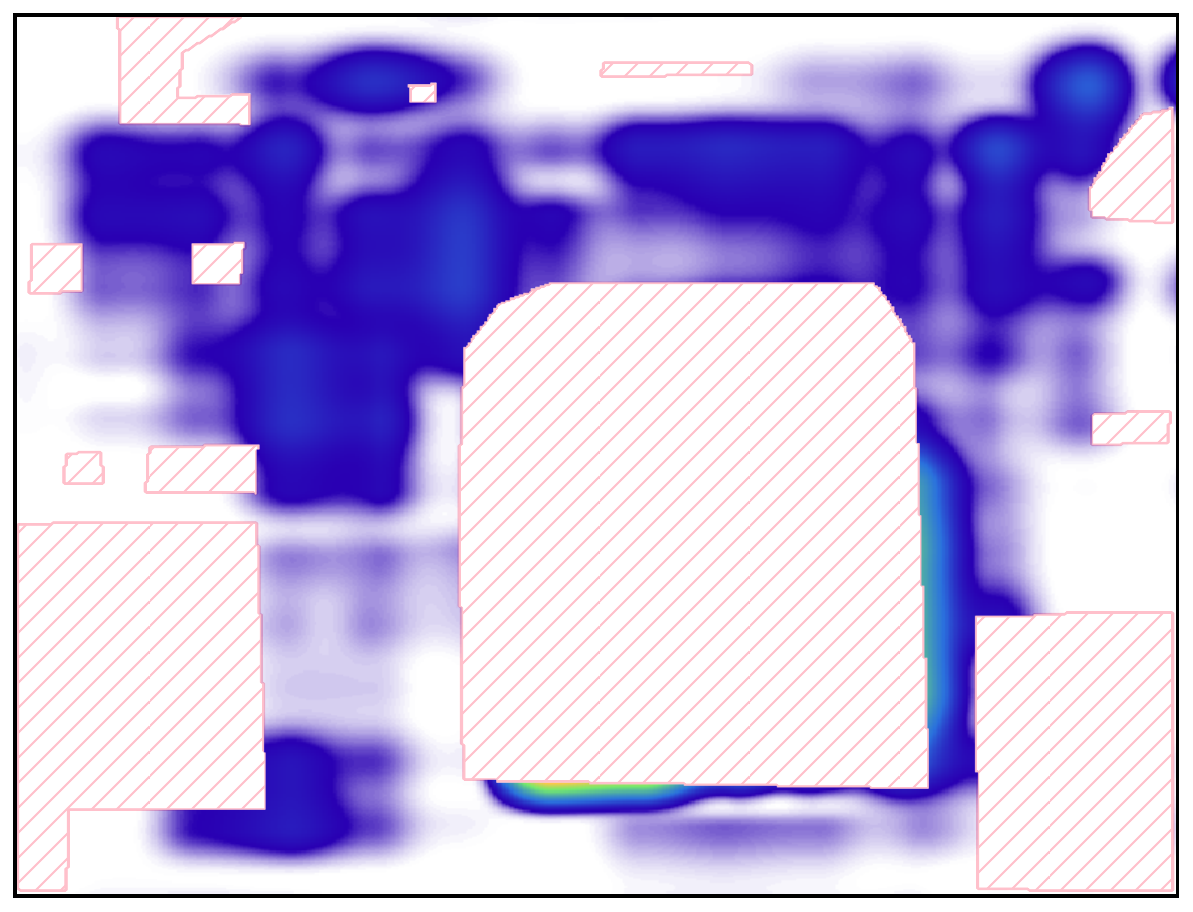} &
\includegraphics[scale=0.15]{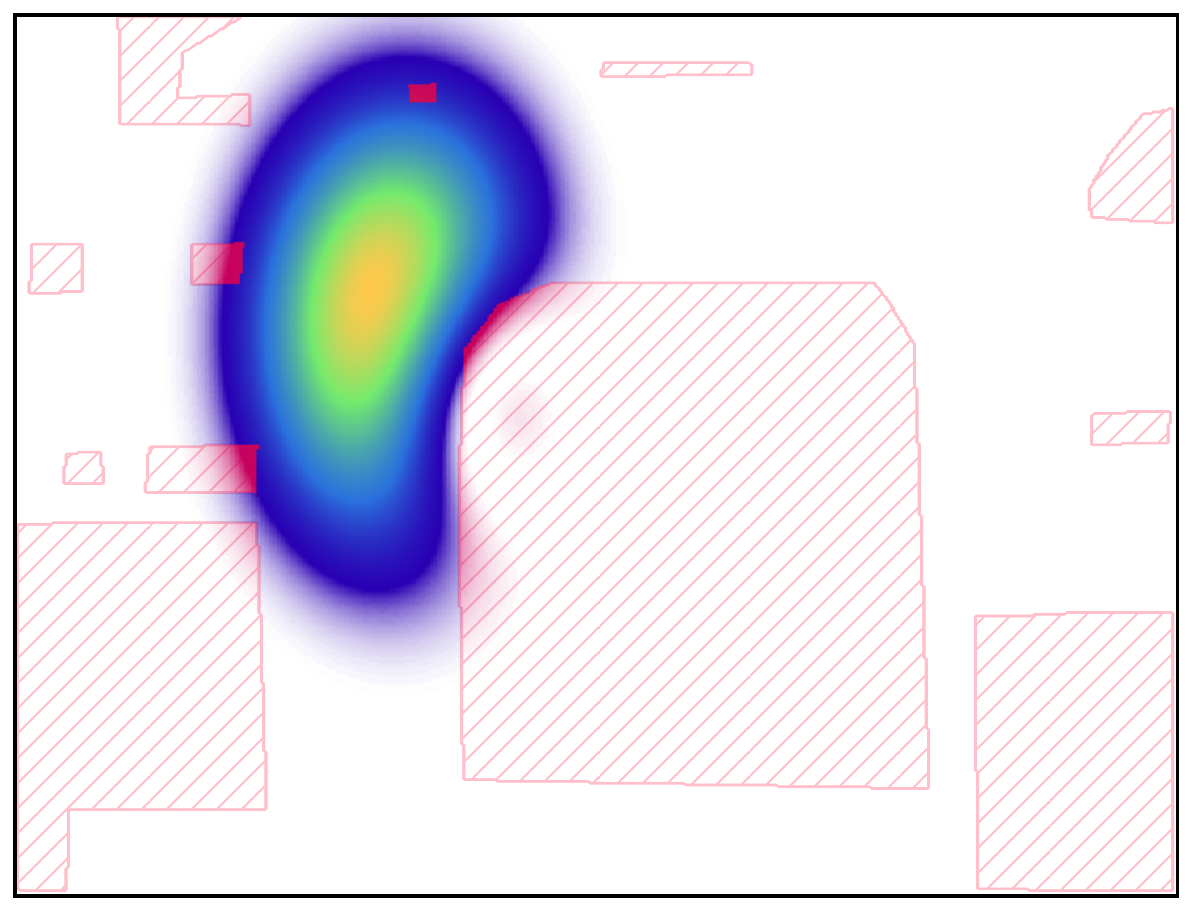}
& 
\includegraphics[scale=0.15]{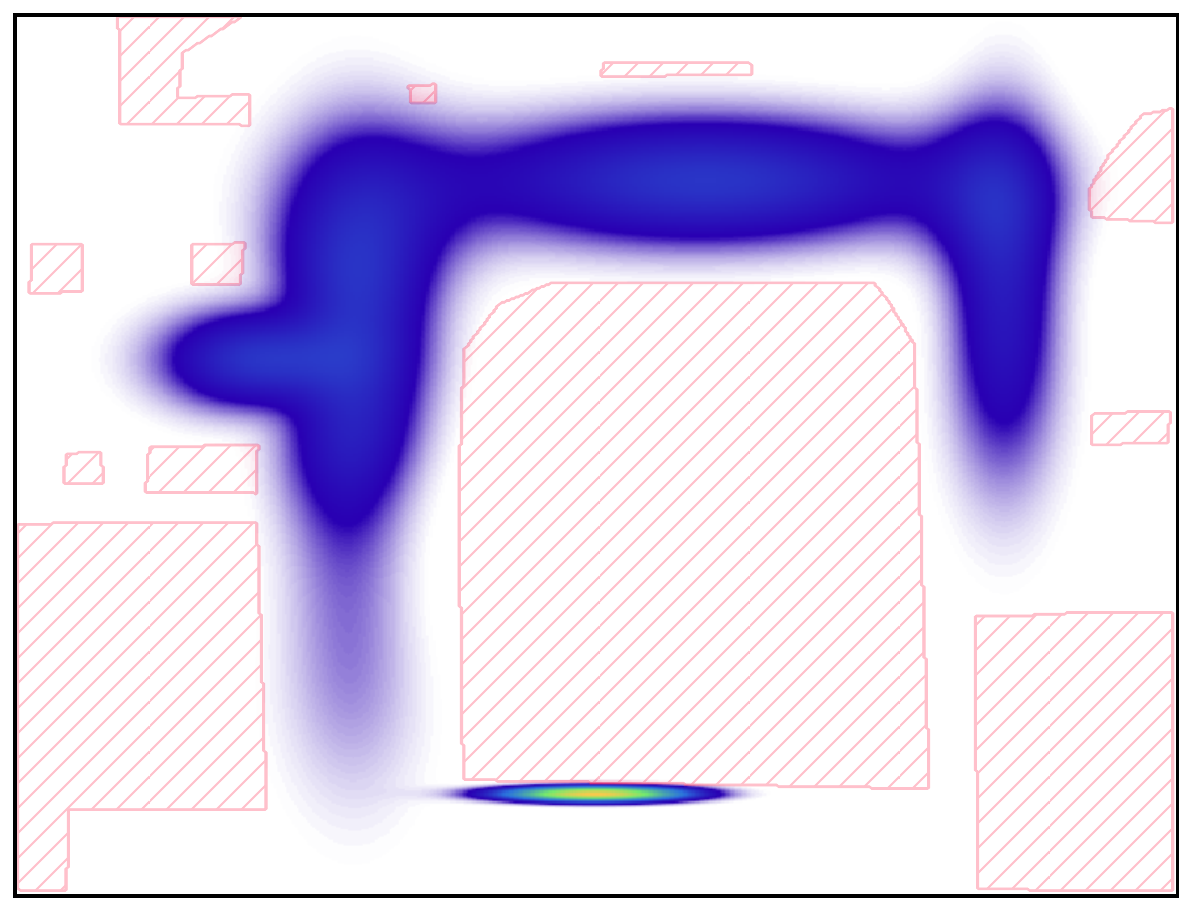}\\
\includegraphics[scale=0.15]{figures/pal_app/scenario_2/pal_target_2.pdf} &
\includegraphics[scale=0.15]{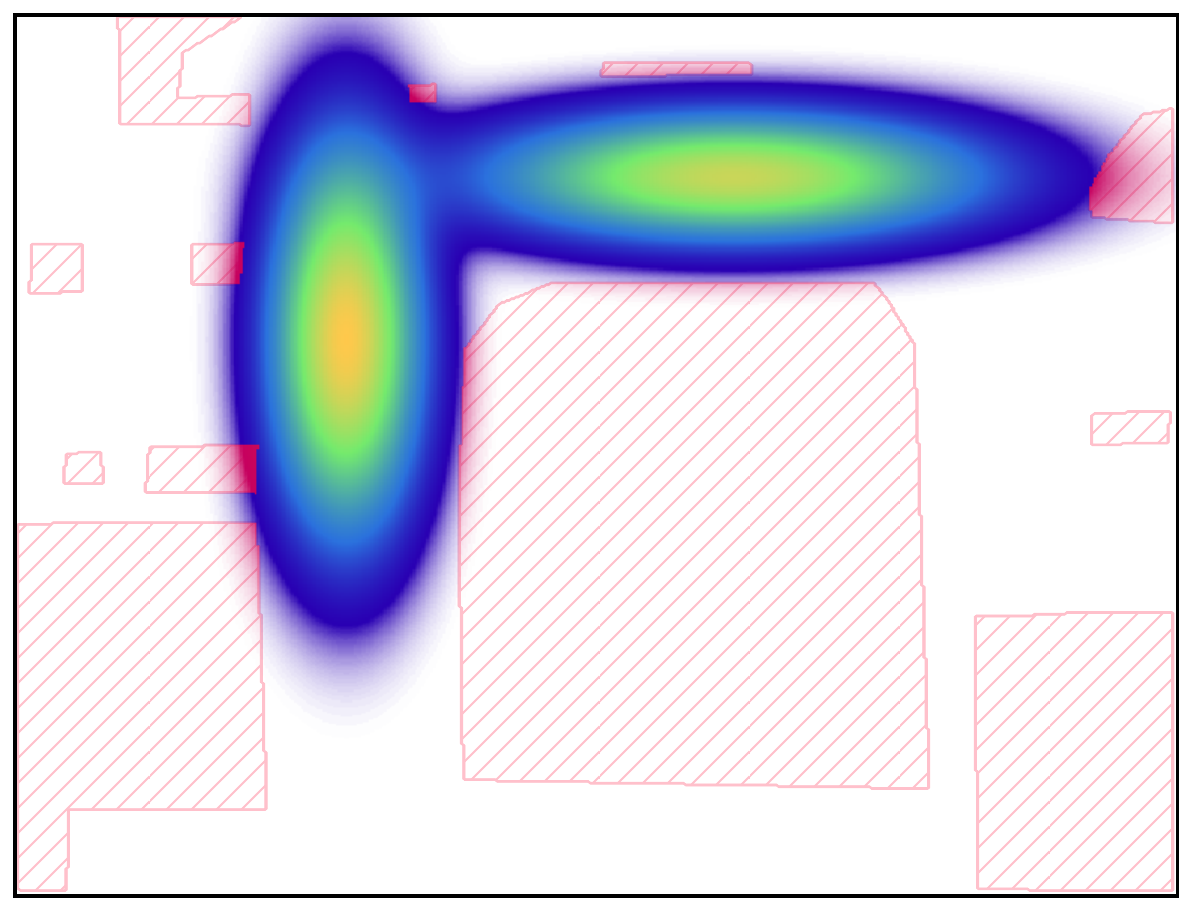}
& 
\includegraphics[scale=0.15]{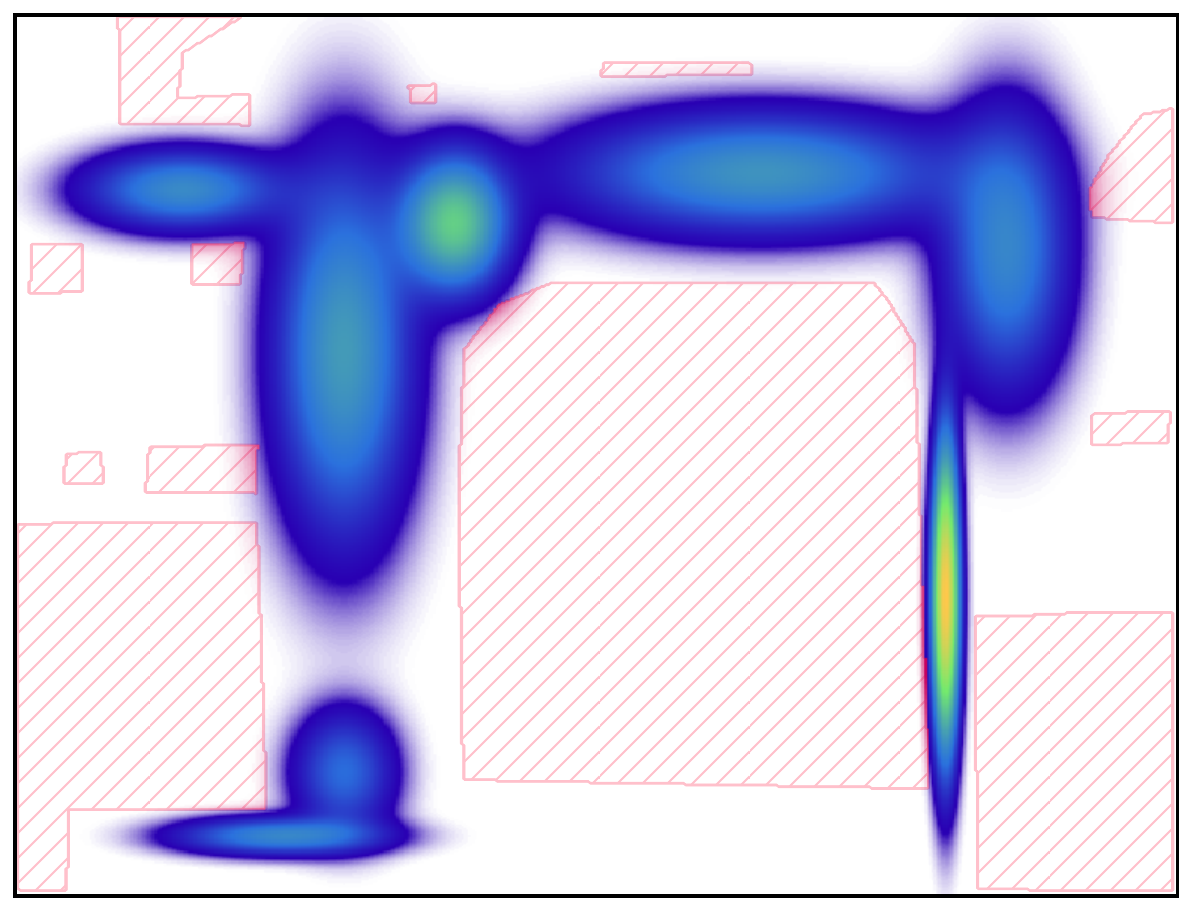}
\end{tabular}
}
\caption{A second scenario from the SDD dataset. \textbf{SMMs (top)} and \textbf{GMMs (bottom)} result in very different fits to the target for the same component budget. We once again observe that for $K=2$, the SMM effectively learns to use subtraction in order to model the absence of density induced by a constraint.}
\label{fig:pal_2}
\end{figure}
\textbf{RQ4) SMMs for learning under constraints.} 
Lastly, we evaluate SMMs on a real-world scenario where they are intuitive and effective VI proposals. 
In particular, we consider a \emph{neuro-symbolic} setting, where inputs violating domain constraints results in $0$-density. 
It is natural to model such densities with SMMs, as they can learn to ``cut'' the density of invalid areas. 
Consider the targets in \cref{fig:pal_fig_1,fig:pal_2} fit with the \emph{probabilistic algebraic layer} (PAL) \citep{kurscheidt2025probabilistic} to the \emph{Stanford Drone Dataset} (SDD) \citep{robicquet2016learning}, which captures the walkable area of a map. 
The densities are naturally constrained due to obstacles, such as a roundabout and buildings. These constraints were recently manually annotated by \citet{kurscheidt2025probabilistic}, resulting in challenging benchmark densities. 
We find that SMMs trained with RLOO + rejection can learn to use subtraction to approximately adhere to some of the domain constraints, even at a low component budget. 
However, we also find that general difficulties in fitting SMMs observed in the previous subsections persist: Akin to the $10$-dimensional funnel, we find it difficult to learn SMMs that cover the full target, even at high component budgets.
See \cref{app:rq_4} for experimental details, results for additional values of $K$, and a quantitative comparison to GMMs in terms of ELBO.
Using SMMs to model domain constraints is a promising future direction.
It will be interesting to explore in future work how SMMs can be used in rare-event probability estimation in other safety-critical systems, such as aircraft collision avoidance \citep{corso2021survey} and power grids \citep{owen2019importance}, since these problems all involve domain constraints. %

\paragraph{Main insights.} We conclude \cref{sec:experiments} with a summary of the main insights obtained throughout the experiments. 
\textbf{(I1)} On the tested targets, rejection sampling is a surprisingly effective alternative to ARITS and performs well for both VI and IS. 
\textbf{(I2)} \dis{} on the other hand does not reach the same performance as rejection and ARITS and the same holds for $\Delta\text{VI}$. While the safe component can help to mitigate the variance of \dis{}, effectively using \dis{} in practice will require future work on further reducing its variance.  
\textbf{(I3)}  When a density has prominent ``holes'' (such as the \emph{Ring} and \emph{Hollow} targets), we observe a clear benefit of using SMMs, both in terms of fit to the target and normalizing constant estimation.
\textbf{(I4)} Lastly, while SMMs show promising initial results, there are several open learning challenges, including finding robust initializations for VI with SMMs, and ways to effectively fit squared SMMs with a high number of components, despite parameter sharing across components. %

\section{DISCUSSION}
In this work, we laid the foundations to effectively use SMMs for IS and VI, overcoming the lack the latent variable interpretation of classical additive MMs.
To this end, we introduced and compared several estimators based on three key sampling routines---ARITS, rejection sampling and \dis---highlighting in theory and in practice the induced trade-offs in terms of accuracy of estimation, scalability, and statistical stability. 
Gaining a deeper theoretical understanding of the optimization challenges that negative components and the squaring operation bring will be crucial for further stabilizing approximate inference with SMMs.

\textbf{Future work.} We leave for future work the exploration of self-normalized importance sampling \citep{mcbook} and particle filtering (the latter can be viewed as IS with mixture proposals, \citep{li2016efficient, branchini2025foundations}) with SMMs. %
Moreover, we aim to connect our findings to the literature of tensor factorizations and networks, where negative parameters are commonly used \citep{loconte2024relationshiptensorfactorizationscircuits}. We further plan to investigate how to adapt the techniques we proposed for shallow SMMs to the general deep PCs  \citep{vergari2019tractable,choi2020probabilistic}. %

\newpage

\ackaccepted{NB acknowledges support from the ProbAI Hub. 
LZ and AV were supported by the ``UNREAL: Unified Reasoning Layer for Trustworthy ML'' project (EP/Y023838/1) selected by the ERC and funded by UKRI EPSRC.
LDS was supported by the Internal Funds KU Leuven (projects iBOF/21/075 and PDMT2/25/057).
He also acknowledges support from the Flemish Government (AI Research Program) and Una Europa.
NM acknowledges support from the CIFAR Learning in Machines and Brains Programme.
We are grateful to the \href{https://april-tools.github.io/}{april} lab for valuable feedback, in particular Lorenzo Loconte for useful discussion on the implementation of ARITS and the \texttt{cirkit} library, Adri\'{a}n
Javaloy for early helpful discussions around SMMs, and Leander Kurscheidt for providing the neuro-symbolic targets.
}

\paragraph{Contributions.}
NB suggested using the difference representation of SMMs for IS and, together with LZ, developed the theory of $\Delta$IS, including its formalization, proofs, and derivations. LZ led experiments with help from NB and LDS, who respectively suggested benchmarks and provided GPU-accelerated implementations of $\Delta$VI and other baselines. LZ also identified the instability of $\Delta$IS, proposed a safe component to address it, and suggested the stratified variant. NM and VE provided useful discussion and feedback on the manuscript. AV supervised all the stages of the project.

\bibliographystyle{plainnat}
\bibliography{bib.clean}

\section*{Checklist}

\begin{enumerate}
  \item For all models and algorithms presented, check if you include:
  \begin{enumerate}
    \item A clear description of the mathematical setting, assumptions, algorithm, and/or model. Yes, provided throughout the paper, primarily \cref{sec:smms} and \cref{sec:approximate_inference}, \cref{app:sampling_algs}, \cref{app:proofs}.
    \item An analysis of the properties and complexity (time, space, sample size) of any algorithm. Yes, see \cref{app:complexity}.
    \item (Optional) Anonymized source code, with specification of all dependencies, including external libraries. Yes, see \url{https://github.com/april-tools/delta-vi}.
  \end{enumerate}

  \item For any theoretical claim, check if you include:
  \begin{enumerate}
    \item Statements of the full set of assumptions of all theoretical results. Yes, see \cref{proposition:properties}, \cref{proposition:variance_rs}, and \cref{app:proofs}.
    \item Complete proofs of all theoretical results. Yes, see \cref{app:proofs}.
    \item Clear explanations of any assumptions. Yes, see \cref{proposition:properties}, \cref{proposition:variance_rs}, and \cref{app:proofs}.
  \end{enumerate}

  \item For all figures and tables that present empirical results, check if you include:
  \begin{enumerate}
    \item The code, data, and instructions needed to reproduce the main experimental results (either in the supplemental material or as a URL). Yes, see \url{https://github.com/april-tools/delta-vi}.
    \item All the training details (e.g., data splits, hyperparameters, how they were chosen). Yes, see \cref{app:exp_setup}, \cref{tab:hyper_grid}, \cref{tab:hyper_grid_logreg} and \cref{app:rq_4}.
    \item A clear definition of the specific measure or statistics and error bars (e.g., with respect to the random seed after running experiments multiple times). Yes, see \cref{sec:experiments}.
    \item A description of the computing infrastructure used. (e.g., type of GPUs, internal cluster, or cloud provider). Yes, see \cref{app:exp_setup}.
  \end{enumerate}

  \item If you are using existing assets (e.g., code, data, models) or curating/releasing new assets, check if you include:
  \begin{enumerate}
    \item Citations of the creator If your work uses existing assets. Yes, see \cref{app:exp_setup}.
    \item The license information of the assets, if applicable. Yes, see \cref{app:exp_setup}.
    \item New assets either in the supplemental material or as a URL, if applicable. Yes, see \url{https://github.com/april-tools/delta-vi}.
    \item Information about consent from data providers/curators. Not Applicable.
    \item Discussion of sensible content if applicable, e.g., personally identifiable information or offensive content. Not Applicable.
  \end{enumerate}

  \item If you used crowdsourcing or conducted research with human subjects, check if you include:
  \begin{enumerate}
    \item The full text of instructions given to participants and screenshots. Not Applicable.
    \item Descriptions of potential participant risks, with links to Institutional Review Board (IRB) approvals if applicable. Not Applicable.
    \item The estimated hourly wage paid to participants and the total amount spent on participant compensation. Not Applicable.
  \end{enumerate}

\end{enumerate}

\onecolumn
\appendix

\aistatstitle{Supplementary material for: How to Approximate Inference with Subtractive Mixture Models}

\begin{figure*}[ht!]
    \centering
    \raisebox{20pt}{\scalebox{6}{{(}}}\includegraphics[height=.15\textwidth, page=1]{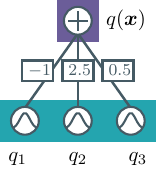}
    \raisebox{20pt}{\scalebox{6}{{)}}\raisebox{30pt}{\scalebox{1.5}{$^2$}}}
    \raisebox{30pt}{\scalebox{3}{=}}
    \includegraphics[height=.15\textwidth, page=2]{figures/smm.pdf}
    \raisebox{30pt}{\scalebox{3}{=}}
    \includegraphics[height=.15\textwidth, page=3]{figures/smm.pdf}
    \raisebox{30pt}{\scalebox{3}{$-$}}
    \includegraphics[height=.15\textwidth, page=4]{figures/smm.pdf}

    \caption{
    \textbf{A squared mixture can be split into its positive and negative parts} as illustrated via its representation as a computational graph, also called circuit \citep{choi2020probabilistic,loconte2024relationshiptensorfactorizationscircuits}.
    }
    \label{fig:smms-pcs}
\end{figure*}
\section{SAMPLING ALGORITHMS}
\label{app:sampling_algs}
In this section, we provide further details on the sampling algorithms discussed throughout the paper.
\Cref{alg:ancestral-sampling} provides the algorithm for ancestral mixture sampling. 
\Cref{alg:stratified} covers stratified sampling for additive mixture models. \Cref{alg:ARITS} gives detailed pseudocode for our ARITS implementation. We derive the variance of a MC estimator based on rejection in \Cref{app:rejection}. We discuss the complexity of the sampling strategies in \Cref{app:complexity}.

\subsection{Ancestral sampling}
\label{app:ancestral}

        \begin{algorithm}[H]
            \SetKwInOut{Input}{Input}
            \SetKwInOut{Output}{Output}
            \textbf{Input:} an additive MM $q$ \cref{eq:mm}, and 
            sample budget $S$\;
            \textbf{Output:} $S$ i.i.d. samples from $q$\;
            $\mathcal{X}\leftarrow\{\}$\; 
            \For{$s \in \{1, ..., S\}$}{
                $k\sim\mathsf{Categorical}(\boldsymbol{\alpha})$\;
                $\boldsymbol{x}^{(s)}\sim q_k$\;
                $\mathcal{X}\leftarrow\mathcal{X}\cup\{\boldsymbol{x}^{(s)}\}$\;
            }  
            \Return{$\mathcal{X}$}
            \caption{{\small ancestralSampling($q, S$)}}\label{alg:ancestral-sampling}
        \end{algorithm}

\subsection{Stratified sampling}
\label{app:stratified}

        \begin{algorithm}[H]
            \SetKwInOut{Input}{Input}
            \SetKwInOut{Output}{Output}
            \textbf{Input:} an additive MM $q$ \cref{eq:mm}, and 
            sample budget $S$\;
            \textbf{Output:} up to $S$ stratified samples from $q$\;
            $\mathcal{X}\leftarrow\{\}$\; 
            $S_1,\ldots,S_K\leftarrow\lfloor\alpha_1 S\rfloor,\ldots,\lfloor\alpha_K S\rfloor$\;
            \For{$k \in \{1, ..., K\}$}{
                $\mathcal{X}^{(k)}\leftarrow\{\}$\;
                \For{$s \in \{1, ..., S_k\}$}{
                $\boldsymbol{x}^{(k, s)}\sim q_k$\;
                $\mathcal{X}^{(k)}\leftarrow\mathcal{X}^{(k)}\cup\{\boldsymbol{x}^{(k,s)}\}$\;
                }
                $\mathcal{X}\leftarrow\mathcal{X}\cup\mathcal{X}^{(k)}$\;
            }  
            \Return{$\mathcal{X}$}
            \caption{{\small stratifiedSampling($q, S$)}}\label{alg:stratified}
        \end{algorithm}

\subsection{Auto-regressive Inverse Transform Sampling (ARITS)}\label{app:arits}
For performing ARITS in practice, we use a binary search for numerically inverting the conditional CDF. %
\cref{alg:ARITS} provides the detailed algorithm. In our experiments, the start and end points of the binary search are set to $L=-100$ and $B=100$ respectively for RQ 1) and to $L=-50$ and $B=50$ for learning\footnote{To ensure the validity of the algorithm, we always check whether these bounds result in a (conditional) CDF of $0$ and $1$ respectively.}.
The search is stopped $|L-B| > \epsilon$, for $\epsilon=10^{-6}$. 
Note that computing the conditional CDF in the algorithm is tractable for SMMs as they are smooth and decomposable circuits \citep{vergari2021compositional}.

\begin{algorithm}[ht!]
    \SetKwInOut{Input}{Input}
    \SetKwInOut{Output}{Output}
    \Input{a SMM $q$ (\cref{eq:smm}),\newline 
    sample budget $S$,\newline
    initial upper bound $B$, \newline
    initial lower bound $L$,\newline
    tolerance $\epsilon$
    }
    \Output{$S$ i.i.d. samples from $q$}
    \For{$s \in \{1, \ldots, S$\}}{
    $x^{(s)} \leftarrow \{\}$\;
    \For{$d \in \{1, ..., D\}$}{
        $u \sim \text{Unif}(0,1)$\;
        \tcc{Pre-compute the evidence of previously sampled dimensions}
        \If{$d > 1$}{
        $e \gets q(x_1^{(s)}, ..., x^{(s)}_{d-1})$\;
        }
        \Else{
            $e \gets 1$
        }
        \tcc{Perform binary search to numerically invert conditional CDF}
        \While{$|L - B| > \epsilon$}{
            $M \gets L + (B-L)/2$\;
            \tcc{Compute conditional CDF at midpoint}
            $c \gets q(x_d^{(s)} \leq M, \; x^{(s)}_1, ..., x^{(s)}_{d-1})/e$\;
            \If{$c > u$}{
                   $B \gets M$
                }
            \Else{
                    $L \gets M$
            }
            
        }
        \tcc{Recompute midpoint and set $x_d^{(s)}$}
        $x^{(s)}_d \leftarrow L + (B-L)/2$
    }
    $\vx^{(s)} \leftarrow (x_1^{(s)}, \ldots, x_D^{(s)})$\;
    $\mathcal{X}\leftarrow\mathcal{X}\cup\{\vx^{(s)}\}$\;
    }
    \Return{$\mathcal{X}$}
    \caption{binarySearchArits($q$, $S$, $L$, $B$, $\epsilon$)}\label{alg:ARITS}
\end{algorithm}

\subsection{Rejection Sampling for SMMs}\label{app:rejection}

\paragraph{Derivation of rejection sampling variance.}

\begin{definition} %
    We define a zero-truncated binomial distribution $\mathsf{TrBin}(S;a)$ on $\{1,\dots,S\}$ with probability mass function (PMF)
    $$
    \mathbb{P}[K=k] = \frac{\binom{S}{k} a^{k}(1-a)^{S-k} }{1 - (1-a)^S} , ~ k ~\in \{1,\dots,S\} , 
    $$
    and parameters $a \in [0,1]$ and $S \in \mathbb{N}$.
\end{definition}

Given the above definition, it is straightforward to derive the variance of the rejection estimator by applying the law of total variance 
\begin{align}
\begin{split}
\mathbb{V}_{\substack{\vx \sim q_{\text{SMM}}\\ K \sim \mathsf{TrBin}(S;a)}} \left [ \widehat{I}_{\text{RS}} \right ] &= \mathbb{E}_{K \sim \mathsf{TrBin}}\left [ \mathbb{V}_{\vx \sim q_{\text{SMM}}}[\widehat{I}_{\text{RS}} | K] \right ]  \\
&+ \mathbb{V}_{K\sim\mathsf{TrBin}} \left [ \mathbb{E}_{\vx \sim q_{\text{SMM}}}[\widehat{I}_{\text{RS}} | K] \right ]
\end{split}
\end{align}
Since $\mathbb{E}_{\vx \sim q_{\text{SMM}}}[\widehat{I}_{\text{RS}} | K] = I$\footnote{Conditioning on $K$, the number of accepted samples is equivalent to conditioning on the acceptance pattern, i.e., binary r.v.s. which denote exactly which sample was accepted.} for any $K$ the second term is zero, so expanding the first term (using that accepted samples are i.i.d.), 
$$
\mathbb{V}[\widehat{I}_{\text{RS}}] = \mathbb{V}_{\vx \sim q_{\text{SMM}}}[h(\vx)] \cdot  \mathbb{E}\left [\frac{1}{K} \right ]
$$

\begin{figure}[t]
    \centering
    \includegraphics[width=0.77\linewidth]{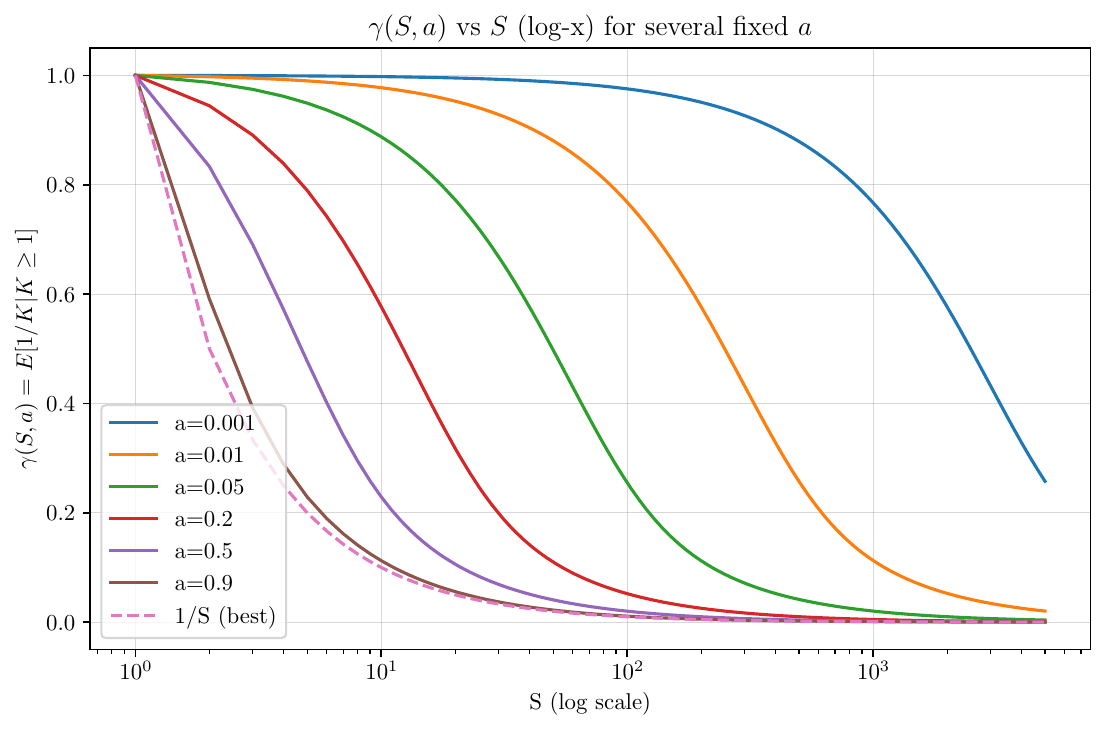}
    \caption{$\gamma(a,S)$ versus $S$ (log-scale $x$-axis) for several values of $a$, the acceptance probability of rejection. The best convergence rate is $1/S$, which is almost achieved when $a$ is close to $1$.}
    \label{fig:rejection_inflation}
\end{figure}

In the following, we set $\gamma(S, a) := \mathbb{E}[1/K]$ under the defined truncated binomial 
\begin{equation}\label{eq:inflation_factor_deriv}
\begin{split}
\gamma(S,a) := \mathbb{E}[1/K ] = \frac{\sum_{k=1}^{S}\frac{1}{k} \binom{S}{k} a^{k}(1-a)^{S-k} }{1 - (1-a)^S}.
\end{split}
\end{equation}
The dependence on $a$ is intuitive: as $a$ goes from $0$ to $1$, \Ref{eq:inflation_factor_deriv} goes from $1$ to $1/S$, i.e., impacting the MC convergence rate. See \Cref{fig:rejection_inflation} for an illustration. Note that advanced schemes recycling rejected samples would be possible \citep{casella1998post}.

\subsection{COMPLEXITY ANALYSIS}
\label{app:complexity}

To derive the complexity of our sampling algorithm, we will consider classical additive MMs and SMMs to be represented as \textit{\textbf{probabilistic circuits}} (PCs) \citep{choi2020probabilistic,vergari2021compositional}, computational graphs involving three types of computational units: sum, product and input units.
Sum units compute linear combinations of their inputs and are used to represent the sum operation in MMs (\cref{eq:mm}) and SMMs (\cref{eq:smm}).
Input distributions are custom neurons that encode parametric PDFs or mass functions.
Product units encode local factorizations and appear in our models when we consider input distributions that 
factorize into independent marginals.
\cref{fig:smms-pcs} illustrates one squared SMM as a sum unit over many multivariate (and possibly unnormalized) Gaussian PDF units.
Alternatively, we could have replaced each input unit in \cref{fig:smms-pcs} with a product unit over $D$ input units each encoding a (possibly unnormalized) univariate Gaussian PDF.
Under this construction, we discuss complexity next.

\textbf{Ancestral sampling (\cref{alg:ancestral-sampling}).} For an additive MM, we first sample one component out of $K$ from a Categorical, which can be done in $\mathcal{O}(K)$, assuming unitary cost to sample from uniform distributions.
Then, we can sample from the relative isotropic Gaussian component, i.e., sampling from a univariate Gaussian $D$ times.
Assuming the cost of sampling from a univariate Gaussian is unitary, one has to perform $\mathcal{O}(S(K + D))$ operations.
If we adopt \textit{stratified sampling} (\cref{alg:stratified}) for an additive MM instead, we can first precompute how many samples each mixture component will yield by multiplying its corresponding mixture coefficient with $S$.
We then visit the circuit in a feedforward way, sampling from all $KD$ input units at once and propagate the partial samples by concatenating them along columns (dimensions $D$) when encountering a product unit and 
along rows (samples $S$) when encountering a sum unit.
Again assuming the cost of sampling to be unitary, we have to compute $\mathcal{O}(S(K + KD))=\mathcal{O}(SKD)$ operations, which corresponds to evaluating the full circuit $S$ times.

\textbf{ARITS (\cref{alg:arits}).} To generate a single sample, we have to compute the inverse of the marginal CDF $D$ times. To do that, we first have to marginalize out the circuit, which can be done exactly in time linear in the circuit size, i.e. $\mathcal{O}(KD)$.
To then invert the CDF, we have to evaluate the marginal circuit in a binary search process, retrieving a sample up to precision $\epsilon$, which has a cost of $\mathsf{b}(\epsilon)$.
As such, the overall procedure should take $\mathcal{O}(SKD^{2})\mathsf{b}(\epsilon)$.

\textbf{Rejection sampling (\cref{alg:rejection}).} The worst-case complexity is the same as ancestral sampling, as we have to sample $S$ times from the positive part of an SMM. While for the analysis we assumed that we have at least one acceptance, in practice it is not guaranteed to have at least one acceptance and therefore one should consider the runtime of rejection sampling more generally as a random variable (with not necessarily finite variance).

\section{PROOFS AND DERIVATIONS}\label{app:proofs}
\subsection{Properties of \dis}\label{proof:theorem_1}
We now prove the properties of \dis{} stated in \cref{proposition:properties}. %

\subsubsection{Consistency of $\Delta\text{IS}$}\label{app:consistency}
This property amounts to showing consistency separately for the two expectations. The conditions are slightly different than for UIS because the IS weight is different (and we have a combination of two estimators). 
Let the following assumptions be satisfied (beyond the usual $q \gg p$, $\int |f| p < \infty $)
\begin{itemize}
    \item \textbf{(A1)} $q > 0$ $q_{+}$-a.e. and $q > 0$ $q_{-}$-a.e. (that is, $q \gg q_{+}$ and $q \gg q_{-}$)
    \item  \textbf{(A2)}  $
\mathbb{E}_{x\sim q_{+}}\left[ | f(\vx) |\frac{p(\vx)}{q(\vx)}  \right] < \infty\text{ and } \mathbb{E}_{x\sim q_{-}}\left[ | f(\vx) |\frac{p(\vx)}{q(\vx)} \right] < \infty
$.
\end{itemize}
The above are sufficient for an individual strong law of large numbers (SLLN) holding for the two estimators within $\Delta\text{IS}$ as both $S_{+} \rightarrow \infty$ and $S_{-} \rightarrow \infty$. Finally, since the mapping $(x,y) \rightarrow (Z_{+}/Z)x - (Z_{-}/Z)y$ is continuous, by the continuous mapping theorem we have the desired SLLN for $\Delta\text{IS}$, 
$$
\mathbb{P}_{\substack{\vx_+ \sim q_+\\\vx_- \sim q_-}} \left[ \lim_{\substack{S_{+}, S_{-}\rightarrow \infty}}\widehat{I}_{\Delta\text{IS}} = I \right] = 1. 
$$

\subsubsection{Unbiasedness of $\Delta\text{IS}$ for UIS}\label{app:unbiasedness}
We show next show that $\Delta\text{IS}$ is unbiased.

Recall that our estimator for $I = \int f(\vx)p(\vx)d\vx$ is given as
\begin{equation*}
\widehat{I}_{\Delta\text{IS}} = \frac{Z_{+}}{Z}\frac{1}{S_{+}} \sum\nolimits_{s=1}^{S_{+}}  f(\vx_{+}^{(s)}) w(\vx_{+}^{(s)}) - \frac{Z_{-}}{Z}\frac{1}{S_{-}} \sum\nolimits_{s=1}^{S_{-}} f(\vx_{-}^{(s)})w(\vx_{-}^{(s)}), \ \text{where}
\begin{array}{l}
\vx_{+}^{(s)}\sim q_{+}(\vx_{+}) \\[2pt]
\vx_{-}^{(s)} \sim q_{-}(\vx_{-})
\end{array}.
\end{equation*}
Note that $q(\boldsymbol{x})=\frac{1}{Z}(Z_+q_+(\boldsymbol{x}) - Z_-q_-(\boldsymbol{x}))$ and $\{ \vx_{+}^{(s)}\}_{s=1}^{S_+} \overset{\text{i.i.d.}}{\sim} q_{+}$ and $\{ \vx_{-}^{(s)}\}^{S_-}_{s=1} \overset{\text{i.i.d.}}{\sim} q_{-}$. %
Assuming that $\int |f(\vx)| p(\vx) dx < \infty$ and $q(\vx) \neq 0$ almost-everywhere in the support of $q_{+}$ and $q_{-}$, we have
\begin{align}
\mathbb{E}_{\substack{\vx_+ \sim q_+\\\vx_- \sim q_-}}[\widehat{I}_{\Delta\text{IS}}] 
= & \mathbb{E}_{\substack{\vx_+ \sim q_+\\\vx_- \sim q_-}}\left[\frac{Z_+}{Z} \frac{1}{S_{+}} \sum_{s=1}^{S_+}  f(\boldsymbol{x}_{+}^{(s)}) \frac{p(\boldsymbol{x}_{+}^{(s)})}{q(\boldsymbol{x}_{+}^{(s)})} - \frac{Z_-}{Z}\frac{1}{S_{-}} \sum_{s=1}^{S_-} f(\boldsymbol{x}_{-}^{(s)}) \frac{p(\boldsymbol{x}_{-}^{(s)})}{q(\boldsymbol{x}_{-}^{(s)})}\right]\\
= & \frac{Z_+}{Z} \frac{1}{S_{+}}\sum_{s=1}^{S_+} \mathbb{E}_{q_+}\left[f(\boldsymbol{x}_{+}^{(s)}) \frac{p(\boldsymbol{x}_{+}^{(s)})}{q(\boldsymbol{x}_{+}^{(s)})}\right] - \frac{Z_-}{Z} \frac{1}{S_{-}}\sum_{s=1}^{S_-} \mathbb{E}_{q_-}\left[f(\boldsymbol{x}_{-}^{(s)}) \frac{p(\boldsymbol{x}_{-}^{(s)})}{q(\boldsymbol{x}_{-}^{(s)})}\right]\\ %
= & \frac{Z_+}{Z}\mathbb{E}_{q_+}\left[f(\boldsymbol{x}_{+}) \frac{p(\boldsymbol{x}_{+})}{q(\boldsymbol{x}_{+})}\right] - \frac{Z_-}{Z} \mathbb{E}_{q_-}\Big[f(\boldsymbol{x}_{-}) \frac{p(\boldsymbol{x}_{-})}{q(\boldsymbol{x}_{-})}\Big]\\ %
= & \frac{Z_+}{Z}\int f(\boldsymbol{x}) \frac{p(\boldsymbol{x})}{q(\boldsymbol{x})}q_+(\vx)d\boldsymbol{x} - \frac{Z_-}{Z} \int f(\boldsymbol{x}) \frac{p(\boldsymbol{x})}{q(\boldsymbol{x})}q_-(\boldsymbol{x})d\boldsymbol{x} \\
= & \int f(\boldsymbol{x}) \frac{p(\boldsymbol{x})}{q(\boldsymbol{x})}\frac{1}{Z}(Z_+q_+(\boldsymbol{x}) - Z_-q_-(\boldsymbol{x}))d\boldsymbol{x} \\
= & \int f(\boldsymbol{x}) \frac{p(\boldsymbol{x})}{q(\boldsymbol{x})}q(\boldsymbol{x})d\boldsymbol{x} %
= \int f(\boldsymbol{x}) p(\boldsymbol{x})d\boldsymbol{x} = I. 
\end{align}

\subsubsection{Variance of \dis}\label{app:variance}
Using the independence of $\vx_{+}$ and $\vx_{-}$ as well as the fact that $\{ \vx_{+}^{(s)}\}_{s=1}^{S_+} \overset{\text{i.i.d.}}{\sim} q_{+}$ and $\{ \vx_{-}^{(s)}\}^{S_-}_{s=1} \overset{\text{i.i.d}}{\sim} q_{-}$, we arrive at the following variance expression, where $w(\vx)=\frac{p(\vx)}{q(\vx)}$,
\begin{align*}
    \mathbb{V}_{\substack{\vx_+ \sim q_+\\\vx_- \sim q_-}}[\widehat{I}_{\Delta\text{IS}}] &= \mathbb{V}_{q_{+}}\left[ \frac{Z_+}{Z}\frac{1}{S_{+}} \sum\nolimits_{s=1}^{S_{+}}  f(\vx_{+}^{(s)}) w(\vx_{+}^{(s)}) \right ] + \mathbb{V}_{q_{-}}\left [\frac{Z_-}{Z}\frac{1}{S_{-}} \sum\nolimits_{s=1}^{S_{-}} f(\vx_{-}^{(s)})w(\vx_{-}^{(s)}) \right ] \\
    &= \frac{Z_+^2}{Z^2} \frac{1}{S_{+}} \left ( \mathbb{E}_{q_{+}}\left [ (f(\vx_{+}) w(\vx_{+}))^2 \right ] -  \left (\mathbb{E}_{q_{+}}\left [ f(\vx_{+}) w(\vx_{+}) \right ] \right )^2 \right ) \\
    & + \frac{Z_-^2}{Z^2} \frac{1}{S_{-}} \left ( \mathbb{E}_{q_{-}}\left [ (f(\vx_{-}) w(\vx_{-}))^2 \right ] -  \left (\mathbb{E}_{q_{-}}\left [ f(\vx_{-}) w(\vx_{-}) \right ] \right )^2 \right ).
\end{align*}

\subsubsection{Optimal UIS proposal for \dis}\label{app:variance}
\paragraph{$f\geq 0$ or $f\leq 0$.} It is easy to see that when $f(\vx)>0$ a.e. (or $f(\vx) \leq 0)$ a.e., the optimal proposal for \dis{} is $q^{\bigstar}(\vx)=\frac{f(\vx)p(\vx)}{I}$, where $I=\int f(\vx)p(\vx)d\vx$ is the integral of interest. 
Plugging $q^{\bigstar}$ into \cref{eq:dis} gives
\begin{align*}
\frac{Z_+}{Z}\frac{1}{S_{+}} \sum\limits_{s=1}^{S_{+}} \frac{f(\vx^{(s)})p(\vx^{(s)})}{f(\vx^{(s)})p(\vx^{(s)})/I} - \frac{Z_-}{Z}\frac{1}{S_{-}} \sum\limits_{s=1}^{S_{-}} \frac{f(\vx^{(s)})p(\vx^{(s)})}{f(\vx^{(s)})p(\vx^{(s)})/I} 
= \left(\frac{Z_+}{Z}-\frac{Z_-}{Z}\right)I = I.
\end{align*}
Since $I$ is a constant, we have $\mathbb{V}_{\substack{\vx_+ \sim q_+\\\vx_- \sim q_-}} [\widehat{I}_{\Delta\text{IS}}]=0$ for $q=q^{\bigstar}$.

\paragraph{General $f$.} We now prove the optimal proposal for general $f$, which might take both positive and negative values. This proposal is given as $q^{\bigstar}(\vx) = \frac{|f(\vx)|p(\vx)}{\int |f(\vx)|p(\vx)d\vx}$. The proof has two main steps (1) explicitly relating the variance of \dis{} to the standard UIS variance and (2) expressing the optimal proposal $q^{\bigstar}$ as $q_+$.

Note first that
\begin{equation}\label{eq:useful_identity_proposal}
    q = Z^{-1}(Z_{+} q_{+} - Z_{-}q_{-}) \Rightarrow q_{+} = Z/Z_{+} q + Z_{-} /Z_{+} q_{-}.
\end{equation}

We now rewrite the variance of \dis{} in terms of the variance of UIS with the same proposal $q$. Let $w(\vx) := \frac{f(\vx)p(\vx)}{q(\vx)}$.
We know from \cref{app:variance} that
\begin{align}\label{eq:var-deltais-ours}
\mathbb{V}_{\substack{\vx_+ \sim q_+\\\vx_- \sim q_-}}[\widehat{I}_{\Delta\text{IS}}] = \frac{1}{S_+}\frac{Z_+^2}{Z^2}\mathbb{V}_{q_+}{\Big[}w(\vx){\Big]} + \frac{1}{S_-}\frac{Z_-^2}{Z^2}\mathbb{V}_{q_-}{\Big[}w(\vx){\Big]}.
\end{align}

Now, by rewriting the first term $\mathbb{V}_{q_+}{\Big[}w(\vx){\Big]} $ as a function of $q \text{ and } q_{-}$ we will introduce the term $\mathbb{V}_{q}[w(\vx)]$ (variance of UIS) and establish an inequality between $\mathbb{V}_{q}[w(\vx)]$ and $\mathbb{V}_{\substack{\vx_+ \sim q_+\\\vx_- \sim q_-}}[\widehat{I}_{\Delta\text{IS}}]$.
Using 
\begin{align}
    \mathbb{V}_{q_{+}}[w(\vx)] &= \mathbb{E}_{q_{+}}[w(\vx)^2] - \mathbb{E}_{q_{+}}[w(\vx)]^2 
\end{align}

and \cref{eq:useful_identity_proposal}, 
letting $a:= Z/Z_{+}$, $b:= Z_{-}/Z_{+}$, we have (abbreviating $w(\vx)$ as $w$)
\begin{align}\label{eq:intermediate_1}
\mathbb{V}_{q_{{+}}}[w] = \overbrace{a \mathbb{E}_{q}[w^2] + b \mathbb{E}_{q_{-}}[w^2]}^{ = \mathbb{E}_{q_{+}}[w(\vx)^2]} - (a \mathbb{E}_q[w] + b \mathbb{E}_{q_{-}}[w])^2
\end{align}

We rewrite \cref{eq:intermediate_1} by expressing the second moments in terms of the corresponding variances, resulting in
\begin{align}
    \mathbb{V}_{q_{{+}}}[w] = a \mathbb{V}_q[w] + b \mathbb{V}_{q_{-}}[w] + \left[a \mathbb{E}_{q}^2[w] + b \mathbb{E}_{q_{-}}^2[w] - (a \mathbb{E}_{q}[w] + b \mathbb{E}_{q_{-}}[w])^2 \right].
\end{align}

The bracket in the above can be simplified. Expanding the square in the bracket:
$$
\begin{aligned}
& a \mathbb{E}_q^2[w]+b \mathbb{E}_{q_{-}}^2[w]-\left(a \mathbb{E}_q[w]+b \mathbb{E}_{q_{-}}[w]\right)^2 \\
& \quad=a \mathbb{E}_q^2[w]+b \mathbb{E}_{q_{-}}^2[w]-a^2 \mathbb{E}_q^2[w]-b^2 \mathbb{E}_{q_{-}}^2[w]-2 a b \mathbb{E}_q[w] \mathbb{E}_{q_{-}}[w] \\
& \quad=a(1-a) \mathbb{E}_q^2[w]+b(1-b) \mathbb{E}_{q_{-}}^2[w]-2 a b \mathbb{E}_q[w] \mathbb{E}_{q_{-}}[w].
\end{aligned}
$$

Since $a+b=1$, we have $1-a=b$ and $1-b=a$, and therefore
$$
\begin{aligned}
a(1-a) \mathbb{E}_q^2[w]+b(1-b) \mathbb{E}_{q_{-}}^2[w]-2 a b \mathbb{E}_q[w] \mathbb{E}_{q_{-}}[w]
& =a b \mathbb{E}_q^2[w]+a b \mathbb{E}_{q-}^2[w]-2 a b \mathbb{E}_q[w] \mathbb{E}_{q-}[w] \\
& =a b\left(\mathbb{E}_q[w]-\mathbb{E}_{q-}[w]\right)^2.
\end{aligned}
$$
We can finally rewrite \cref{eq:intermediate_1} as
\begin{align}
    \mathbb{V}_{q_{{+}}}[w] = a \mathbb{V}_q[w] + b \mathbb{V}_{q_{-}}[w] + ab (\mathbb{E}_{q}[w] - \mathbb{E}_{q_{-}}[w])^2 .
\end{align}

Noting that $b \mathbb{V}_{q_{-}}[w] \geq 0$ and $ab(\mathbb{E}_{q}[w] - \mathbb{E}_{q_{-}}[w])^2 \geq 0$, it follows that
\begin{align}\label{eq:final_useful_ineq}
     \mathbb{V}_{q_{{+}}}[w] \geq a \mathbb{V}_{q}[w] = \frac{Z}{Z_{+}} \mathbb{V}_{q}[w].
\end{align}

Using this identity and going back to \cref{eq:var-deltais-ours}, we obtain
\begin{align}
   \mathbb{V}_{\substack{\vx_+ \sim q_+\\\vx_- \sim q_-}}[\widehat{I}_{\Delta\text{IS}}]&=\frac{1}{S_+}\frac{Z_+^2}{Z^2}\mathbb{V}_{q_+}{\Big[}w(\vx){\Big]} + \frac{1}{S_-}\frac{Z_-^2}{Z^2}\mathbb{V}_{q_-}{\Big[}w(\vx){\Big]} \\
    &\geq \frac{1}{S_+}\frac{Z_+^2}{Z^2}\mathbb{V}_{q}{\Big[}w(\vx){\Big]}  \\
    &\geq \frac{1}{S_+} \cdot \frac{Z}{Z_{+}} \cdot \frac{Z_+^2}{Z^2}\mathbb{V}_{q}{\Big[}w(\vx){\Big]} = \frac{1}{S_+} \cdot \frac{Z}{Z_{+}} \mathbb{V}_{q}{\Big[}w(\vx){\Big]},
\end{align}
which relates the variance of \dis{} and $\mathbb{V}_{q}{\Big[}w(\vx){\Big]}$. To complete the first part of the proof, we show 
$$
\frac{Z_{+}}{Z} \frac{1}{S_{+}} \geq \frac{1}{S} .
$$
which implies $\frac{1}{S_+} \cdot \frac{Z}{Z_{+}} \mathbb{V}_{q}{\Big[}w(\vx){\Big]} \geq 1/S \mathbb{V}_{q}{\Big[}w(\vx){\Big]} = \mathbb{V}_{x \sim q}[\widehat{I}_{\text{UIS}}]$. Using that
$$
\frac{Z_{+}}{Z S_{+}}-\frac{1}{S}=\frac{Z_{+} S-Z S_{+}}{Z S_{+} S},
$$
and further rewriting the numerator using $S=S_{+}+S_{-}$and $Z=Z_{+}-Z_{-}$, we obtain
$$
\begin{aligned}
Z_{+} S-Z S_{+} & =Z_{+}\left(S_{+}+S_{-}\right)-\left(Z_{+}-Z_{-}\right) S_{+} \\
& =Z_{+} S_{-}+Z_{-} S_{+}  \geq 0.
\end{aligned}
$$
We have hence shown $\mathbb{V}_{\substack{\vx_+ \sim q_+\\\vx_- \sim q_-}}[\widehat{I}_{\Delta\text{IS}}] \geq \mathbb{V}_{x \sim q}[\widehat{I}_{\text{UIS}}]$. 

To prove that the optimal proposal $q^{\bigstar} \propto |f|p$ is the same for $\widehat{I}_{\text{$\Delta\text{IS}$}}$ and $\widehat{I}_{\text{UIS}}$, we will show $\mathbb{V}_{\substack{\vx_+ \sim q_+\\\vx_- \sim q_-}}[\widehat{I}_{\Delta\text{IS}}]$ using $q = q^{\bigstar}$ is no greater than using any other proposal $q$. In particular, there is a choice of $q(\vx) = (Z_{+}/Z) ~q_{+}(\vx) - (Z_{-}/Z) ~ q_{-}(\vx)$ that obtains minimum variance: Set $q = q_{+}$ and $S=S_{+}$, which implies $Z_{-}=0$ (no negative component). Then, the $\Delta$IS estimator is equivalent to UIS. We can now simply set $q = q_{+} = q^{\bigstar}$ in $\Delta$IS. Then, for any $q$ the below holds 
\begin{align}
    \underbrace{\mathbb{V}_{q^{\bigstar}}[\widehat{I}_{\Delta\text{IS}}]}_{\text{choosing}~ S=S_{+}} = \mathbb{V}_{q^{\bigstar}}[\widehat{I}_{\text{UIS}}] \leq \mathbb{V}_{q}[\widehat{I}_{\text{UIS}}] \leq \mathbb{V}_{q}[\widehat{I}_{\Delta\text{IS}}] \quad \blacksquare . 
\end{align}

\subsection{Derivation of $\Delta\text{VI}$}\label{app:delta_vi}
Let $q_{\theta}$ denote an SMM proposal with $K$ components and corresponding learnable parameters $\theta = (\theta_1, \ldots,\theta_K)$. 
The learnable parameters include both the mixture weights and the component parameters.
Applying \Cref{eq:dex}, we decompose the RKL objective:
\begin{align}
\mathbb{E}_{q_{\theta}}{\Big[}\log\Big(\frac{q_{\theta}(\vx)}{\widetilde{p}(\vx)}\Big){\Big]} 
&= \frac{Z_+}{Z}\mathbb{E}_{q_{+}}{\Big[}\log\Big(\frac{q_{\theta}(\vx)}{\widetilde{p}(\vx)}\Big){\Big]} 
- \frac{Z_-}{Z} \mathbb{E}_{q_{-}}{\Big[}\log\Big(\frac{q_{\theta}(\vx)}{\widetilde{p}(\vx)}\Big){\Big]} \\
&= \frac{Z_+}{Z}\int{\Big[}\log\Big(\frac{q_{\theta}(\vx)}{\widetilde{p}(\vx)}\Big){\Big]}q_+(\vx)d\vx 
- \frac{Z_-}{Z} \int {\Big[}\log\Big(\frac{q_{\theta}(\vx)}{\widetilde{p}(\vx)}\Big){\Big]}q_-(\vx) d\vx.
\end{align}
We then apply stratification with respect to the underlying components, as described in \citet{morningstar2021automatic}, within $q_+$ and $q_-$ respectively. 
For the first term, we plug in $q_+(\vx) = \frac{1}{Z_+}\sum_{k \in \mathcal{K_+}} \alpha_k~ \widetilde{q}_k(\vx)$, where $\mathcal{K_+}$ denotes the set of indices corresponding to positively weighted components.
We denote the normalizing constant of $\widetilde{q}_k$ as by $Z_k$, i.e., $Z_k = \int \widetilde{q}_k(\vx) d\vx$.

We obtain the following for the first part 
\begin{align}
\frac{Z_+}{Z}\int{\Big[}\log\Big(\frac{q_{\theta}(\vx)}{\widetilde{p}(\vx)}\Big){\Big]}q_+(\vx)d\vx 
&= \frac{Z_+}{Z}\int {\Big[}\log\Big(\frac{q_{\theta}(\vx)}{\widetilde{p}(\vx)}\Big){\Big]}\frac{1}{Z_+}\sum_{k=1} \alpha_k~ \widetilde{q}_k(\vx)d\vx\\
&=\frac{1}{Z}\int {\Big[}\log\Big(\frac{q_{\theta}(\vx)}{\widetilde{p}(\vx)}\Big){\Big]}\sum_{k \in \mathcal{K_+}} \alpha_kZ_k~ \frac{\widetilde{q}_k(\vx)}{Z_k} d\vx\\
&=\frac{1}{Z}\int {\Big[}\log\Big(\frac{q_{\theta}(\vx)}{\widetilde{p}(\vx)}\Big){\Big]}\sum_{k \in \mathcal{K_+}} \alpha_k Z_k~ q_k(\vx) d\vx\\
&=\sum_{k \in \mathcal{K_+}}\frac{\alpha_kZ_k}{Z}\int  {\Big[}\log\Big(\frac{q_{\theta}(\vx)}{\widetilde{p}(\vx)}\Big){\Big]}q_k(\vx)d\vx\\
&=\sum_{k \in \mathcal{K_+}}\frac{\alpha_kZ_k}{Z}\mathbb{E}_{q_k}  {\Big[}\log\Big(\frac{q_{\theta}(\vx)}{\widetilde{p}(\vx)}\Big){\Big]}.
\end{align}

Analogously, we can rewrite the second expectation w.r.t. $q_-(\vx) = \frac{1}{Z_-}\sum_{j \in \mathcal{K_-}} |\alpha_{j}|~ \widetilde{q}_j(\vx)$, resulting in 
\begin{align*}
\frac{Z_-}{Z} \int {\Big[}\log\Big(\frac{q_{\theta}(\vx)}{\widetilde{p}(\vx)}\Big){\Big]}q_-(\vx) d\vx = \sum_{j \in \mathcal{K_-}}\frac{|\alpha_j|Z_j}{Z}\mathbb{E}_{q_j}  {\Big[}\log\Big(\frac{q_{\text{SMM}}(\vx)}{\widetilde{p}(\vx)}\Big){\Big]}.
\end{align*}

Putting the two terms back together, we arrive at
\begin{align}
\mathbb{E}_{q_{\theta}}{\Big[}\log\Big(\frac{q_{\theta}(\vx)}{\widetilde{p}(\vx)}\Big){\Big]} 
&= \sum_{k \in \mathcal{K_+}}\frac{\alpha_kZ_k}{Z}\mathbb{E}_{q_k}  {\Big[}\log\Big(\frac{q_{\theta}(\vx)}{\widetilde{p}(\vx)}\Big){\Big]} -
\sum_{j \in \mathcal{K_-}}\frac{|\alpha_j|Z_j}{Z}\mathbb{E}_{q_j}  {\Big[}\log\Big(\frac{q_{\theta}(\vx)}{\widetilde{p}(\vx)}\Big){\Big]}\\
&=\sum_{k \in \mathcal{K_+}}\frac{\alpha_kZ_k}{Z}\mathbb{E}_{q_k}  {\Big[}\log\Big(\frac{q_{\theta}(\vx)}{\widetilde{p}(\vx)}\Big){\Big]} + \sum_{j \in \mathcal{K_-}}\frac{\alpha_jZ_j}{Z}\mathbb{E}_{q_j}  {\Big[}\log(\frac{q_{\theta}(\vx)}{\widetilde{p}(\vx)}){\Big]}\\
&=\sum_{k=1}^{K}\frac{\alpha_kZ_k}{Z}\mathbb{E}_{q_k}  {\Big[}\log\Big(\frac{q_{\theta}(\vx)}{\widetilde{p}(\vx)}\Big){\Big]},
\end{align}
which is our $\Delta\text{VI}$ objective. The last few steps use that for $j \in \mathcal{K_-}$, the associated coefficients $\alpha_j$ were negative before taking the absolute value by construction, and $\mathcal{K}_+ \cup \mathcal{K}_-$ results in the full set of components.

\paragraph{Estimation of the $\Delta\text{IS}$ gradient.} We have shown that we can safely rewrite the initial objective in terms of a decomposition into individual components, despite negative coefficients, namely
\begin{align*}
\mathbb{E}_{q_{\theta}}{\Big[}\log\Big(\frac{q_{\theta}(\vx)}{\widetilde{p}(\vx)}\Big){\Big]} 
&= \sum_{k=1}^{K}\frac{\alpha_kZ_k}{Z}\mathbb{E}_{q_k}  {\Big[}\log\Big(\frac{q_{\theta}(\vx)}{\widetilde{p}(\vx)}\Big){\Big]}.
\end{align*}
We assume that the components $q_k$ are \emph{reparameterizable} with respect to some parameter-independent reference distribution $q_0$, i.e., $\vx = h_{\theta_k}(\vz),~ \vz \sim q_0$, for some mapping $h_{\theta_k}: \mathbb{R}^D \rightarrow \mathbb{R}^D$.
Rewriting the above in terms of $\vz$, we obtain
\begin{align*}
\mathbb{E}_{q_{\theta}}{\Big[}\log\Big(\frac{q_{\theta}(\vx)}{\widetilde{p}(\vx)}\Big){\Big]} 
&= \sum_{k=1}^{K}\frac{\alpha_kZ_k}{Z}\mathbb{E}_{\vz \sim q_0}  {\Big[}\log\Big(\frac{q_{\theta}(h_{\theta_k}(\vz))}{\widetilde{p}(h_{\theta_k}(\vz))}\Big){\Big]}.
\end{align*}

In practice, given a sampling budget of $S$ samples during training, we distribute the budget evenly across components, i.e., $S_k=\lfloor\frac{S}{K}\rfloor$. The (unbiased) estimator of the objective is then given as
\begin{align*}
\sum_{k=1}^K \frac{\alpha_kZ_k}{Z} \frac{1}{S_k}\sum_{s=1}^{S_k} \log\Big(\frac{q_{\theta}(h_{\theta_k}(\vz^{(s)}))}{\widetilde{p}(h_{\theta_k}(\vz^{(s)}))}\Big), \vz^{(s)} \overset{\text{i.i.d.}}{\sim} q_0,
\end{align*}
which is a fully differentiable objective in terms of mixture weights, allowing us to compute the corresponding gradients via automatic differentiation \citep{kucukelbir2017automatic, morningstar2021automatic}.

\section{EXPERIMENTAL SETUP}\label{app:exp_setup}
In the following, we describe the experimental setup used for the research questions discussed in \Cref{sec:experiments}.
We implement all of our experiments in Python using the \texttt{cirkit} library\footnote{\url{https://github.com/april-tools/cirkit}} \citep{The_APRIL_Lab_cirkit_2024}, which was released under the GPL-3.0 license.
Moreover, we adapt code from the repositories of EigenVI\footnote{\url{https://github.com/dicai/eigenVI}} \citep{cai2024eigenvi}, released under the Apache-2.0 license, and BeyondELBOs\footnote{\url{https://github.com/DenisBless/variational_sampling_methods}} \citep{blessing2024beyond}, released under the MIT license.
Experiments reporting runtime are run on a single NVIDIA RTX A6000 (48GiB VRAM) each and runtimes are measured using \texttt{time.perf\_counter}. Experiments on synthetic and neuro-symbolic targets were also run on NVIDIA RTX A6000 (48GiB VRAM).
Bayesian logistic regression experiments were run on NVIDIA A100 and RTX 3080 Ti GPUs.
We use Weights and Biases\footnote{\url{https://github.com/wandb/wandb}} and PyTorch Lightning\footnote{\url{https://github.com/Lightning-AI/pytorch-lightning}} for training.

\subsection{Target Specifications}\label{app:targets}
In the following, we define the targets used throughout the paper.

\subsubsection{Two-dimensional Targets}
The first three targets are taken from \citet{cai2024eigenvi}.
\paragraph{GMM3 ($\boldsymbol{d=2}$).}  \emph{GMM3} is a $3$-component GMM defined as follows:
\begin{align*}
    p_{\text{GMM3}}(\vx) = 0.4\cdot \mathcal{N}(\vx| [-1, 1]^T, \Sigma) + 0.3\cdot \mathcal{N}(\vx| [1.1, 1.1]^T, \Sigma_2) + 0.3\cdot \mathcal{N}(\vx| [-1, -1]^T, \Sigma_2),
\end{align*}
where $\Sigma = \begin{pmatrix}
0.5 & 0 \\
0 & 0.5
\end{pmatrix}$ and $\Sigma_2 = \begin{pmatrix}
1 & 0 \\
0 & 1
\end{pmatrix}.$

\paragraph{GMM4 ($\boldsymbol{d=2}$).} \emph{GMM4} is a $4$-component GMM defined as follows:
\begin{align*}
    p_{\text{GMM4}}(\vx) = \frac{1}{4}\cdot \mathcal{N}(\vx|[0, 2]^T, \Sigma_1) + \frac{1}{4}\cdot \mathcal{N}(\vx|[-2, 0]^T, \Sigma_2)  + \frac{1}{4}\cdot \mathcal{N}(\vx|[2,0]^T, \Sigma_2) + \frac{1}{4}\cdot \mathcal{N}(\vx|[0,-2]^T, \Sigma_1)
\end{align*}
where $\Sigma_1 = \begin{pmatrix}
0.15^{0.9} & 0 \\
0 & 1
\end{pmatrix}$ and $\Sigma_2 = \begin{pmatrix}
1 & 0 \\
0 & 0.15^{0.9}
\end{pmatrix}.$

\paragraph{Funnel ($\boldsymbol{d=2}$).} The two-dimensional funnel for $\vx=(x_1, x_2)$ is given by
\begin{align*}
    p_{\text{Funnel}}(\vx) = \mathcal{N}(x_1|0, \sigma^2)\mathcal{N}(x_2|0, \exp{(x_1/2)}),
\end{align*}
where $\sigma=1.2$. Note that we add a small constant ($10^{-6}$) to the log-density of the funnel to match the implementation by \citet{cai2024eigenvi}.

\paragraph{Ring ($\boldsymbol{d=2}$).} We define the \emph{Ring} target as a squared SMM with real mixture weights. Note that the SMM need to be renormalized after squaring, hence a normalizing constant appears in the following definition.
\begin{align*}
    p_{\text{Ring}}(\vx) = \frac{1}{Z_{\text{Ring}}}\big{(}\mathcal{N}(\vx|[0,0]^T, \Sigma_1) - 0.46 \cdot \mathcal{N}(\vx|[0,0]^T, \Sigma_2)\big{)}^2 
\end{align*}
where $\Sigma_1 = \begin{pmatrix}
3^2 & 0 \\
0 & 3^2
\end{pmatrix}$ and $\Sigma_2 = \begin{pmatrix}
2^2 & 0 \\
0 & 2^2
\end{pmatrix}.$

\paragraph{DeepRing ($\boldsymbol{d=2}$).} For our analysis of the effect of a safe component on $\Delta\text{IS}$ (see \ref{app:rq_31}), we additionally define the following \emph{DeepRing}, which has a mode in the middle.
\begin{align*}
    p_{\text{DeepRing}}(\vx) = \frac{1}{Z_{\text{DeepRing}}}\big{(} 0.16\cdot \mathcal{N}(\vx|[0,0]^T, \Sigma_1) - 0.36 \cdot \mathcal{N}(\vx|[0,0]^T, \Sigma_2)\big{)}^2 
\end{align*}
where $\Sigma_1 = \begin{pmatrix}
0.6^2 & 0 \\
0 & 0.6^2
\end{pmatrix}$ and $\Sigma_2 = \begin{pmatrix}
1 & 0 \\
0 & 1
\end{pmatrix}.$

\subsubsection{Higher-dimensional Targets}

All of the \emph{Hollow} targets are squared SMMs with two components and with real mixture weights.
\paragraph{Hollow ($\boldsymbol{d=16}$).} 

\begin{align*}
    p_{\text{Hollow}(d=16)}(\vx) = \frac{1}{Z_{\text{Hollow(d=16)}}}\big{(}\mathcal{N}(\vx|[0,0]^T, \sigma_1^2\cdot I^{16\times16}) - 0.3 \cdot \mathcal{N}(\vx|[0,0]^T, \sigma_2^2\cdot I^{16\times16})\big{)}^2 
\end{align*}
where $\sigma_1 = 7$ and $\sigma_2 =6$. $I^{16\times16}$ denotes the 16-dimensional identity matrix.

\paragraph{Hollow ($\boldsymbol{d=32}$).} 

\begin{align*}
    p_{\text{Hollow}(d=32)}(\vx) = \frac{1}{Z_{\text{Hollow(d=32)}}}\big{(}\mathcal{N}(\vx|[0,0]^T, \sigma_1^2\cdot I^{32 \times 32}) - 0.11 \cdot \mathcal{N}(\vx|[0,0]^T, \sigma_2^2\cdot I^{32 \times 32})\big{)}^2 
\end{align*}
where $\sigma_1 = 7$ and $\sigma_2 = 6$. $I^{32 \times 32}$ denotes the $32$-dimensional identity matrix.

\paragraph{Hollow ($\boldsymbol{d=64}$).} 

\begin{align*}
    p_{\text{Hollow}(d=64)}(\vx) = \frac{1}{Z_{\text{Hollow(d=64)}}}\big{(}\mathcal{N}(\vx|[0,0]^T, \sigma_1^2\cdot I^{64 \times64}) - 0.074 \cdot \mathcal{N}(\vx|[0,0]^T, \sigma_2^2\cdot I^{64\times 64})\big{)}^2 
\end{align*}
where $\sigma_1 = 7$ and $\sigma_2 = 6.5$. $I^{64 \times 64}$ denotes the $64$-dimensional identity matrix.

\paragraph{Funnel ($\boldsymbol{d=10}$).} The 10-dimensional \emph{Funnel} is taken from \citet{blessing2024beyond}:
\begin{align*}
    p_{\text{Funnel($D=10$)}}(\vx) = \mathcal{N}(x_1|0, \sigma^2)\mathcal{N}(x_{2}|0, \exp{(x_1)})\ldots \mathcal{N}(x_{10}|0, \exp{(x_1)}),
\end{align*}
where $\sigma=3$.

\paragraph{Bayesian Logistic Regression.} These targets are obtained as posterior distributions of a real valued parameter $\vx \in \mathbb{R}^D$, given a dataset $\{y_{n}, \boldsymbol{z}_{n}\}_{n=1}^{N}$ where $y \in \{ 0,1 \}$ are labels and $\boldsymbol{z}_{n} \in \mathbb{R}^{d_z}$ are covariates. The targets are then given as
$$
p(\vx) :=  p(\vx | \{y_n, \boldsymbol{z}_{n}\}_{n=1}^{N}) \propto \prod_{n} p(y_{n} | \vx, \boldsymbol{z}_n) \cdot p_0(\vx),
$$
where $p(y_{n} | \vx, \boldsymbol{z}_n) = \operatorname{Bernoulli}(y_n; \operatorname{sigmoid}(\vx^\top\boldsymbol{z}_n))$ and $p_0$ is a standard Gaussian prior, $p_0(\vx) = \mathcal{N}(0; I^{D\times D})$. We obtain different targets by using various datasets from \citet{blessing2024beyond}.

\subsubsection{Stanford Drone Dataset}\label{app:pal}
The \emph{Stanford Drone Dataset (SDD)} \citep{robicquet2016learning} consists of top-down  street views of the Stanford campus including the trajectories of pedestrians and vehicles.
We target densities that were fit to the given trajectories with the \emph{probabilistic algebraic layer} (PAL) by \citep{kurscheidt2025probabilistic} using $10$ mixture components and $14$ knots per spline. 
As in \citet{kurscheidt2025probabilistic}, we refer to the first target (\cref{fig:pal_fig_1} and top half of \cref{fig:pal_1_app}) as \emph{scenario 1} and the second target (\cref{fig:pal_2} and bottom half of \cref{fig:pal_1_app}) as \emph{scenario 2}.
Note that PAL models \emph{hard constraints} and hence sets the density of invalid areas (such as buildings, obstacles, and roundabouts) to $0$.
This is impractical for VI with the RKL since the target density appears in the denominator of the objective.
Therefore, we pre-process the PAL densities, such that invalid areas are assigned a log-density of $-20$ for scenario 1 and $-25$ for scenario 2.
We found that these values lead to stable training dynamics for both SMMs and GMMs.

\subsection{RQ1: Scaling sampling with SMMs}\label{app:rq_0}
In our first set of experiments, addressing RQ1), our sampling distribution is a squared SMM with Gaussian inputs. 
The means are initialized with a $\text{Unif}([-0.5, 0.5])$ distribution  %
and standard deviations are drawn from a $\text{Unif}([2, 3])$ distribution. 
The mixture weights are initialized with a $\text{Unif}([-1, 1])$ distribution. We repeat the random initialization until the generated model has at least one negatively weighted component after squaring. 
\cref{tab:rq_1_summary_stats} gives an overview of the average accepted probability when performing rejection sampling (\cref{alg:rejection}) on the resulting SMMs.
The target function $f$ is initialized with $100$ Gaussian components for all settings. 
The means for are initialized using a standard normal distribution and the standard deviations are sampled from a $\text{Unif}([1, 2])$ distribution. 
The weights of the sum layer are sampled from a $\text{Unif}([10000, 100000])$ distribution to encourage a non-zero target expectation in high dimensions.
All methods were sampled with a maximum batch size of $5000$.

\begin{table*}
\caption{Average acceptance probability of the random instances generated for RQ1, given as $\frac{Z}{Z_+}$. Each cell reports the mean and standard deviation over the $30$ generated instances for the corresponding combination of features $(D)$ and components ($K$).}\label{tab:rq_1_summary_stats}
\begin{center}
\begin{tabular}{llll}
\toprule
& \multicolumn{3}{c}{$K$}\\
& \multicolumn{1}{c}{$2$} & \multicolumn{1}{c}{$4$} & \multicolumn{1}{c}{$6$} \\ \cline{2-4}\\[0.01pt]
$D$& \multicolumn{1}{c}{$\frac{Z}{Z_+}$} & \multicolumn{1}{c}{$\frac{Z}{Z_+}$} & \multicolumn{1}{c}{$\frac{Z}{Z_+}$} \\
\midrule
$16$ & $0.489 \pm 0.240$ & $0.373 \pm 0.217$ & $0.349 \pm 0.247$ \\
$32$ & $0.591 \pm 0.193$ & $0.423 \pm 0.171$ & $0.517 \pm 0.200$ \\
$64$ & $0.719 \pm 0.142$ & $0.673 \pm 0.191$ & $0.611 \pm 0.180$ \\
\bottomrule
\end{tabular}
\end{center}
\end{table*}

\subsection{RQ2: Comparing VI strategies for SMMs}\label{app:rq_1}
In the following, we describe the experimental setup that was used to obtain the results in \Cref{tab:rq_1} and \Cref{tab:high_dim}.
For our \emph{squared, complex SMMs}, we do the following:
For each target density, we train models from $5$ different initializations and perform hyperparameter search according to \Cref{tab:hyper_grid}. 
Within each run, we save the $5$ best checkpoints according to the training loss. 
The resulting set of models is what we select the models in \Cref{tab:rq_1} and \Cref{tab:high_dim} from. 
In particular, \emph{we choose the model with the best average training loss} according to the following scheme: For $\Delta\text{VI}$ and RLOO (Rej.), recomputing the training loss is inexpensive, and hence we repeat the estimation $30$ times for targets with $D < 16$, and $10$ times otherwise. To further stabilize the selection for models on synthetic targets, we only consider models for which the average train loss exceeds $-0.1$, as we expect it to be positive for these targets.
For RLOO with ARITS, we repeat the loss estimation $10$ times for targets with less than $16$ dimensions, and compute it only once otherwise. %
All models models in \Cref{tab:rq_1} and \Cref{tab:high_dim} were trained with $10^5$ samples per update step, except for the $10$-dimensional \emph{Funnel} \citep{blessing2024beyond}, which we trained with $10^4$ samples per update for all models.
We use the Adam optimizer \citep{kingma2014adam} for all models.

For \emph{EigenVI} \citep{cai2024eigenvi}, we use a $\text{Unif}([-9,9])$ proposal for all targets and fit the models based on $10^5$ samples from this proposal. 
We use normalized Hermite polynomials as the basis functions of EigenVI, as was done by \citet{cai2024eigenvi} in their experiments.
We train a single EigenVI model with this setup for each target as EigenVI does not rely on stochastic gradient descent. 

For the logistic regression targets (\Cref{tab:logreg_elbo}), we did not run ARITS due computational constraints. The hyperparameter details are in \cref{tab:hyper_grid_logreg}. The model selection procedure used here is the same as of the other targets. The weights for SMMs are initialised so that the real part has a probability of $0.5$ of being negative. This is to encourage initial negative contributions. Then the positive weights absolute values are initialised $\mathsf{Unif}(0.5,2.0)$, while absolute values for the negative weights in $\mathsf{Unif}(0.1, 0.5)$.

\paragraph{Evaluation.} We report the \emph{evidence lower bound (ELBO)} for all settings and the forward Kullback-Leibler divergence (FKL) for synthetic targets. 
We estimate the forward KL for the synthetic densities as
\begin{align*}
\text{FKL}(p, q) = \frac{1}{S}\sum_{s=1}^{S} \log\Big{(}\frac{p(\vx^{(s)})}{q(\vx^{(s)})}\Big{)}, ~~\vx^{(s)} \overset{\text{i.i.d.}}{\sim} p.
\end{align*}
We do not compute the FKL for the Bayesian logistic regressions in \Cref{tab:logreg_elbo} as we do not have access to ground-truth target samples in this setting. 
For synthetics targets, for which we know the normalizing constant, we additionally report the reverse Kullback-Leibler divergence (RKL), which is given as
\begin{align*}
\text{RKL}(q, p) = \frac{1}{S}\sum_{s=1}^{S} \log\Big{(}\frac{q(\vx^{(s)})}{p(\vx^{(s)})}\Big{)}, ~~\vx^{(s)} \overset{\text{i.i.d.}}{\sim} q.
\end{align*}

Lastly, for Bayesian logistic regression targets, for which we neither have access to ground-truth samples nor the normalizing constants, we report the ELBO given as 
\begin{align*}
\text{ELBO}(q, p) = -\frac{1}{S}\sum_{s=1}^{S} \log\Big{(}\frac{q(\vx^{(s)})}{\widetilde{p}(\vx^{(s)})}\Big{)}, ~~\vx^{(s)} \overset{\text{i.i.d.}}{\sim} q.
\end{align*}

\begin{table*}[!htbp]
\caption{Initializations and hyperparameter search spaces for \cref{tab:rq_1} and \cref{tab:high_dim}. Weight decay is applied to the mixture weights only for all models.}\label{tab:hyper_grid}
\centering
\resizebox{\textwidth}{!}{  
\begin{tabular}{lll}
\toprule

\emph{2D Synthetics: GMM3, GMM4, Funnel, Ring}\\ \midrule
\textbf{Model} &  \textbf{Parameter} & \textbf{Initialization}\\
\midrule
\multirow{3}{*}{GMM}   &  mean of Gaussian & $\text{Unif}(-1, 1)$ \\
                        &  stddev of Gaussian & $\text{Unif}(1, 3)$\\
                        & mixture weights & $\frac{1}{K}$\\
\hline
\multirow{4}{*}{SMM ($\mathbb{C}$)}  &  mean of Gaussian & $\text{Unif}(-1, 1)$ \\
                        &  stddev of Gaussian & $\text{Unif}(1, 3)$\\
                        & mixture weights (real) & $\text{Unif}(0,1)$\\
                        & mixture weights (imaginary) & $\mathcal{N}(0,1)$\\               
\midrule
\textbf{Loss} &  \textbf{Hyperparameter/Config} & \textbf{Range} \\
\midrule
\multirow{4}{*}{$\Delta\text{VI}$ and GMM} & lr & \{0.01, 0.001, 0.0001\}\\
& patience & 3000\\
& weight decay & None\\ 
& max steps & 15000\\\midrule
\multirow{4}{*}{RLOO + ARITS}& lr & \{0.01\}\\
& patience & 500\\
& weight decay & None\\ 
& max steps & 15000\\\midrule
\multirow{4}{*}{RLOO + Rejection} & lr & \{0.01\}\\
& patience & 500\\
& weight decay & None\\
& max steps & 15000\\ \midrule

 \emph{Hollow $(d=16, 32, 64)$ $\ldots$ Parameters equal across $d$ unless denoted otherwise}\\ \midrule
\textbf{Model} & \textbf{Parameter} & \textbf{Initialization}\\
\midrule
\multirow{3}{*}{GMM}   &  mean of Gaussian & $\text{Unif}(-1, 1)$ \\
                        &  stddev of Gaussian & $d \in\{16, 64\}: \text{Unif}(5, 7)$, $D=32: \text{Unif}(6, 8)$\\
                        & mixture weights & $\frac{1}{K}$\\
\hline
\multirow{4}{*}{SMM ($\mathbb{C}$)}  &  mean of Gaussian & $\text{Unif}(-1, 1)$ \\
                        &  stddev of Gaussian & 
                         $d \in\{16, 64\}: \text{Unif}(5, 7)$, $D=32: \text{Unif}(6, 8)$\\
                        & mixture weights (real) & $\text{Unif}(0,1)$\\
                        & mixture weights (imaginary) & $\mathcal{N}(0,1)$\\               
\midrule
\textbf{Loss} &  \textbf{Hyperparameter/Config} & \textbf{Range} \\
\midrule
\multirow{4}{*}{$\Delta\text{VI}$ and GMM} & lr & \{0.01, 0.001\}\\
& patience & None \\
& weight decay & \{0.0, 0.001\}\\ %
& max steps & 30000\\\midrule
\multirow{4}{*}{RLOO + ARITS}& lr & $d \in \{16, 32\}: 0.01$, $D=64: 0.001$\\ %
& patience & $100$\\
& weight decay & \{0.0, 0.001\}\\ 
& max steps & 30000\\\midrule
\multirow{4}{*}{RLOO + Rejection} & lr & $d \in \{16, 32\}: 0.01$, $D=64: 0.001$\\
& patience & 1000 \\
& weight decay & \{0.0, 0.001\}\\
& max steps & 30000\\ \midrule

\emph{Funnel $(d=10)$}\\ \midrule
\textbf{Model} &  \textbf{Parameter} & \textbf{Initialization}\\
\midrule
\multirow{3}{*}{GMM}   &  mean of Gaussian & $\text{Unif}(-4, 4)$ \\
                        &  stddev of Gaussian & $\text{Unif}(2, 4)$\\
                        & mixture weights & $\frac{1}{K}$\\
\hline
\multirow{4}{*}{SMM ($\mathbb{C}$)}  &  mean of Gaussian & $\text{Unif}(-4, 4)$ \\
                        &  stddev of Gaussian & $\text{Unif}(2, 4)$\\
                        & mixture weights (real) & $\text{Unif}(0,1)$\\
                        & mixture weights (imaginary) & $\mathcal{N}(0,1)$\\               
\midrule
\textbf{Loss} &  \textbf{Hyperparameter/Config} & \textbf{Range} \\
\midrule
\multirow{4}{*}{$\Delta\text{VI}$ and GMM} & 
lr & \{0.01, 0.001\}\\
& patience & None\\
& weight decay & \{0, 0.001\}\\ 
& max steps & 60000\\\midrule
\multirow{4}{*}{RLOO + Rejection} & 
lr & \{0.01, 0.001\}\\
& patience & None\\
& weight decay & \{0, 0.001\}\\ 
& max steps & 60000\\
\bottomrule
\end{tabular}
}
\end{table*}

\begin{table*}[!htbp]
\begin{center}
\caption{Selected hyperparameters for \Cref{tab:rq_1} and \Cref{tab:high_dim}. For the full search space see \Cref{tab:hyper_grid}. The number of components, $K$, was fixed for these experiments. Weight dacay was only used for higher-dimensional targets and was only applied to the mixture weights.}\label{tab:hyper_selected}

\begin{tabular}{llllll}\toprule
Target & Hyperparameter & GMM & $\Delta\text{VI}$ & RLOOKL (Rej.) & RLOOKL (ARITS)\\ \midrule

\multirow{2}{*}{GMM3 ($D=2$)} & lr & 0.001 & 0.001 & 0.01 & 0.01\\
& $K$ & 3 & 3 & 3 & 3\\ \hline

\multirow{2}{*}{GMM4 ($D=2$)} & lr & 0.001 & 0.01 & 0.01 & 0.01\\
& $K$ & 4 & 4 & 4 & 4\\ \hline

\multirow{2}{*}{Funnel ($D=2$)} & lr & 0.01 & 0.001 & 0.01 & 0.01\\
& $K$ & 16 & 16 & 16 & 16\\ \hline

\multirow{2}{*}{Ring ($D=2$)} & lr & 0.01 & 0.01 & 0.01 & 0.01\\
& $K$ & 2 & 2 & 2 & 2\\ \midrule

\multirow{3}{*}{Hollow ($D=16$)} & lr & 0.01 & 0.01 & 0.01 & 0.01\\
 & weight decay & 0.001 & 0 & 0 & 0.001\\
 & $K$ & 2 & 2 & 2 & 2\\ \hline
 
\multirow{3}{*}{Hollow ($D=32$)} & lr & 0.01 & 0.001 & 0.01 & 0.01\\
& weight decay & 0 & 0 & 0 & 0.001\\
& $K$ & 2 & 2 & 2 & 2\\ \hline

\multirow{3}{*}{Hollow ($D=64$)} & lr & 0.01 & 0.001 & 0.001 & /\\
& weight decay & 0 & 0.001 & 0.001 & /\\
& $K$ & 2 & 2 & 2 & /\\ \midrule

\multirow{3}{*}{Funnel ($D=10$)} & lr & 0.001 & 0.01 & 0.01 & /\\
& weight decay & 0.001 & 0 & 0 & /\\
& $K$ & 16 & 16 & 16 & /\\
\bottomrule
\end{tabular}
\end{center}
\end{table*}

\begin{table*}[!htbp]
\caption{Initializations and hyperparameter search spaces for BLR posteriors.}\label{tab:hyper_grid_logreg}
\centering
\begin{tabular}{lll}
\toprule
\emph{Logistic regressions}\\ \midrule
\textbf{Model} & \textbf{Parameter} & \textbf{Initialization} \\
\midrule
\multirow{3}{*}{GMM}
  & mean of Gaussian & $\text{Unif}(-2, 2)$ \\
  & stddev of Gaussian & $\text{Unif}(6, 8)$ \\
  & mixture weights & $1/K$ \\
\hline
\multirow{4}{*}{SMM ($\mathbb{C}$)}
  & mean of Gaussian & $\text{Unif}(-2, 2)$ \\
  & stddev of Gaussian & $\text{Unif}(6, 8)$ \\
  & mixture weights & positive: $\mathsf{Unif}(0.5, 2.0)$;  \\
    &  & negative: $\mathsf{Unif}(0.1, 0.5)$;  \\
  &  & Prob. of negative: $0.5$  \\

  \\
\midrule
\textbf{Loss} & \textbf{Hyperparameter/Config} & \textbf{Range} \\
\midrule
\multirow{6}{*}{$\Delta\text{VI}$ and GMM}
  & lr & $\{0.001,\,0.01\}$ \\
  & patience & $1500$ \\
  & max steps & $10000$ \\
  & weight decay & $0$ \\
  & $K$ (components) & $\{4,\,8,\,16\}$ \\
  & samples/update & $5000$ \\
  & optimizer & Adam \\ \midrule
\multirow{6}{*}{RLOO + Rejection}
  & lr & $\{0.001,\,0.01\}$ \\
  & patience & $1500$ \\
  & max steps & $10000$ \\
  & weight decay & $0$ \\
  & $K$ (components) & $\{4,\,8,\,16\}$ \\
  & samples/update & $5000$ \\
\bottomrule
\end{tabular}
\end{table*}

\newpage

\section{Additional Results and Experiments}\label{app:additional_results}
\subsection{Complete Results for RQ1}\label{tab:table_runtime}
\Cref{tab:full_rq_1} reports the concrete runtime and estimation error for the results visualized in \Cref{runtime_exp}. Additionally, it reports the results when choosing $K=2$ and $K=4$, while \Cref{runtime_exp} only shows $K=6$. The general setup is the same across all experiments (see \Cref{app:rq_0}). All methods were sampled with a batch size of $5000$. We note that for rejection sampling, the very first instance that was run in our setup took considerably longer than the remaining, explaining the comparatively high standard deviation of the cell $(D=16, K=2, S=10000)$. The remaining runtimes were stable.

\begin{table}[h]
\caption{\textbf{Rejection sampling and \dis{} are consistently faster than ARITS for expectation estimation while achieving comparable estimation quality when given a sufficient sampling budget.} Results for MC comparing our method (\dis) with ARITS for a varying number of features ($d$) and components ($K$) as well as different sampling budgets ($S$) for \dis. The error is given as $\log(|\widehat{I} -  I|)-\log(I)$, hence lower is better, and time is in seconds. Results are averaged over $30$ initializations of $q_{\text{SMM}}$ and $f$ (mean $\pm$ stddev).}\label{tab:full_rq_1}
\resizebox{\textwidth}{!}{
\begin{tabular}{lrrrrrrrr}
\toprule
&&&\multicolumn{6}{c}{\textbf{Number of components $(\boldsymbol{K})$}}\\
\cmidrule(lr){4-9}
\multicolumn{3}{c}{} & \multicolumn{2}{c}{$\boldsymbol{2}$} & \multicolumn{2}{c}{$\boldsymbol{4}$} & \multicolumn{2}{c}{$\boldsymbol{6}$} \\
\cmidrule(lr){4-5} \cmidrule(lr){6-7} \cmidrule(lr){8-9}
\textbf{Method} &  \boldsymbol{$D$} & \boldsymbol{$S$} & $\log(| \widehat{I} -  I|) - \log(I)$ ($\downarrow$) & Time (s) & $\log(| \widehat{I} -  I|) - \log(I)$ ($\downarrow$)  & Time (s) & $\log(| \widehat{I} -  I|) -\log(I)$ ($\downarrow$) & Time (s) \\
\midrule
$\Delta\text{IS}$ & 16 & 10000 & -2.221 $\pm$ 1.453 & 0.028 $\pm$ 0.002 & -1.667 $\pm$ 1.671 & 0.029 $\pm$ 0.000 & -1.425 $\pm$ 1.544 & 0.033 $\pm$ 0.000 \\
$\Delta\text{IS}$ & 16 & 100000 & -2.989 $\pm$ 1.218 & 0.201 $\pm$ 0.003 & -2.436 $\pm$ 1.349 & 0.211 $\pm$ 0.002 & -2.714 $\pm$ 1.818 & 0.232 $\pm$ 0.003 \\
$\Delta\text{IS}$ & 16 & 1000000 & -4.138 $\pm$ 1.379 & 1.901 $\pm$ 0.031 & -3.617 $\pm$ 1.391 & 2.002 $\pm$ 0.025 & -3.854 $\pm$ 1.664 & 2.194 $\pm$ 0.029 \\
Rej. & 16 & 10000 & -2.727 $\pm$ 1.082 & 0.028 $\pm$ 0.017 & -3.323 $\pm$ 1.880 & 0.024 $\pm$ 0.001 & -3.185 $\pm$ 1.635 & 0.025 $\pm$ 0.001 \\
Rej. & 16 & 100000 & -4.221 $\pm$ 1.284 & 0.232 $\pm$ 0.011 & -3.874 $\pm$ 0.967 & 0.232 $\pm$ 0.011 & -4.004 $\pm$ 1.029 & 0.235 $\pm$ 0.005 \\
Rej. & 16 & 1000000 & -5.520 $\pm$ 1.365 & 2.283 $\pm$ 0.058 & -5.384 $\pm$ 1.823 & 2.284 $\pm$ 0.046 & -5.189 $\pm$ 0.999 & 2.338 $\pm$ 0.052 \\
ARITS & 16 & 10000 & -3.415 $\pm$ 1.084 & 4.905 $\pm$ 0.073 & -3.823 $\pm$ 1.185 & 5.517 $\pm$ 0.066 & -3.307 $\pm$ 1.116 & 6.794 $\pm$ 0.018 \\ \midrule
$\Delta\text{IS}$ & 32 & 10000 & -0.806 $\pm$ 1.539 & 0.035 $\pm$ 0.001 & -0.009 $\pm$ 1.090 & 0.037 $\pm$ 0.001 & -0.830 $\pm$ 1.161 & 0.040 $\pm$ 0.000 \\
$\Delta\text{IS}$ & 32 & 100000 & -1.860 $\pm$ 1.173 & 0.274 $\pm$ 0.013 & -1.197 $\pm$ 1.370 & 0.284 $\pm$ 0.002 & -1.845 $\pm$ 1.243 & 0.304 $\pm$ 0.003 \\
$\Delta\text{IS}$ & 32 & 1000000 & -2.948 $\pm$ 1.412 & 2.612 $\pm$ 0.030 & -2.340 $\pm$ 1.323 & 2.719 $\pm$ 0.018 & -2.819 $\pm$ 1.213 & 2.919 $\pm$ 0.027 \\
Rej. & 32 & 10000 & -1.795 $\pm$ 0.973 & 0.031 $\pm$ 0.001 & -1.561 $\pm$ 0.822 & 0.030 $\pm$ 0.001 & -1.492 $\pm$ 1.101 & 0.033 $\pm$ 0.001 \\
Rej. & 32 & 100000 & -2.343 $\pm$ 0.964 & 0.291 $\pm$ 0.012 & -2.475 $\pm$ 1.305 & 0.283 $\pm$ 0.011 & -2.304 $\pm$ 1.051 & 0.313 $\pm$ 0.019 \\
Rej. & 32 & 1000000 & -3.828 $\pm$ 0.951 & 2.904 $\pm$ 0.121 & -3.543 $\pm$ 1.161 & 2.825 $\pm$ 0.106 & -3.903 $\pm$ 1.117 & 3.112 $\pm$ 0.127 \\
ARITS & 32 & 10000 & -1.713 $\pm$ 0.816 & 10.316 $\pm$ 0.107 & -1.654 $\pm$ 0.743 & 13.042 $\pm$ 0.050 & -1.865 $\pm$ 0.916 & 19.740 $\pm$ 0.048 \\ \midrule
$\Delta\text{IS}$ & 64 & 10000 & -0.253 $\pm$ 0.858 & 0.051 $\pm$ 0.002 & 0.144 $\pm$ 1.320 & 0.054 $\pm$ 0.003 & 0.054 $\pm$ 1.601 & 0.055 $\pm$ 0.002 \\
$\Delta\text{IS}$ & 64 & 100000 & -0.035 $\pm$ 1.313 & 0.428 $\pm$ 0.010 & -0.588 $\pm$ 1.047 & 0.444 $\pm$ 0.016 & -0.393 $\pm$ 1.224 & 0.455 $\pm$ 0.009 \\
$\Delta\text{IS}$ & 64 & 1000000 & -0.830 $\pm$ 1.612 & 4.136 $\pm$ 0.073 & -1.075 $\pm$ 1.379 & 4.256 $\pm$ 0.098 & -0.878 $\pm$ 1.249 & 4.408 $\pm$ 0.084 \\
Rej. & 64 & 10000 & -0.405 $\pm$ 0.626 & 0.050 $\pm$ 0.002 & -0.356 $\pm$ 0.614 & 0.051 $\pm$ 0.003 & -0.348 $\pm$ 0.843 & 0.057 $\pm$ 0.002 \\
Rej. & 64 & 100000 & -0.967 $\pm$ 0.927 & 0.445 $\pm$ 0.029 & -0.553 $\pm$ 0.498 & 0.436 $\pm$ 0.024 & -0.948 $\pm$ 1.517 & 0.496 $\pm$ 0.027 \\
Rej. & 64 & 1000000 & -1.471 $\pm$ 1.008 & 4.832 $\pm$ 0.185 & -1.316 $\pm$ 0.716 & 4.877 $\pm$ 0.306 & -1.389 $\pm$ 1.286 & 5.462 $\pm$ 0.225 \\
ARITS & 64 & 10000 & -0.566 $\pm$ 0.836 & 22.059 $\pm$ 0.235 & -0.509 $\pm$ 0.781 & 37.049 $\pm$ 0.051 & -0.415 $\pm$ 0.622 & 64.612 $\pm$ 0.194 \\

\bottomrule
\end{tabular}
}
\end{table}

\subsection{Further Evaluation of RQ2}
In the following, we provide additional results and evaluations for our VI experiments. We group these results according to their overarching research question in the main text. 

\subsubsection{RQ2.1: Quantitative Comparison to EigenVI}\label{app:rq21}
In this section, we provide a quantitative comparison of the models depicted in \Cref{tab:rq_1}.
All models, except for EigenVI, were fit with $10^5$ samples per update step with the Adam optimizer \citep{kingma2014adam} and hyperparameter search according to \Cref{tab:hyper_grid}.
The selected hyperparameters for each density can be found in \Cref{tab:hyper_selected}.
The number of components is fixed to $3$ for \emph{GMM3}, $4$ for \emph{GMM4}, $16$ for \emph{Funnel}
and $2$ for \emph{Ring}.
All EigenVI models were fit using $10^5$ samples from a $\text{Unif}([-9, 9])$ distribution.
We provide the results for EigenVI with two different number of parameters: \emph{EigenVI (S):} a model that roughly matches ours in terms of parameter count, \emph{EigenVI (L):} the largest model reported by \citep{cai2024eigenvi}.
Since the \emph{Ring} target was not used by \citet{cai2024eigenvi}, to determine a suitable \emph{EigenVI (L)} model, we increased the parameter count until we observed a good fit.
Following \citep{cai2024eigenvi}, we report the forward KL (FKL) between the target and the variational approximation.
The reported forward FKL is estimated from $10^5$ target samples and averaged over $10$ repeated estimations.
We re-estimated the FKL for EigenVI models ourselves in order to have a consistent setup for all methods. We found the resulting FKL estimates to be very close to the ones reported by \citet{cai2024eigenvi} for GMM3 and GMM4.
For the Funnel, we observe worse results than reported in \citet{cai2024eigenvi} by roughly one order of magnitude.
\Cref{tab:eigen-vi-quants} summarizes the results.
EigenVI can achieve a good fit for all targets in terms of FKL, but generally requires a high parameter count to do so.
This is likely do to the fact that EigenVI uses non-learnable components.
Our squared SMM models with complex mixture weights can achieve better or comparable fits to EigenVI while being more parameter-efficient.
However, our VI strategies require hyperparameter tuning (see \Cref{tab:hyper_grid}) and can be sensitive to the initialization (see \Cref{fig:sample_size_boxplot} and \Cref{fig:init_sensitivity}).

\begin{figure*}[t]
\resizebox{\textwidth}{!}{
\begin{tabular}{ccccccccccc}
\toprule
 & \multicolumn{9}{c}{\boldsymbol{$\Delta\text{VI}$}}\\
\cline{2-11}\\
\includegraphics[scale=0.12]{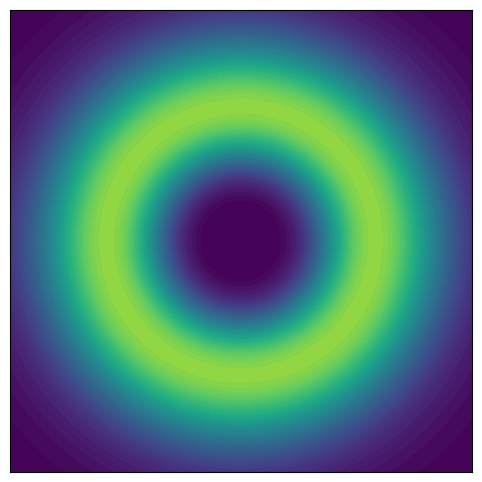} &
\includegraphics[scale=0.12]{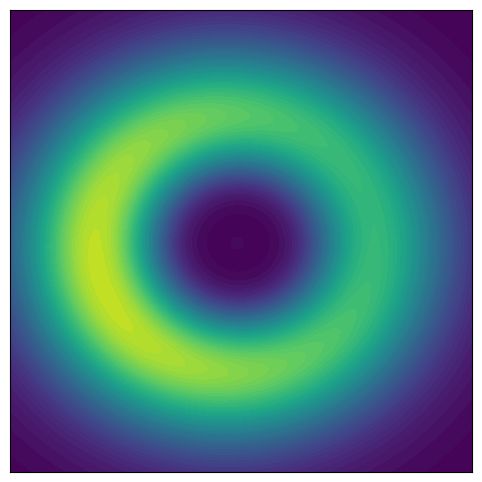} &
\includegraphics[scale=0.12]{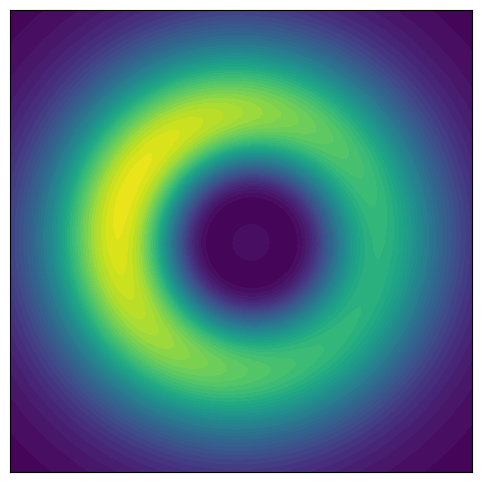} &
\includegraphics[scale=0.12]{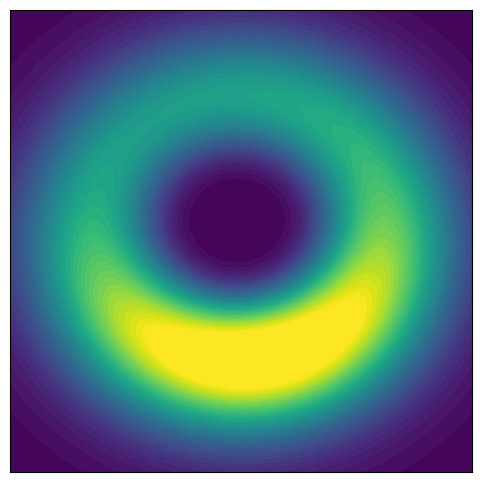} &
\includegraphics[scale=0.12]{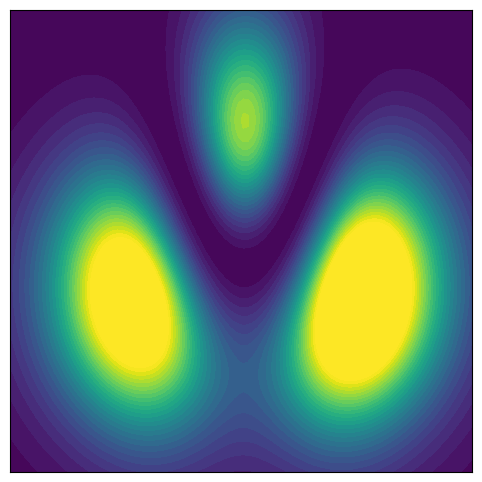} &
\includegraphics[scale=0.12]{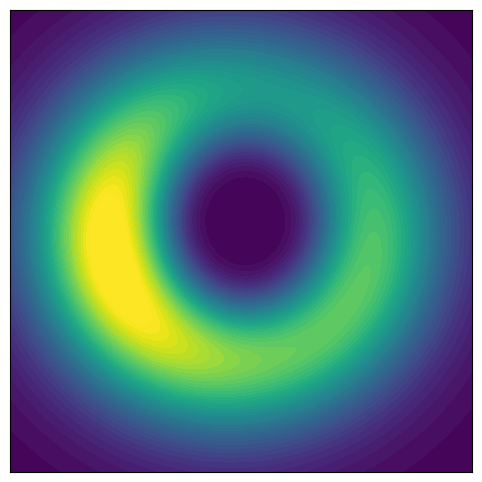} &
\includegraphics[scale=0.12]{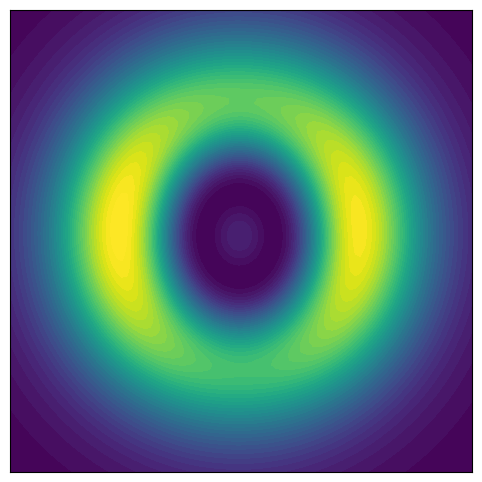} &
\includegraphics[scale=0.12]{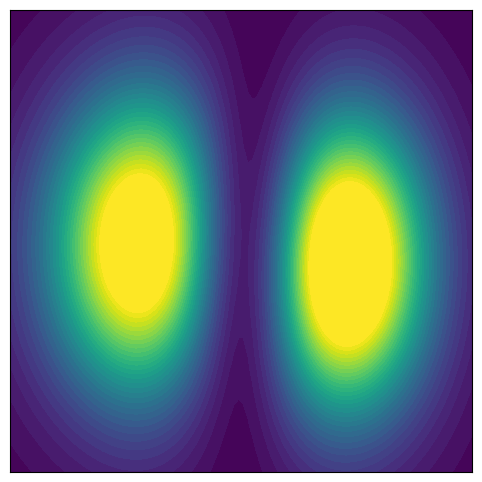} &
\includegraphics[scale=0.12]{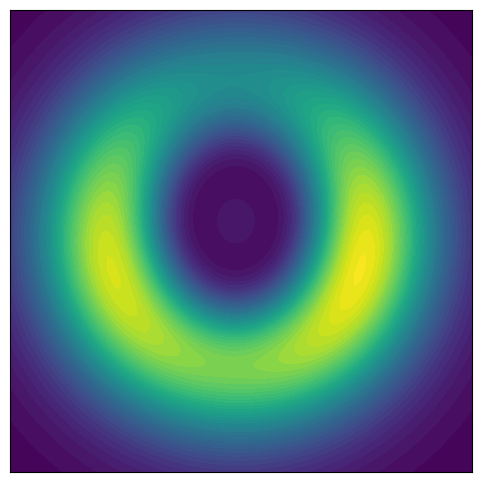} &
\includegraphics[scale=0.12]{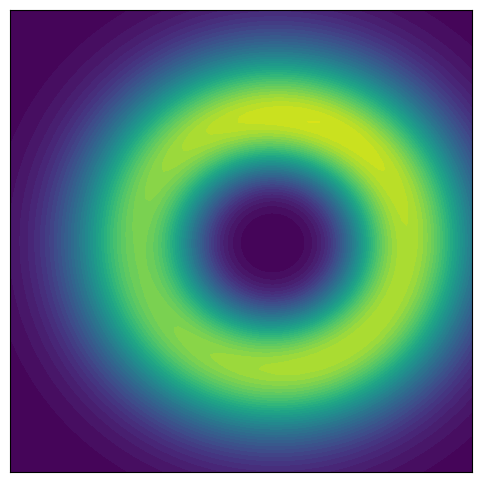} &
\includegraphics[scale=0.12]{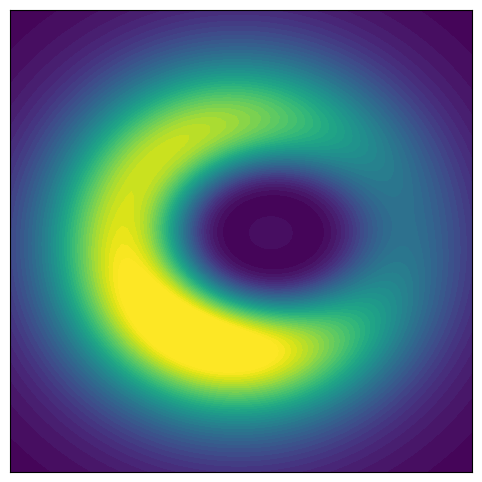}\\

& \multicolumn{9}{c}{\textbf{RLOO (Rej.)}}\\
\cline{2-11}\\
\includegraphics[scale=0.12]{figures/boxplot_models_vis/target.png} &
\includegraphics[scale=0.12]{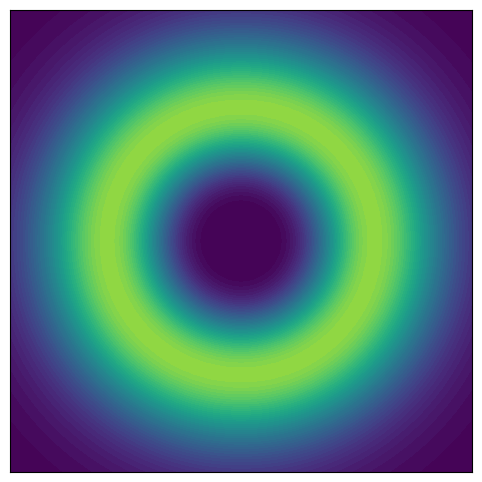} &
\includegraphics[scale=0.12]{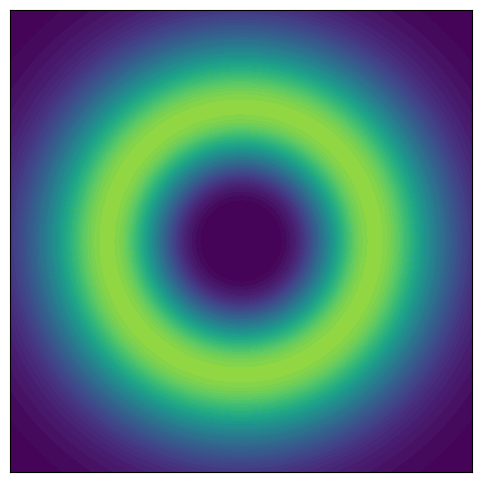} &
\includegraphics[scale=0.12]{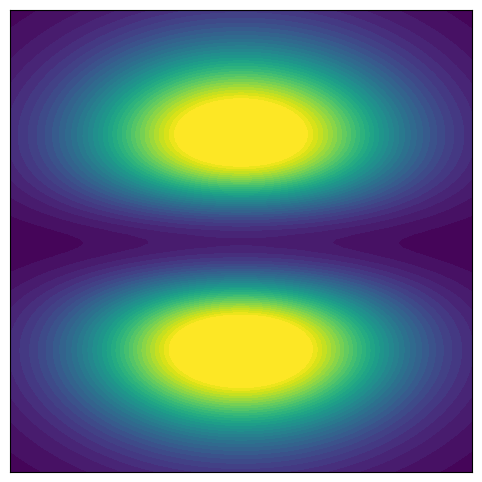} &
\includegraphics[scale=0.12]{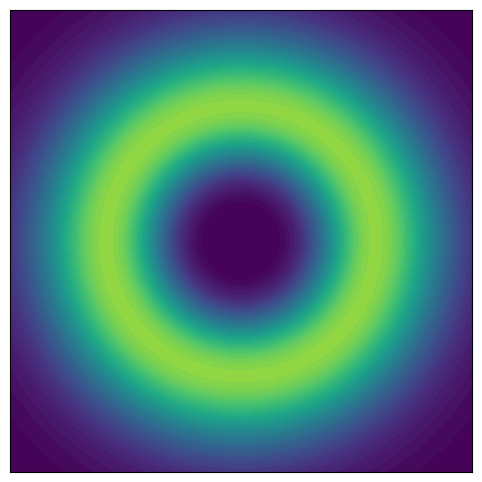} &
\includegraphics[scale=0.12]{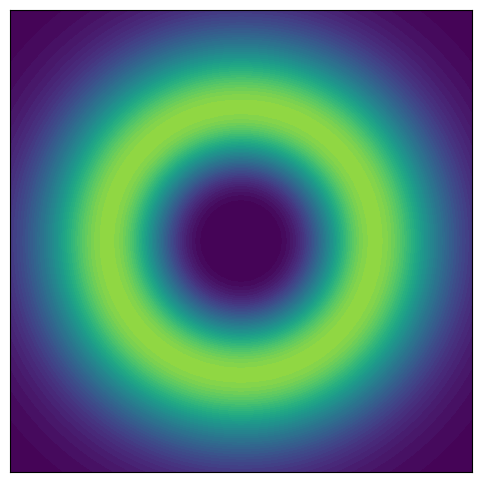} &
\includegraphics[scale=0.12]{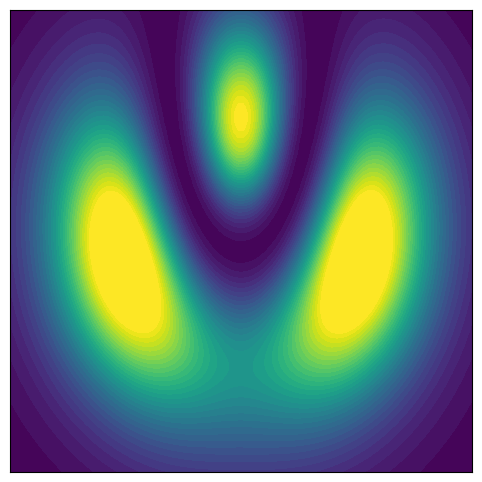} &
\includegraphics[scale=0.12]{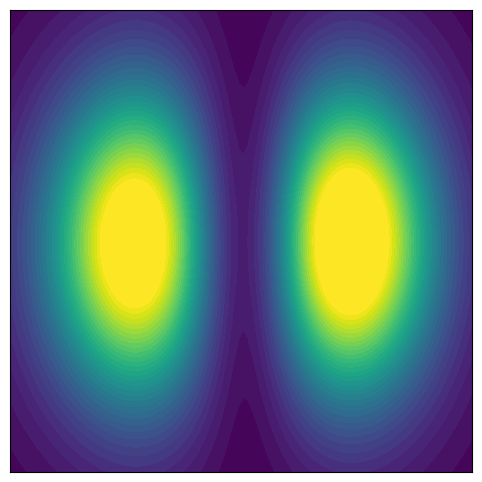} &
\includegraphics[scale=0.12]{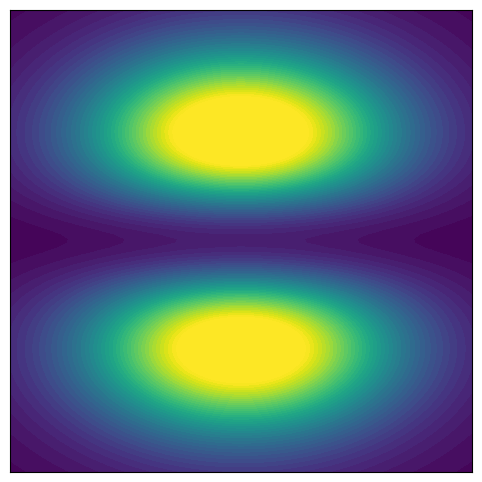} &
\includegraphics[scale=0.12]{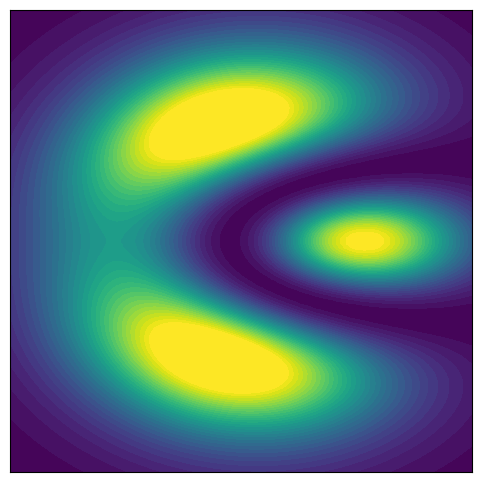} &
\includegraphics[scale=0.12]{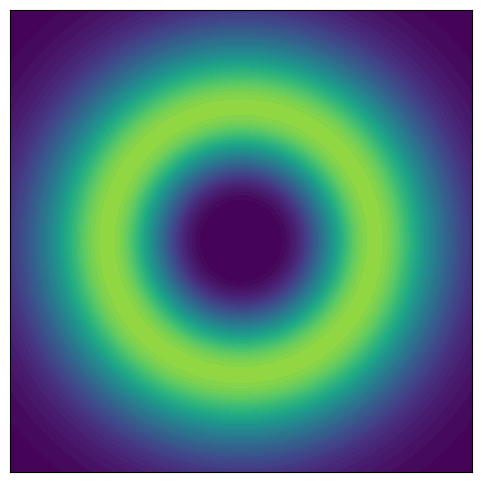} \\

& \multicolumn{9}{c}{\textbf{RLOO (ARITS)}}\\
\cline{2-11}\\
\includegraphics[scale=0.12]{figures/boxplot_models_vis/target.png} &
\includegraphics[scale=0.12]{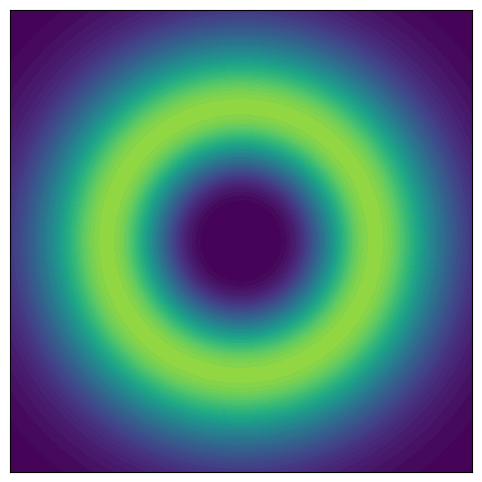} &
\includegraphics[scale=0.12]{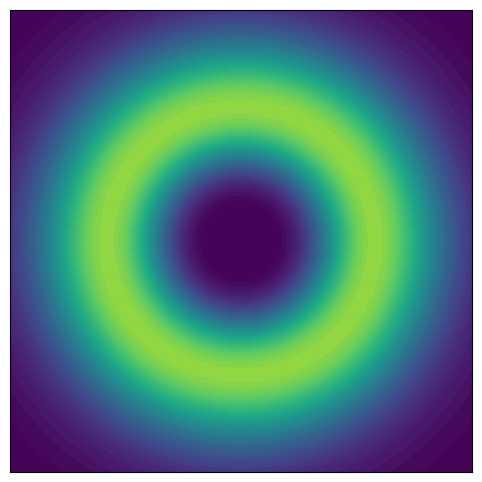} &
\includegraphics[scale=0.12]{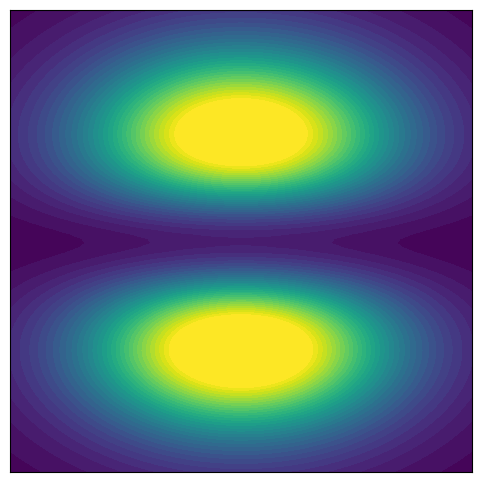} &
\includegraphics[scale=0.12]{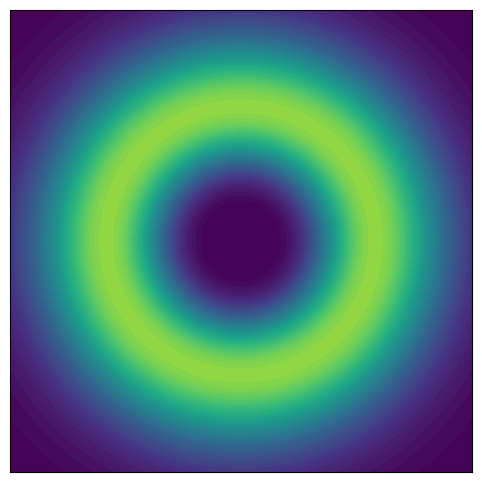} &
\includegraphics[scale=0.12]{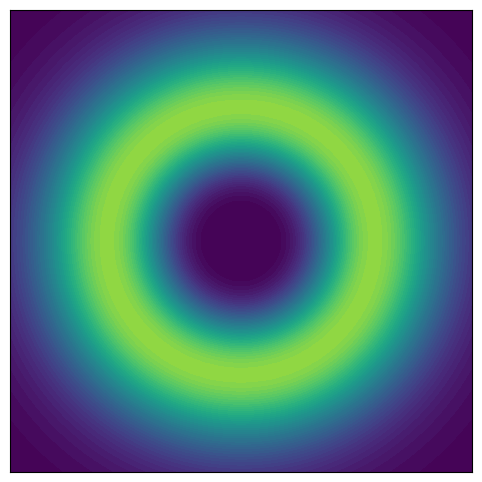} &
\includegraphics[scale=0.12]{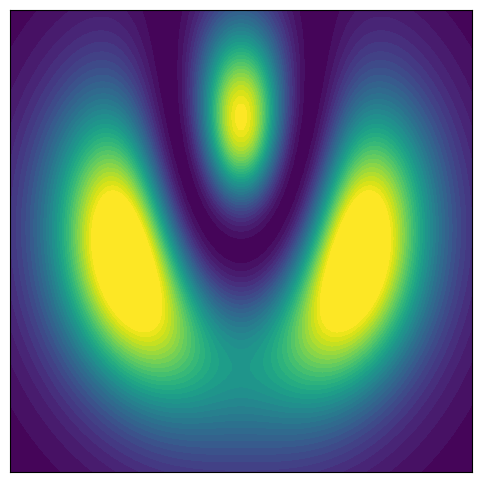} &
\includegraphics[scale=0.12]{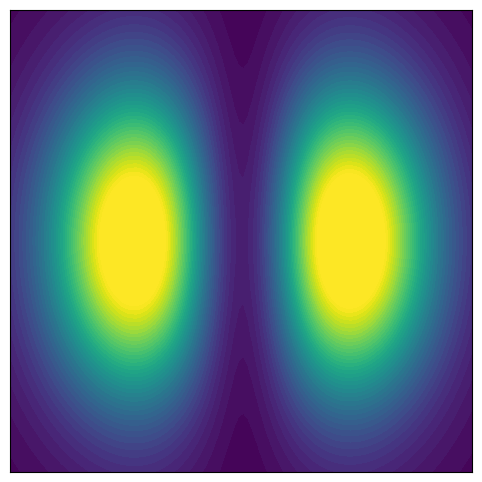} &
\includegraphics[scale=0.12]{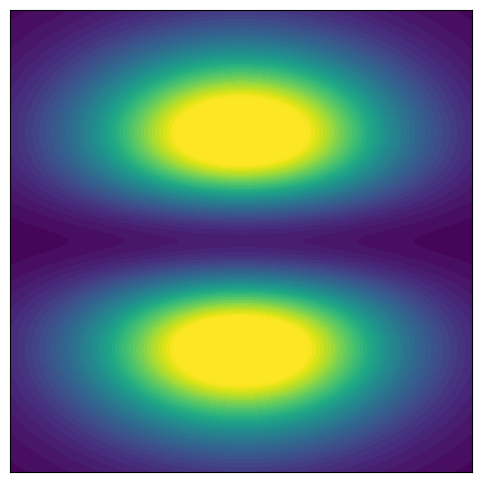} &
\includegraphics[scale=0.12]{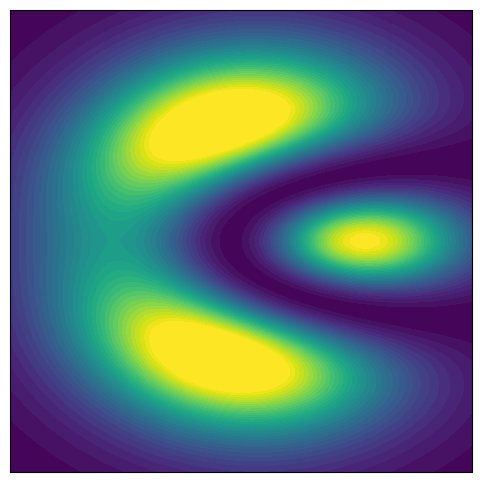} &
\includegraphics[scale=0.12]{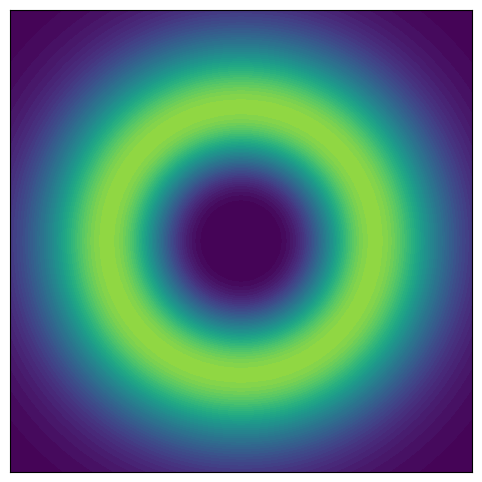} \\ \bottomrule
\end{tabular}
}
\caption{Visual comparisons of models obtained with $\boldsymbol{S=10^6}$ samples per update step for $10$ initializations. 
The first column shows the target.
The depicted models were used to create \Cref{fig:sample_size_boxplot}.}\label{fig:boxplot_model_vis}
\end{figure*}

\begin{table*}[!htbp]
\caption{\textbf{Our proposed VI variants for squared SMMs can achieve comparable performance to EigenVI while being more parameter-efficient.} All models are fit and selected as described in \Cref{app:rq_1}. See \Cref{tab:rq_1} for a visualization of the densities. The FKL is estimated from $10^5$ target samples and estimation is repeated $10$ times (reported is mean $\pm$ stddev). \#P denotes the parameter count. All targets are two-dimensional.}\label{tab:eigen-vi-quants}
\resizebox{\textwidth}{!}{
\begin{tabular}{l ll ll ll ll ll}\toprule
 & \multicolumn{2}{c}{EigenVI (S)} & 
    \multicolumn{2}{c}{EigenVI (L)} &
    \multicolumn{2}{c}{$\Delta\text{VI}$} &
    \multicolumn{2}{c}{RLOO (Rej.)} &
    \multicolumn{2}{c}{RLOO (ARITS)} \\
& \# P & FKL ($\downarrow$) & \# P & FKL ($\downarrow$) & \# P & FKL ($\downarrow$) & \# P & FKL ($\downarrow$) &  \# P & FKL ($\downarrow$)\\ \midrule
 GMM3 &
 $16$ & $1.78\cdot 10^{-2} \pm 5.90\cdot 10^{-4}$ &
 $64$ & $5.55\cdot 10^{-4} \pm 1.07\cdot 10^{-4}$ &
 $18$ & $2.04 \cdot 10^{-4} \pm 8.36 \cdot 10^{-5}$ %
 & %
 $18$ & $2.34 \cdot 10^{-4} \pm 5.37 \cdot 10^{-5}$
 & %
 $18$ & $2.63 \cdot 10^{-4} \pm 9.14 \cdot 10^{-5}$
 \\
 GMM4 & 
 $25$ & $1.38 \cdot 10^{-1} \pm 1.40\cdot 10^{-3}$ &
 $196$ & $4.93\cdot 10^{-3} \pm 4.77\cdot 10^{-4}$ &
 $24$ & $5.51 \cdot 10^{-3} \pm 2.90 \cdot 10^{-4}$
 & 
 $24$ & $1.07 \cdot 10^{-4} \pm 4.81 \cdot 10^{-5}$ 
 & 
 $24$ & $7.44 \cdot 10^{-5} \pm 3.88 \cdot 10^{-5}$
 \\
 Funnel & 
 $100$ &  $2.43\cdot 10^{-1} \pm 6.40\cdot 10^{-3}$ &
 $256$ & $2.65 \cdot 10^{-1} \pm 5.00 \cdot 10^{-3}$ & %
 $96$ & $4.11 \cdot 10^{-2} \pm 8.31 \cdot 10^{-4}$
 & %
 $96$ & $8.17 \cdot 10^{-4} \pm 1.94 \cdot 10^{-4}$& %
 $96$ & $1.28 \cdot 10^{-3} \pm 1.68 \cdot 10^{-4}$ %
 \\
 Ring & 
 $16$ & $2.32\cdot 10^{0} \pm 8.02\cdot 10^{-3}$ &
$196$ & $1.62\cdot 10^{-2} \pm 6.33\cdot 10^{-4}$ &
$12$ & $8.35 \cdot 10^{-2} \pm 1.21 \cdot 10^{-3}$
&%
$12$ & $1.24 \cdot 10^{-5} \pm 1.74 \cdot 10^{-5}$
& %
$12$ & $1.98 \cdot 10^{-6} \pm 1.01 \cdot 10^{-5}$
 \\
\bottomrule
\end{tabular}
}
\end{table*}

\subsubsection{RQ2.2: Model Visualizations for Ring Target}
\label{app:rq22}
The boxplots in \Cref{fig:sample_size_boxplot} are created based on $10$ random initializations, taking the best recorded checkpoint for each and estimating the RKL and FKL between the model and the target from $10^4$ samples.
All models were trained as described in \Cref{app:rq_1} with a learning rate of $10^{-2}$ and no weight decay. 
See \Cref{fig:boxplot_model_vis} for visualizations of the learned models at sample size $10^6$.
At a large sampling budget of $S=10^6$, $\Delta\text{VI}$ manages to capture a ring-like structure for more initializations than the RLOO models, which tend to get stuck in local optima.
However, the best RLOO models are visually a better fit to the target than the best $\Delta\text{VI}$ models.
Potentially, better initializations or a different choice of hyperparameters could mitigate the issues $\Delta\text{VI}$ and the RLOO variants face on this target.

\subsubsection{RQ2.3: Initializations and Resulting Models for Hollow Target}
\label{app:rq23}
\begin{figure*}[t]
\resizebox{\textwidth}{!}{
\begin{tabular}{cccccc}
\toprule
& \multicolumn{5}{c}{\textbf{Initialization}} \\ \cline{2-6}\\
\includegraphics[scale=0.3]{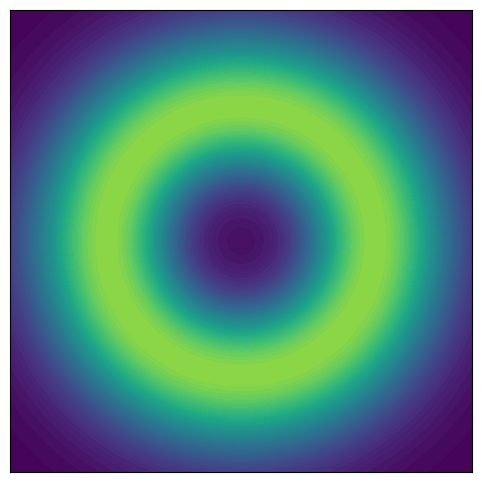} &
\includegraphics[scale=0.3]{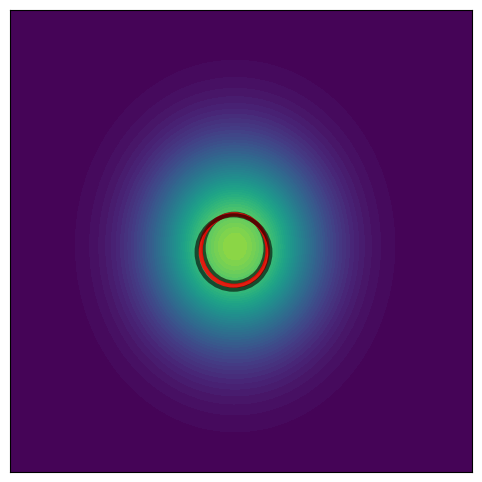} &
\includegraphics[scale=0.3]{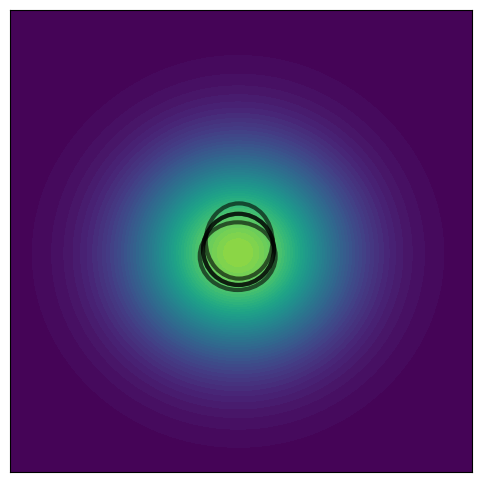} &
\includegraphics[scale=0.3]{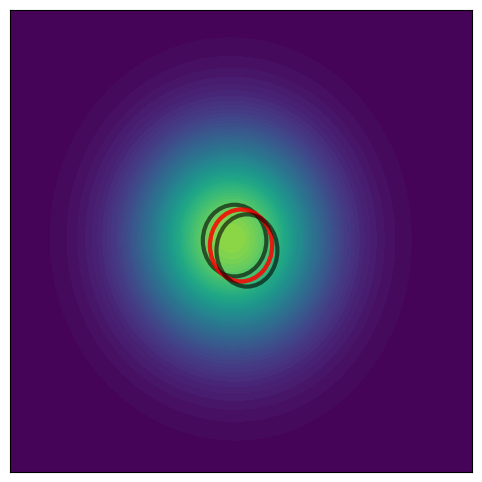} &
\includegraphics[scale=0.3]{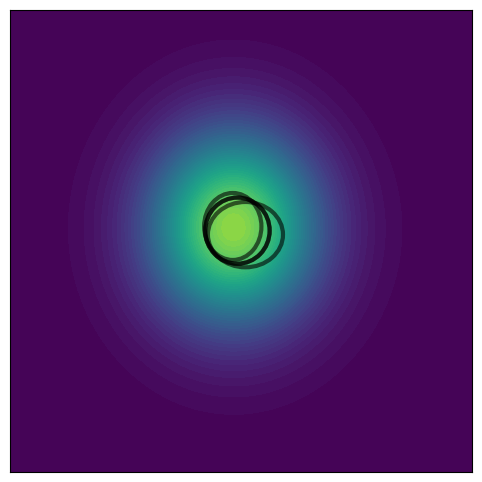} &
\includegraphics[scale=0.3]{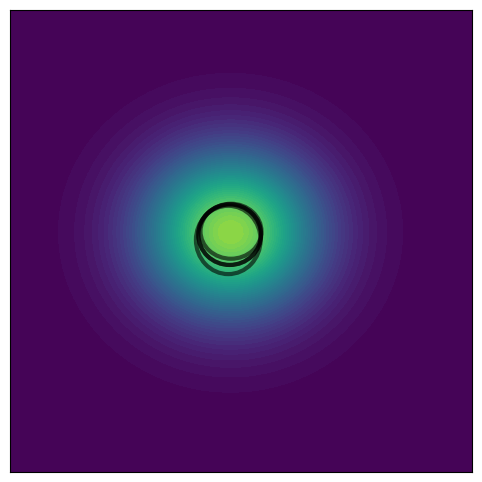} \\

& \multicolumn{5}{c}{\boldsymbol{$\Delta\text{VI}$}} \\ \cline{2-6}\\
\includegraphics[scale=0.3]{figures/init_sensitivity/target.png} &
\includegraphics[scale=0.3]{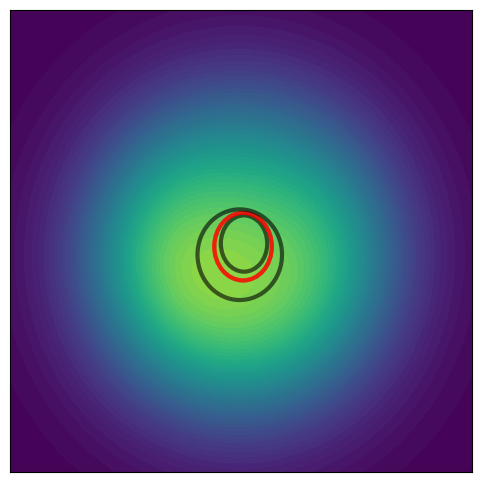} &
\includegraphics[scale=0.3]{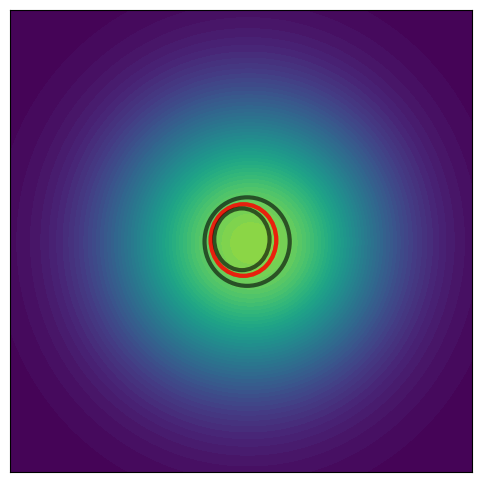} &
\includegraphics[scale=0.3]{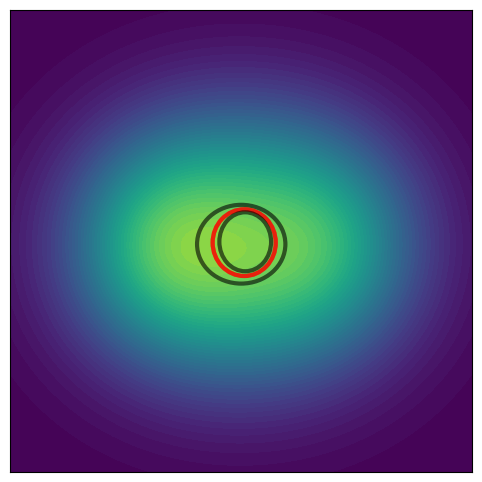} &
\includegraphics[scale=0.3]{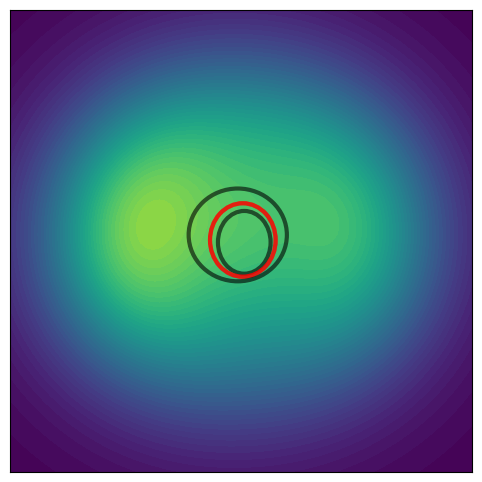} &
\includegraphics[scale=0.3]{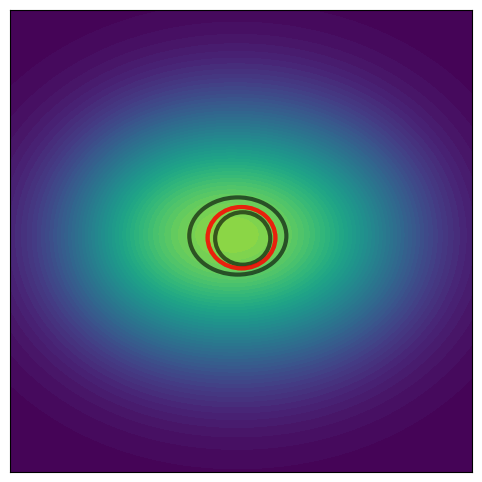} \\

& \multicolumn{5}{c}{\textbf{RLOO (Rej.)}} \\  \cline{2-6}\\
\includegraphics[scale=0.3]{figures/init_sensitivity/target.png} &
\includegraphics[scale=0.3]{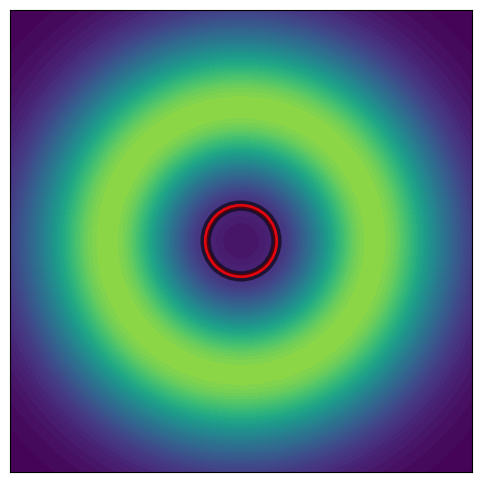} &
\includegraphics[scale=0.3]{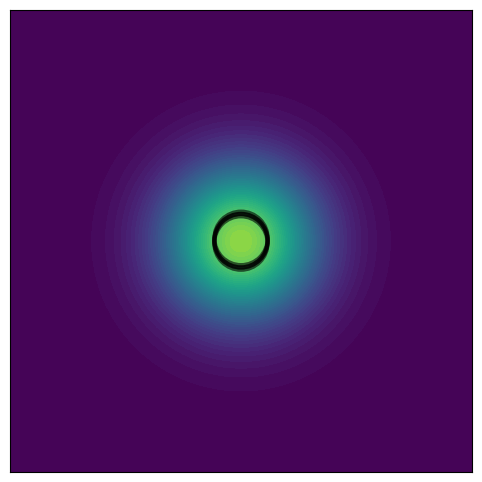} &
\includegraphics[scale=0.3]{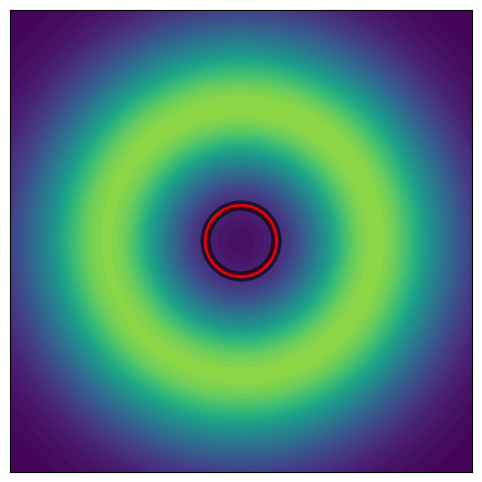} &
\includegraphics[scale=0.3]{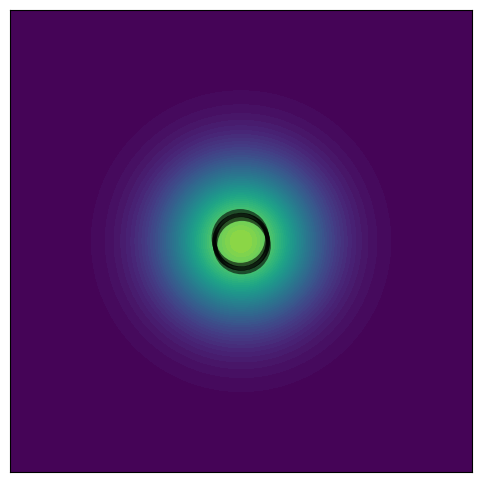} &
\includegraphics[scale=0.3]{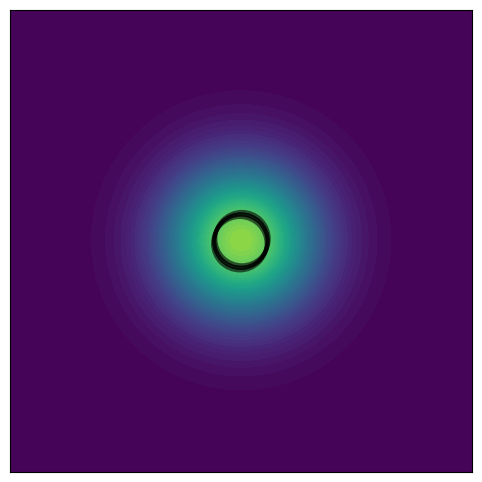} \\

&\multicolumn{5}{c}{\textbf{RLOO (ARITS)}} \\ \cline{2-6}\\
\includegraphics[scale=0.3]{figures/init_sensitivity/target.png} &
\includegraphics[scale=0.3]{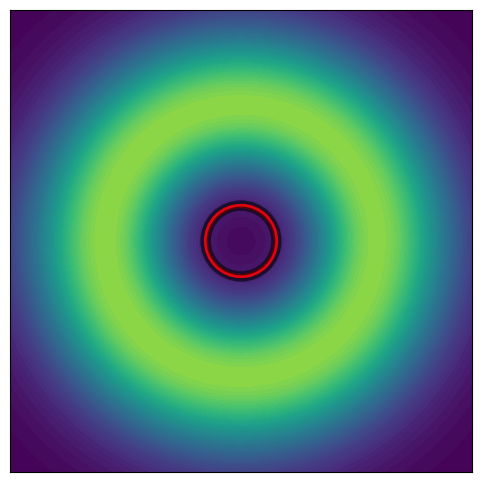} &
\includegraphics[scale=0.3]{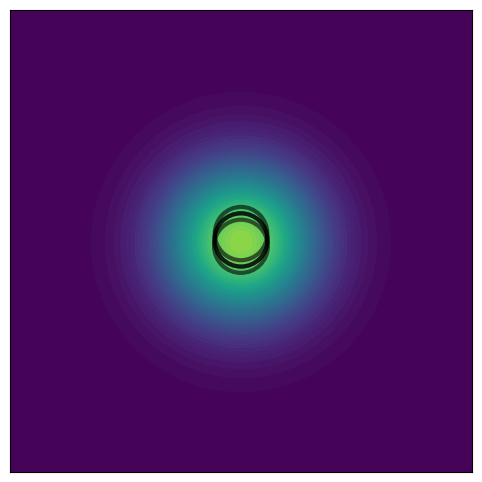} &
\includegraphics[scale=0.3]{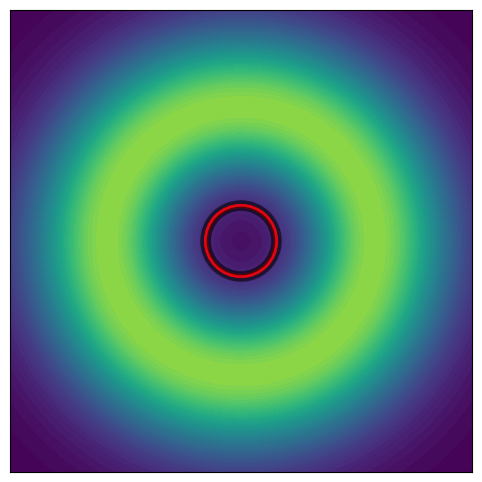} &
\includegraphics[scale=0.3]{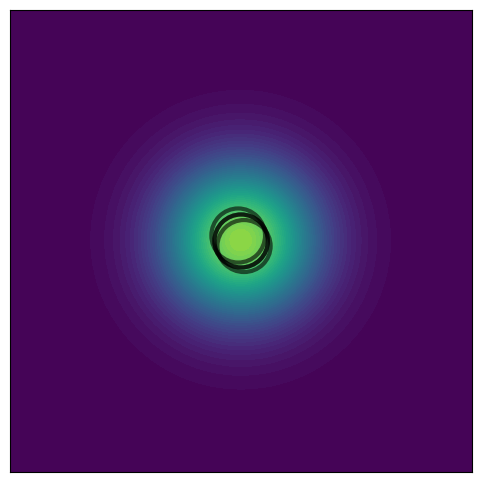} &
\includegraphics[scale=0.3]{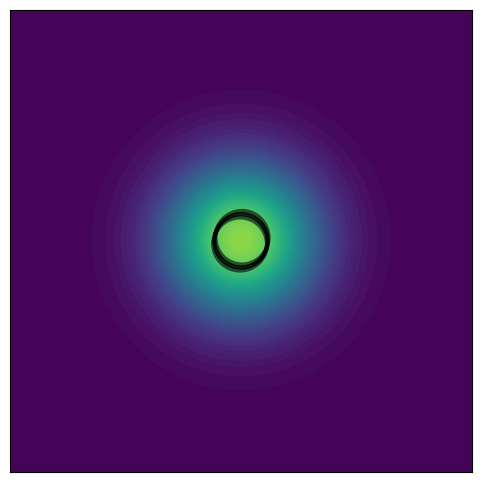} \\
\bottomrule
\end{tabular}
}
\caption{\textbf{The quality of the variational approximation can be sensitive to the initialization.} The figure shows learned models for $5$ different initializations for the \emph{Hollow} target in $16$ dimensions. 
The components are sketched as ellipses with width and height corresponding to their standard deviations.
Positively weighted components are illustrated in black and negative components are shown in red. The first column shows the target. Notably, the initializations look visually similar. Nevertheless, only two of the five RLOO models (with either sampling scheme) converge to a ring-like unnormalized conditional density.}\label{fig:init_sensitivity}
\end{figure*}

As mentioned in the main text, our SMM models can be quite sensitive to the initialization.
We show this behavior on the example of the \emph{Hollow} target with $d=16$ in \Cref{fig:init_sensitivity}.
The figure shows the $5$ different initializations generated in our experiments and the learned models for $\Delta\text{VI}$ and the RLOO variants.
All models were trained with a sampling budget of $S=10^5$ and a learning rate of $0.01$ with no weight decay.
We show the best checkpoint based on training divergence.
We create two-dimensional visualizations of the density by setting all variables except the first two to a fixed value of $5$ and evaluating the density as a function of the first two variables.

\subsection{RQ3: IS with SMMs}
In this section, we provide results for IS with SMM proposals. We group the section into two research questions: (\textbf{RQ3.1}) tackles the effect of a safe component on \dis{} using synthetic proposals whereas (\textbf{RQ3.2}) uses learned proposals from RQ2 and compares various estimation strategies.

\subsubsection{RQ3.1: Safe $\Delta\text{IS}$}\label{app:rq_31}
We study the effect of a safe component on the \emph{DeepRing} and \emph{Hollow} targets. See \Cref{app:targets} for their specification. 
Note that these targets are squared SMMs and we have access to their ground-truth parameters.
To create proposals that are close to the optimal UIS proposal, which in this case is simply $p$ \citep{robert1999monte,mcbook}, we gently noise the standard deviations of the Gaussian components of $p$ while keeping the remaining parameters fixed. %
Let $\boldsymbol{\sigma}_p$ denote the vector consisting of the standard deviations of $p$ \emph{before squaring} and let $|\boldsymbol{\sigma}_p|$ denote its size. 
We obtain a corresponding vector $\boldsymbol{\sigma}_q$ as follows
\begin{align*}
\boldsymbol{Z} &\sim \mathcal{N}(0, I^{(|\boldsymbol{\sigma}_p|)})\\
\boldsymbol{\sigma}_q &:= \boldsymbol{\sigma}_p \cdot \exp(0.01\cdot \boldsymbol{Z}).
\end{align*}
The resulting vector $\boldsymbol{\sigma}_q$ is then used to realize the synthetic proposal $q_{\text{SMM}}$.
We then mix the proposal with a safe component and assess whether this improves estimation quality.
The full proposal is therefore given as
$$q(\vx) = (1-\beta) \cdot q_{\text{SMM}}(\vx) + \beta \cdot q_{\text{safe}}(\vx),$$
where $\beta \in [0, 1)$ is a hyperparameter.

We realize the corresponding estimator for $\mathbb{E}_p[f]$ as follows
\begin{align*}
\widehat{I}_{\Delta\text{IS}}^{(\text{safe})} =~ & (1-\beta) \left[ \frac{Z_+}{Z}  \frac{1}{S_{+}} \sum\nolimits_{s=1}^{S_{+}}  f(\vx_{+}^{(s)}) \frac{p(\vx_{+}^{(s)})}{q(\vx_{+}^{(s)})}
- \frac{Z_-}{Z} \frac{1}{S_{-}} \sum\nolimits_{s=1}^{S_{-}} f(\vx_{-}^{(s)})\frac{p(\vx_{-}^{(s)})} {q(\vx_{-}^{(s)})}\right] \\
& + \beta \left[ \frac{1}{S_{\text{safe}}} \sum\nolimits_{s=1}^{S_{\text{safe}}} f(\vx_{-}^{(s)})\frac{p(\vx_{\text{safe}}^{(s)})} {q(\vx_{\text{safe}}^{(s)})}\right],
\end{align*}
where $\vx_{+} \sim q_+$,  $\vx_{-} \sim q_-$,  $\vx_{\text{safe}} \sim q_{\text{safe}}$, and each set of samples is i.i.d. Note that the first part of the estimator corresponds to a standard $\Delta\text{IS}$ estimator in structure, but with the crucial difference of having the combined proposal $q$ in the denominator of the IS weight as opposed to just $q_{\text{SMM}}$. Similarly, the second term is a weighted IS estimator using samples from the safe component but evaluating the full $q$ in the IS weight. In practice, we split the total sampling budget such that $S:=S_{\Delta\text{IS}} + S_{\text{safe}}$ as $S_{\Delta\text{IS}} = \lfloor (1-\beta) S\rfloor$ and $S_{\text{safe}} = \lfloor \beta S\rfloor$ and then distribute $S_{\Delta\text{IS}}$ across $q_+$ and $q_-$ as described in \Cref{sec:approximate_inference}.

\begin{figure}[t]
\resizebox{\textwidth}{!}{
\begin{tabular}{cc}
DeepRing $(D=2)$ & Hollow ($D=16)$\\
\includegraphics[scale=0.3]{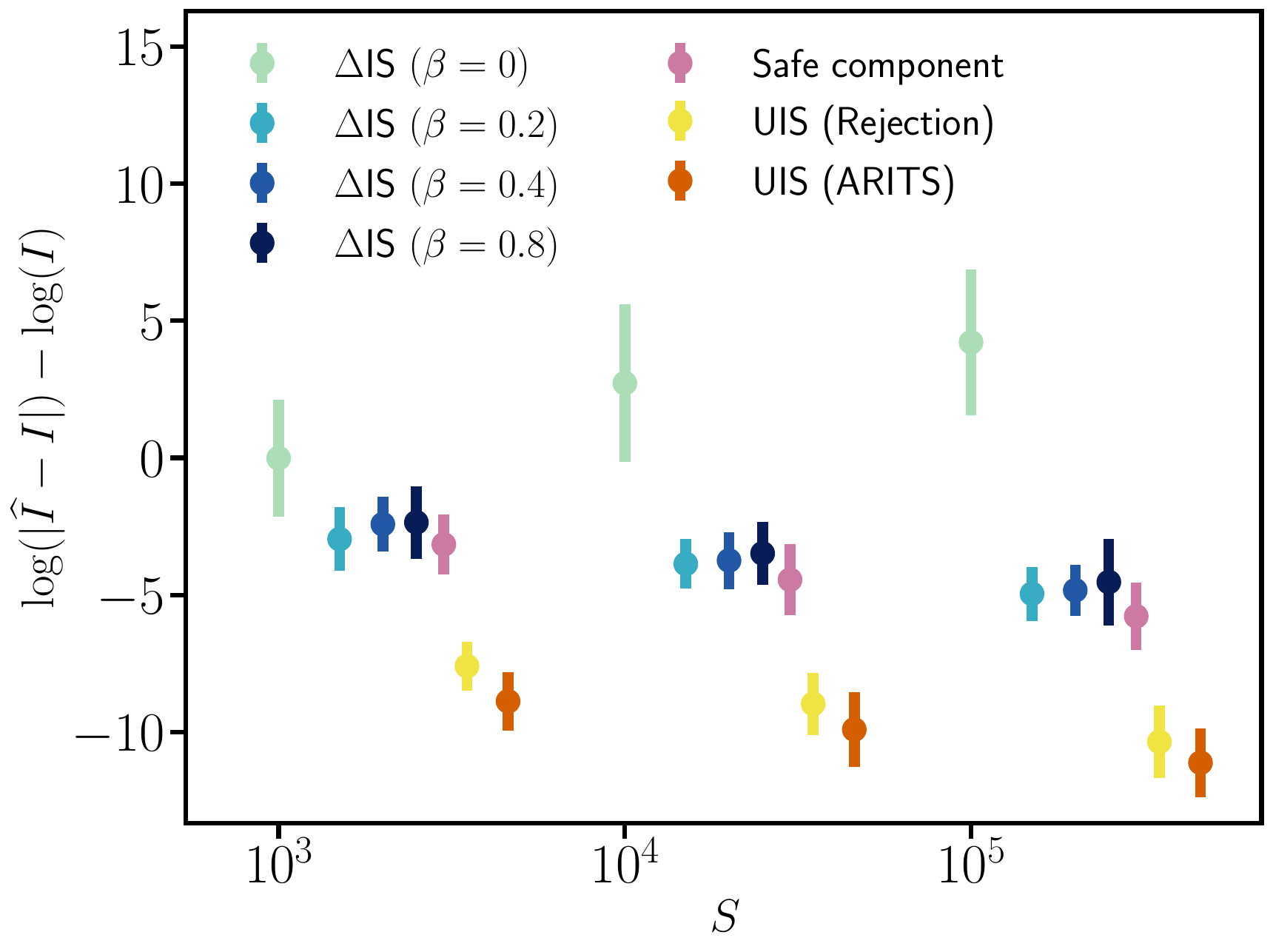} &
\includegraphics[scale=0.3]{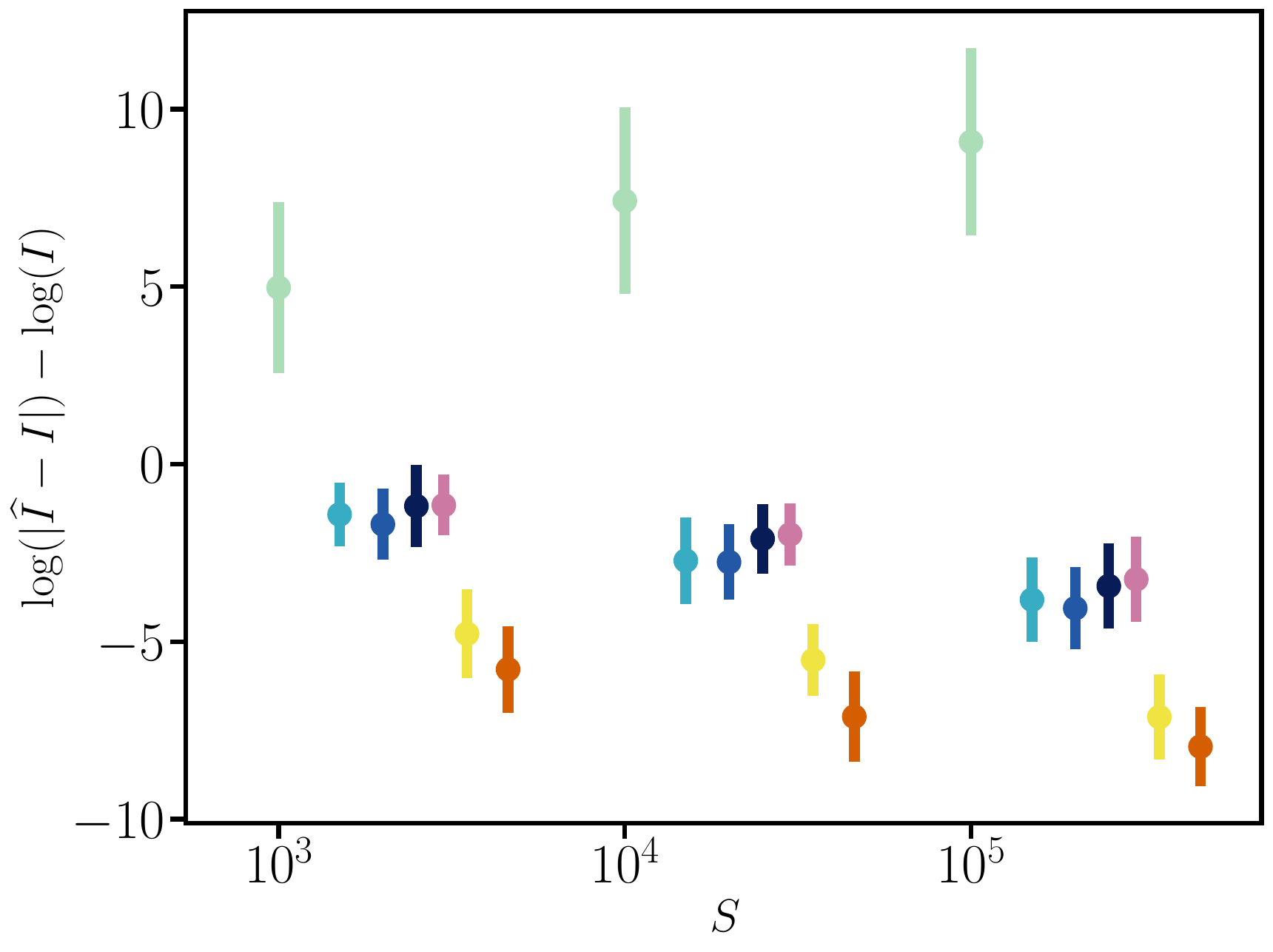}\\
Hollow $(D=32)$& Hollow $(D=64)$\\
\includegraphics[scale=0.3]{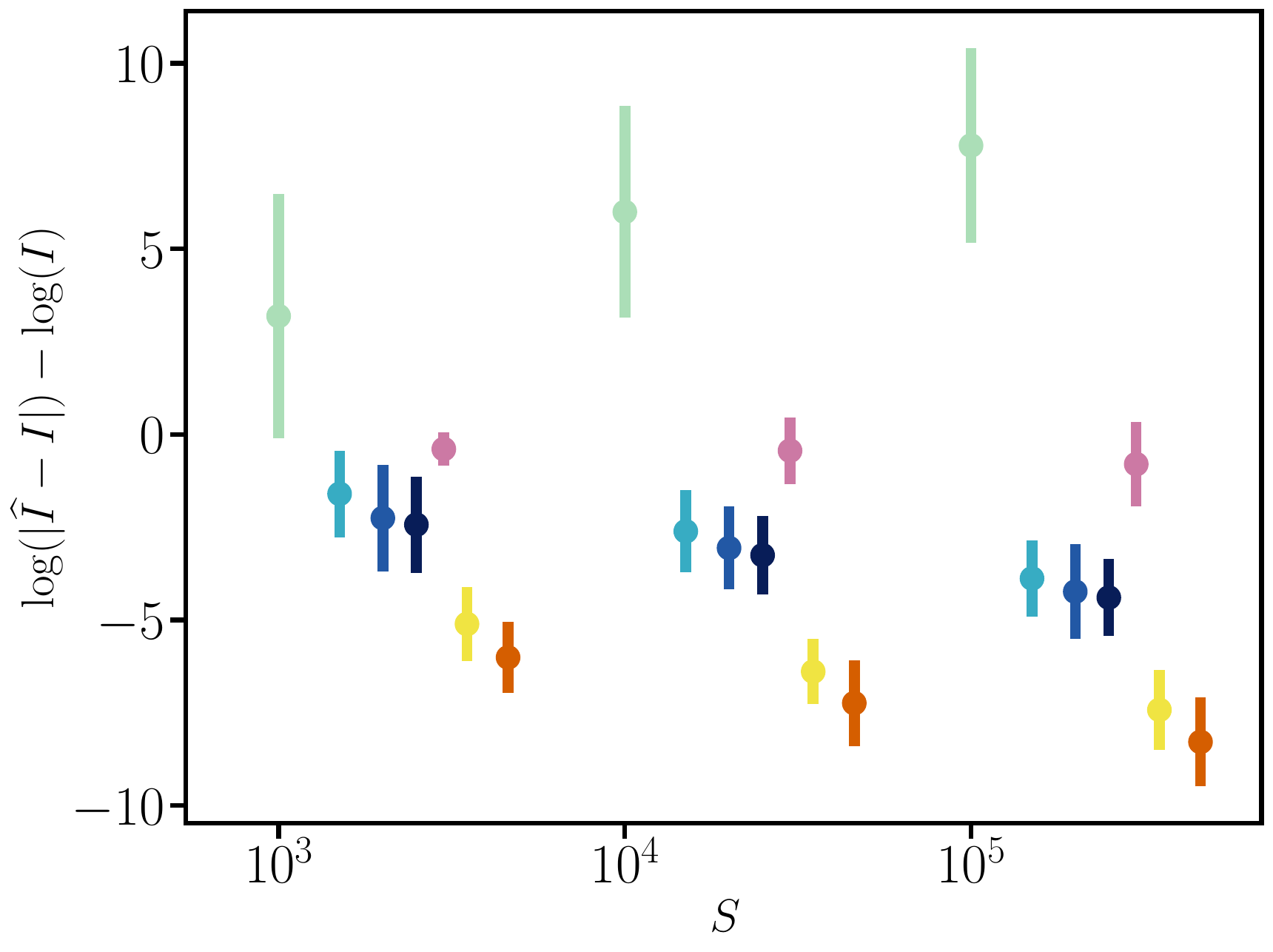} &
\includegraphics[scale=0.3]{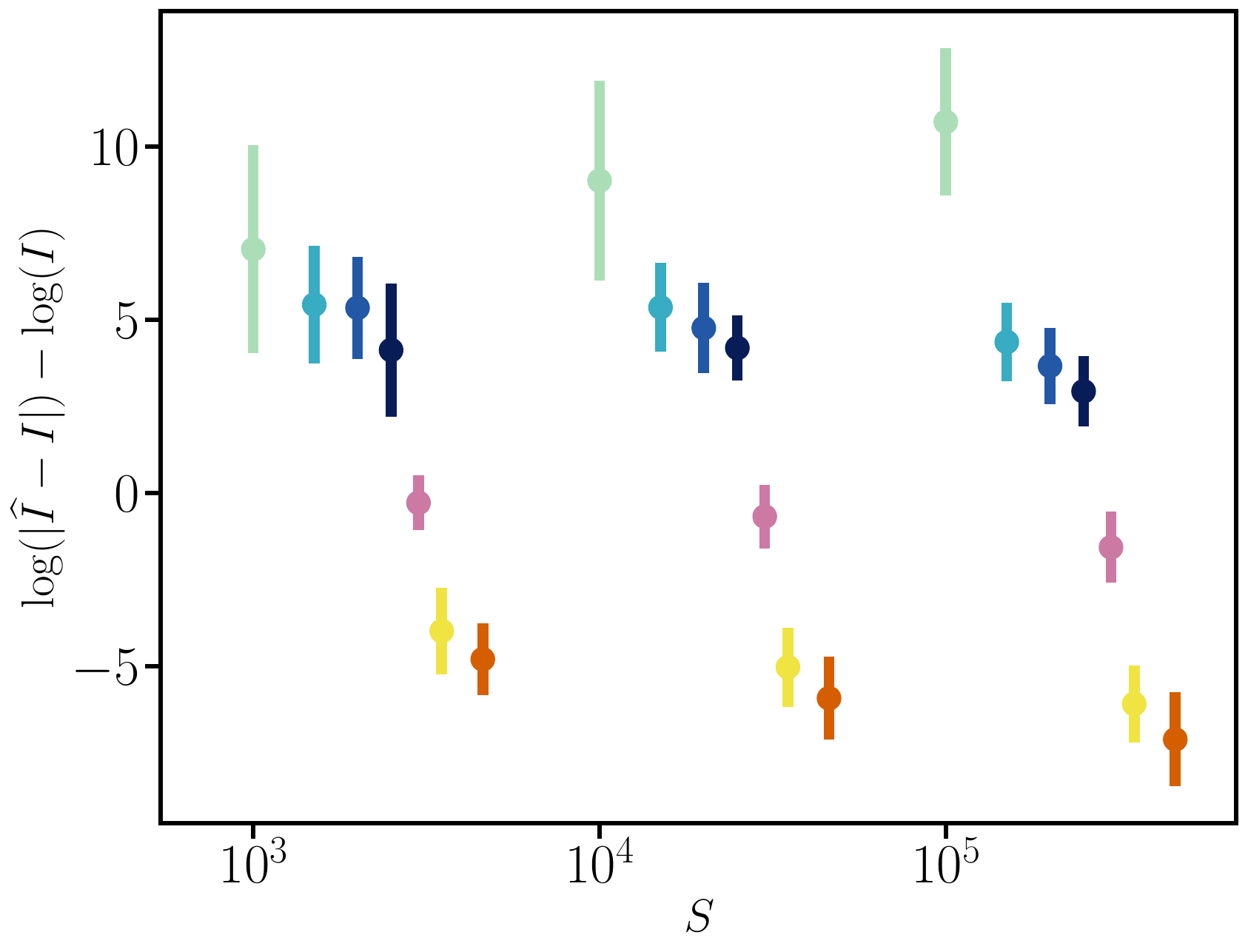}\\
\end{tabular}
}
\caption{\textbf{A safe component can mitigate the potentially large variance of \dis.} Depicted is the estimation error, $\log(|\widehat{I}-I|)-\log(I)$, averaged over $100$ repeated estimations (mean $\pm$ stddev) for various sampling budgets $S$. \emph{Lower is better.} Without a safe component (i.e., $\beta=0$), $\Delta\text{IS}$ can result in high variance and the average estimation error gets \emph{worse} as the sampling budget increases. Standard UIS estimators do not share this behavior despite using the same proposal. Mixing the proposal with a safe component stabilizes the variance of $\Delta\text{IS}$ but does not match the performance of the UIS estimators in these examples.}\label{fig:safe_component}
\end{figure}

In our experiments, we heuristically choose the safe component as a zero-mean isotropic Gaussian with a comparatively high standard deviation for the target. In the experiments below, we use $\sigma=3$ for the \emph{DeepRing} target and $\sigma=8$ for the \emph{Hollow} targets. 
The estimation quality is measured by computing $\log(|\widehat{I} -I|)-\log(I)$ over $100$ repetitions.
\Cref{fig:safe_component} summarizes the results. 
For all targets, $\Delta\text{IS}$ initially results in high variance and atypical behavior for an MC estimator: The average estimation error \emph{increases} as the sampling budget increases.
A possible explanation could be that the noising process sketched above creates ill-defined proposals.
However, we only observe this behavior for $\Delta\text{IS}$: The standard UIS estimators based on rejection and ARITS achieve both (1) better average estimation error than $\Delta\text{IS}$ and (2) a steady decrease in estimation error with increased sampling budget, as is to be expected. %
Adding a safe component stabilizes the $\Delta\text{IS}$ estimator.
Interestingly, the performance is not very sensitive to the value of $\beta$ in the range \{0.2, 0.4, 0.8\}.
Moreover, for the \emph{DeepRing} target and the \emph{Hollow} target in $64$ dimensions, using the safe component in isolation gives better estimates than any of the $\Delta\text{IS}$ estimators.
This is surprising, as the safe component in isolation should be a worse fit to the target than any of the mixed $\Delta\text{IS}$ proposals.
As a result, it seems like the mixed $\Delta\text{IS}$ proposals are a suboptimal choice for the sampling distribution induced by $\Delta\text{IS}$.
Possibly, a more principled safe component that is fit for the specific SMM proposal would improve estimation with $\Delta\text{IS}$.

\subsubsection{RQ3.2: IS with learned proposals}\label{app:rq_32}
In this section, we evaluate the the estimation quality achieved by the IS schemes discussed in \Cref{sec:approximate_inference} using learned proposals.
The integral of interest is the normalizing constant of the target density, i.e., $I =\int \widetilde{p}(\vx)d\vx.$
We use the models selected for \Cref{tab:rq_1} and \Cref{tab:high_dim} as proposals.
For SMMs, we use the proposals trained via RLOO with rejection sampling since this learning scheme generally offered a good compromise between approximation quality and computational efficiency.
For $\Delta\text{IS}$, we perform a grid search of $\beta \in \{0, 0.1,0.2,\ldots, 0.9\}$ and $\sigma_{safe} \in \{3, 5, 7, 9\}$ and select the best setting based on the empirical variance of the estimator with $10^4$ samples based on $30$ repetitions.%
We use a sampling budget of $S=10^4$ for all estimators.
We repeat the estimation $100$ times and report the average estimation error, measured as $\log(|I - \widehat{I}|)-\log(I)$, and its standard deviation across the repetitions.

\Cref{tab:rq_32} summarizes the results. Our main takeawys match the observations made throughout the paper: When a GMM is sufficient to model the target well, as is the case for the \emph{Funnels} and \emph{GMM3} and \emph{GMM4}, using a GMM proposal gives the best results on normalizing constant estimation.
For targets that require an SMM for finding a good proposal, i.e., the \emph{Ring} and the \emph{Hollow} targets, using an SMM proposal with either rejection sampling or ARITS can result in better normalizing constant estimation than using a GMM.
Interestingly, $\Delta\text{IS}$ tends to perform worse than a simple GMM even on targets that benefit from negative components.
For most targets, no safety was chosen for $\Delta\text{IS}$ by the gird search we perform to select $\beta$ and $\sigma_{\text{safe}}$ (see \Cref{tab:chosen_safety}).
How to construct and select a better safe component for a given proposal is an interesting open question.

\begin{table*}[!htbp]
\caption{Selected parameters for the safe component of \dis{} for \Cref{tab:rq_32} and \Cref{fig:rq32}.}\label{tab:chosen_safety}
\begin{center}
\begin{tabular}{lll}
\toprule
Target & $\beta$ & $\sigma_{\text{safe}}$\\ \midrule
GMM3 $(D=2)$ & 0 & /\\
GMM4 $(D=2)$ & 0 & /\\
Funnel $(D=2)$ & 0 & / \\
Ring $(D=2)$ & 0 & / \\
Hollow $(D=16)$ & 0 & /\\
Hollow $(D=32)$ & 0 & /\\
Hollow $(D=64)$ & 0 & /\\
Funnel $(D=10)$ & 0.1 & 3\\\bottomrule
\end{tabular}
\end{center}
\end{table*}

\begin{figure*}[ht!]
\resizebox{\textwidth}{!}{
\begin{tabular}{ccc}
  \multicolumn{1}{c}{GMM3 $(d=2)$} & \multicolumn{1}{c}{GMM4 $(d=2)$} & \multicolumn{1}{c}{Funnel $(d=2)$}\\
\includegraphics[scale=0.4]{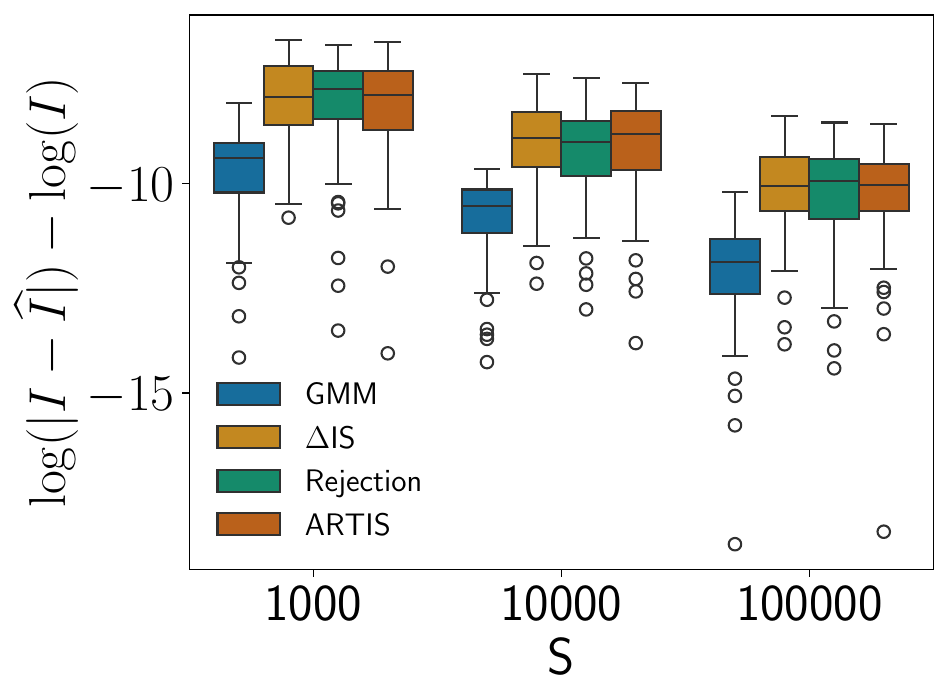} &
\includegraphics[scale=0.4]{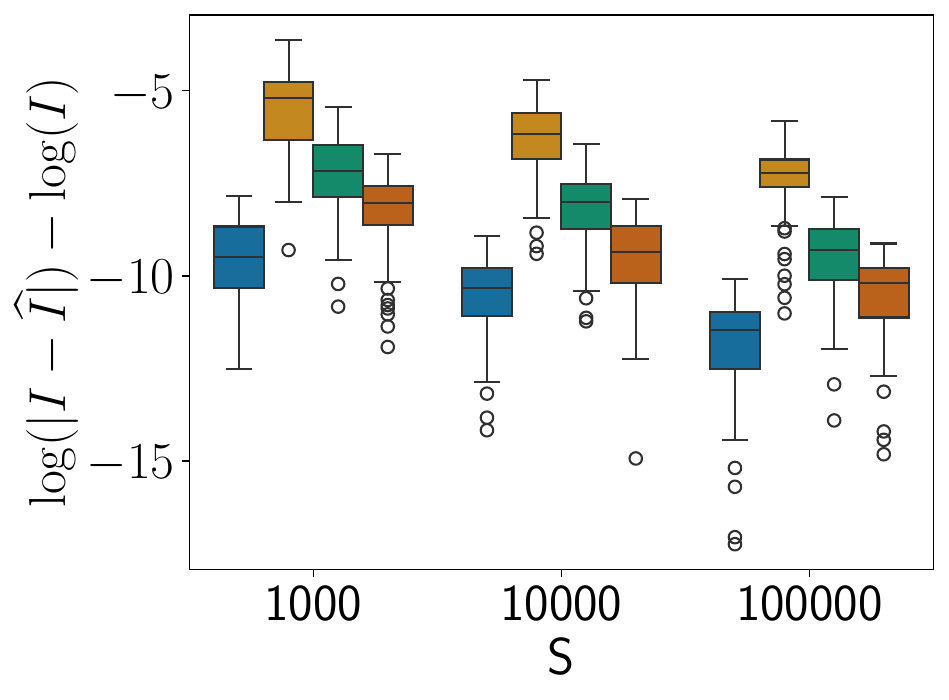} &
\includegraphics[scale=0.4]{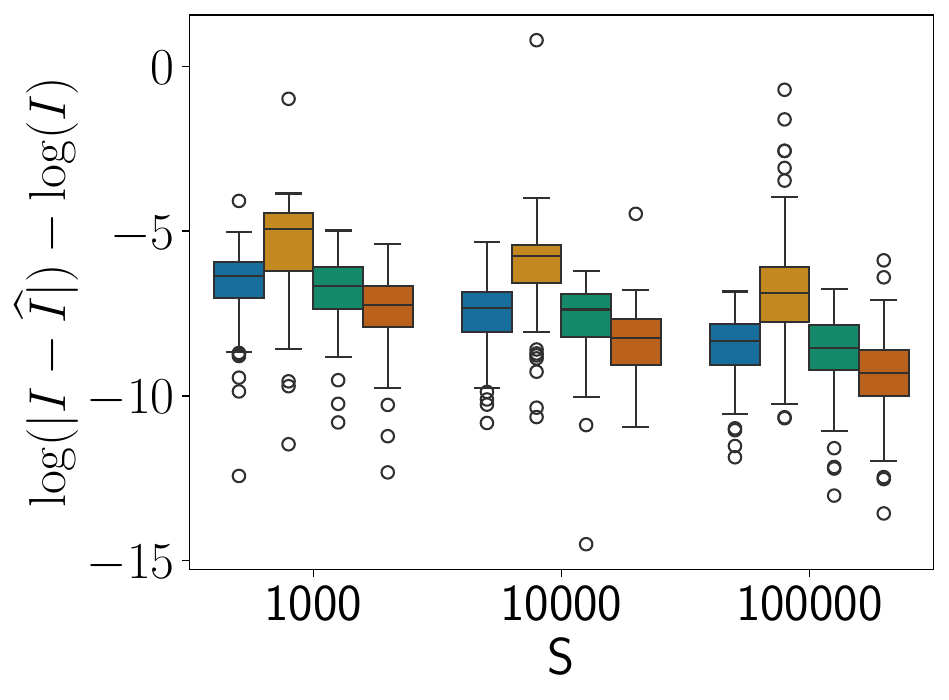} \\
\multicolumn{1}{c}{Ring $(d=2)$} & \multicolumn{1}{c}{Hollow $(d=16)$} & \multicolumn{1}{c}{Hollow $(d=32)$} \\
\includegraphics[scale=0.4]{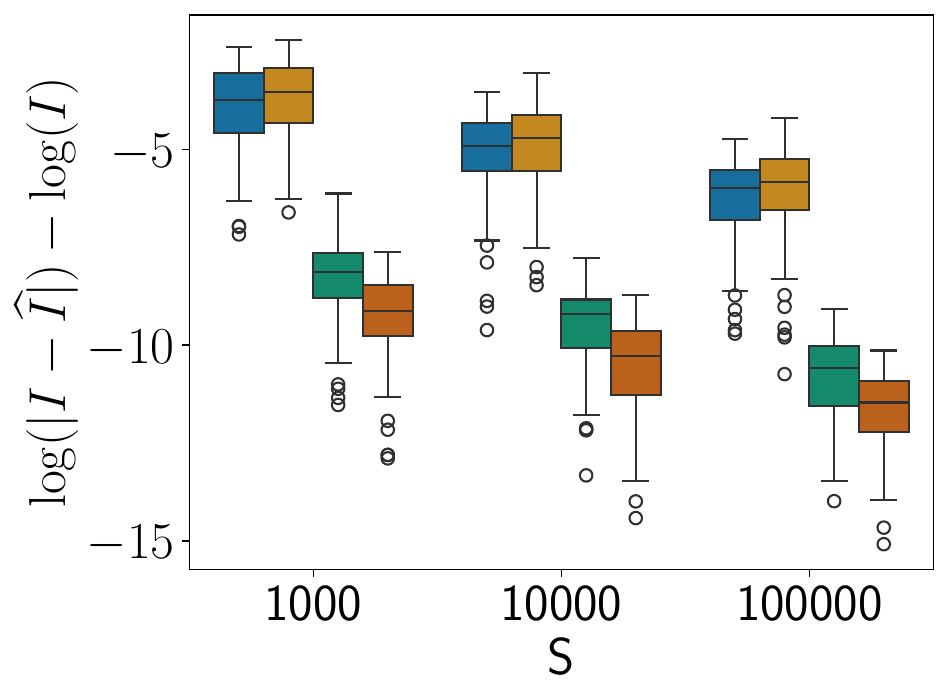} &
\includegraphics[scale=0.4]{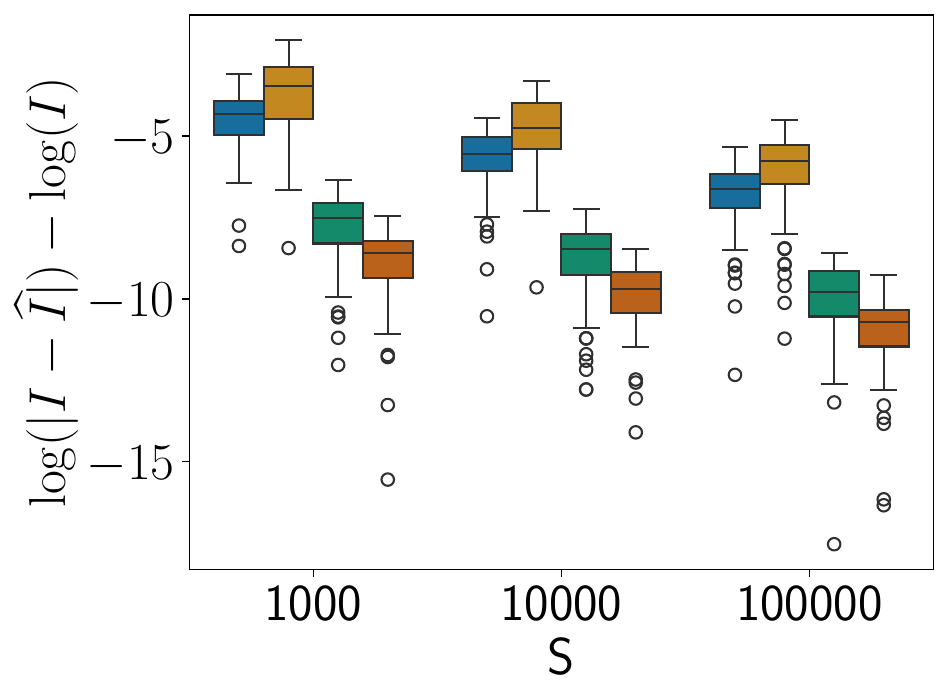} &
\includegraphics[scale=0.4]{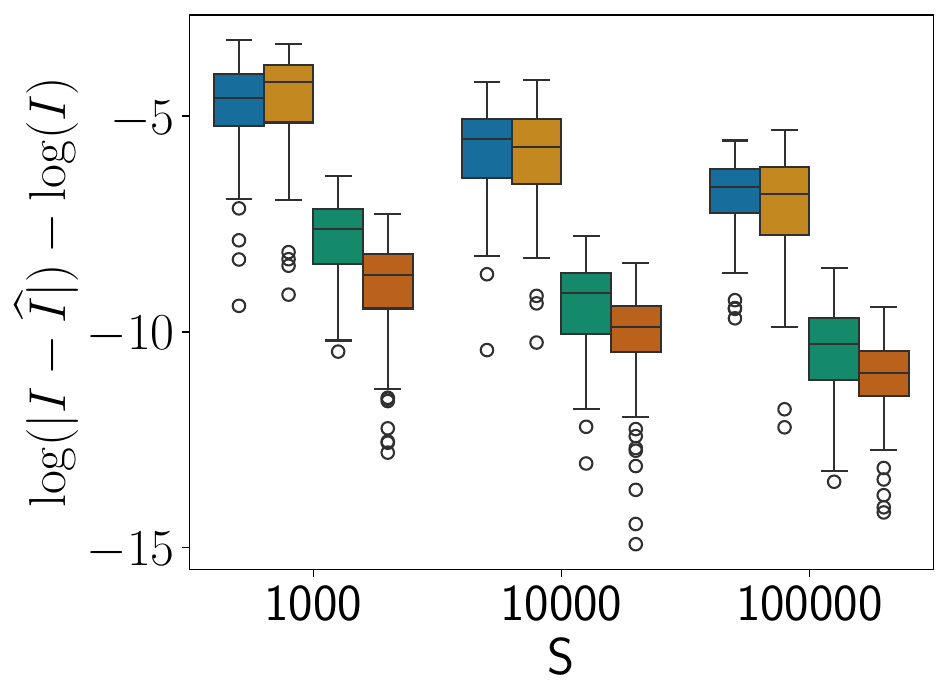}\\
\multicolumn{1}{c}{Hollow $(d=64)$} & \multicolumn{1}{c}{Funnel $(d=10)$} & \\
\includegraphics[scale=0.4]{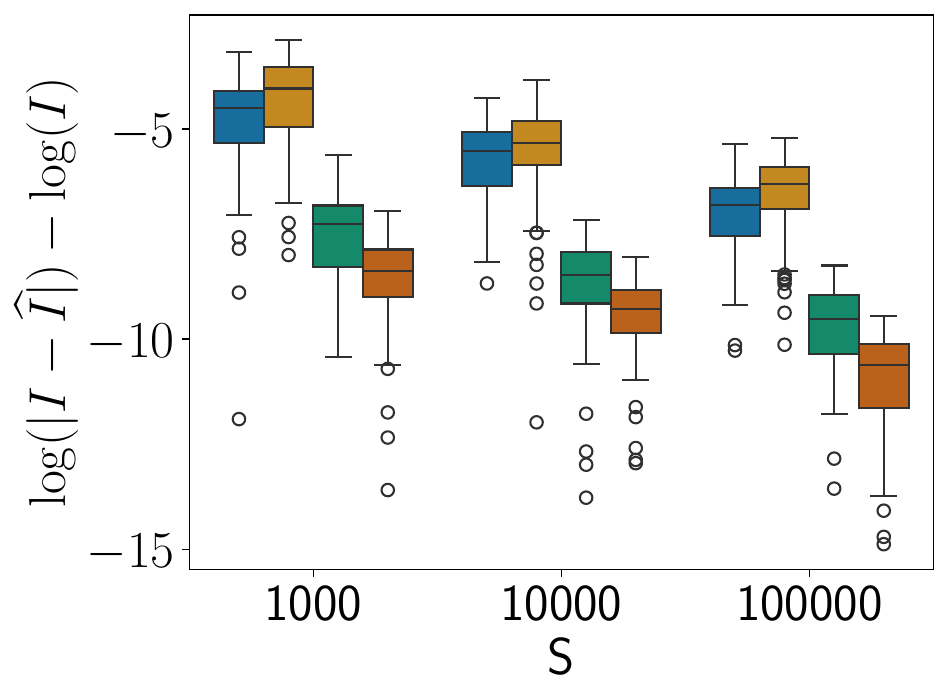} &
\includegraphics[scale=0.4]{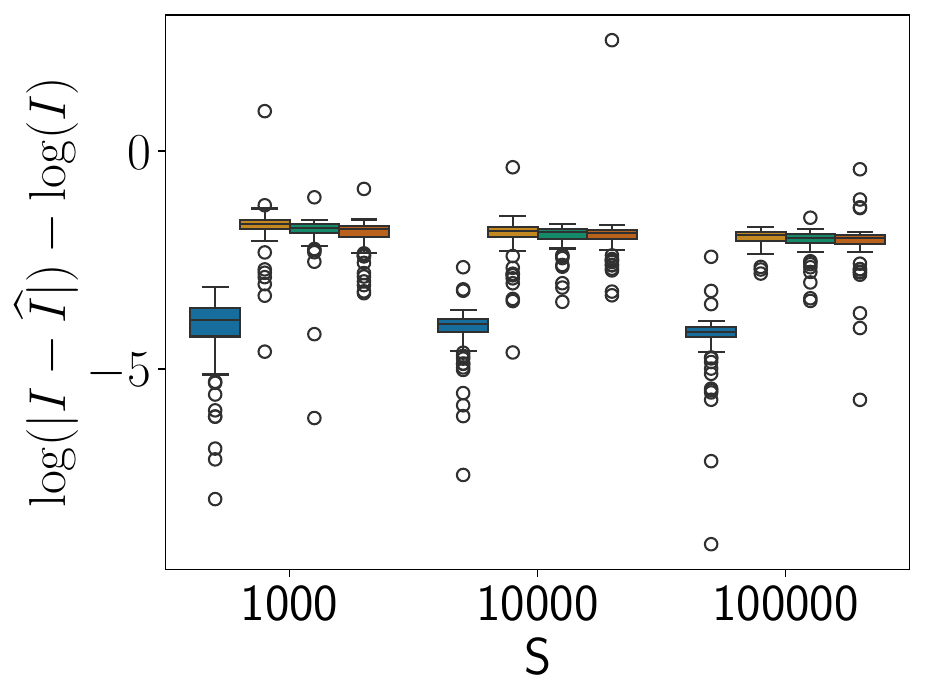}\\
\end{tabular}
}
\caption{\textbf{The trends observed in \Cref{tab:rq_32} hold across various sample sizes.} The boxplots summarize the error for normalizing constant estimation over $100$ repetitions when using GMM and SMM proposals for varying sampling budgets. \emph{Lower is better.} For targets on which the SMM achieves a better fit, we see better estimation performance when using the SMM proposal with either rejection or ARITS. On the remaining targets, GMMs tend to be better. Note that the SMM proposal was always learned with RLOO + rejection.}
\label{fig:rq32}
\end{figure*}

\begin{table}[ht]
\caption{\textbf{SMMs can improve upon GMMs for normalizing constant estimation when an SMM proposal allows for a better fit to the target.} All estimates are computed with a sampling budget of $S=10^{4}$. The estimation error is measured as $\log(|I - \widehat{I}|)-\log(I)$ reported over $100$ repeated estimations (mean $\pm$ standard deviation). The SMM proposals were fit with RLOO using rejection sampling and the GMM proposals were learned via the SELBO \citep{morningstar2021automatic}.}\label{tab:rq_32}
\begin{center}
\resizebox{\textwidth}{!}{
\begin{tabular}{@{}llllll}\toprule
& \multicolumn{1}{c}{UIS (GMM)} & \multicolumn{1}{c}{$\Delta\text{IS}~ \text{(safe)}$} & \multicolumn{1}{c}{UIS (Rej.)} & \multicolumn{1}{c}{UIS (ARITS)} \\  \midrule
GMM3 $(D = 2)$ & $-1.08 \cdot 10^{1} \pm 9.89 \cdot 10^{-1}$ & $-9.04 \cdot 10^{0} \pm 1.03 \cdot 10^{0}$ & $-9.21 \cdot 10^{0} \pm 1.04 \cdot 10^{0}$ & $-9.10 \cdot 10^{0} \pm 1.15 \cdot 10^{0}$ \\
GMM4 $(D = 2)$ & $-1.05 \cdot 10^{1} \pm 1.03 \cdot 10^{0}$ & $-6.34 \cdot 10^{0} \pm 9.29 \cdot 10^{-1}$ & $-8.22 \cdot 10^{0} \pm 1.02 \cdot 10^{0}$ & $-9.55 \cdot 10^{0} \pm 1.16 \cdot 10^{0}$ \\
Funnel $(D = 2)$ & $-7.59 \cdot 10^{0} \pm 1.00 \cdot 10^{0}$ & $-6.09 \cdot 10^{0} \pm 1.38 \cdot 10^{0}$ & $-7.67 \cdot 10^{0} \pm 1.13 \cdot 10^{0}$ & $-8.41 \cdot 10^{0} \pm 1.00 \cdot 10^{0}$ \\
Ring $(D = 2)$ & $-5.13 \cdot 10^{0} \pm 1.16 \cdot 10^{0}$ & $-4.94 \cdot 10^{0} \pm 1.13 \cdot 10^{0}$ & $-9.56 \cdot 10^{0} \pm 1.03 \cdot 10^{0}$ & $-1.05 \cdot 10^{1} \pm 1.13 \cdot 10^{0}$ \\ \midrule
Hollow $(D = 16)$ & $-5.80 \cdot 10^{0} \pm 1.02 \cdot 10^{0}$ & $-4.86 \cdot 10^{0} \pm 1.11 \cdot 10^{0}$ & $-8.85 \cdot 10^{0} \pm 1.22 \cdot 10^{0}$ & $-9.89 \cdot 10^{0} \pm 9.84 \cdot 10^{-1}$ \\
Hollow $(D = 32)$ & $-5.84 \cdot 10^{0} \pm 1.05 \cdot 10^{0}$ & $-5.95 \cdot 10^{0} \pm 1.18 \cdot 10^{0}$ & $-9.36 \cdot 10^{0} \pm 1.06 \cdot 10^{0}$ & $-1.02 \cdot 10^{1} \pm 1.19 \cdot 10^{0}$ \\
Hollow $(D = 64)$ & $-5.83 \cdot 10^{0} \pm 1.02 \cdot 10^{0}$ & $-5.55 \cdot 10^{0} \pm 1.24 \cdot 10^{0}$ & $-8.74 \cdot 10^{0} \pm 1.16 \cdot 10^{0}$ & $-9.46 \cdot 10^{0} \pm 9.42 \cdot 10^{-1}$ \\ \midrule
Funnel $(D = 10)$ & $-4.11 \cdot 10^{0} \pm 5.73 \cdot 10^{-1}$ & $-1.96 \cdot 10^{0} \pm 4.70 \cdot 10^{-1}$ & $-1.95 \cdot 10^{0} \pm 2.91 \cdot 10^{-1}$ & $-1.94 \cdot 10^{0} \pm 5.39 \cdot 10^{-1}$ \\
\bottomrule
\end{tabular}
}
\end{center}
\end{table}

\subsection{RQ2.4: Bayesian Logistic Regressions}
\begin{figure}[t]
  \centering
  \setlength{\tabcolsep}{2pt}
  \begin{tabular}{cc}
    \includegraphics[width=0.35\linewidth]{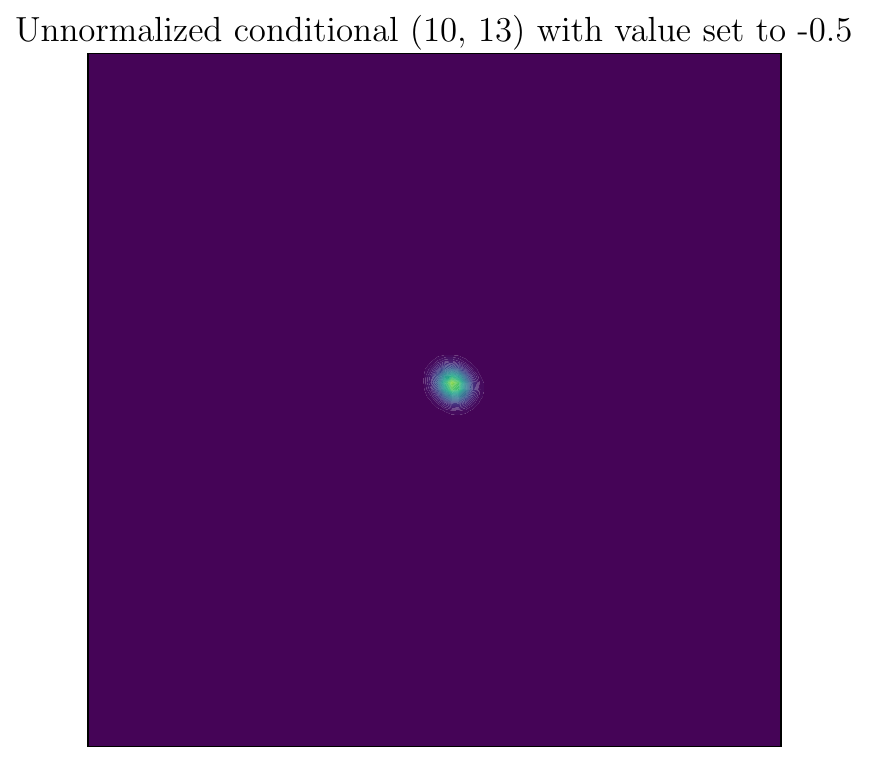} &
    \includegraphics[width=0.35\linewidth]{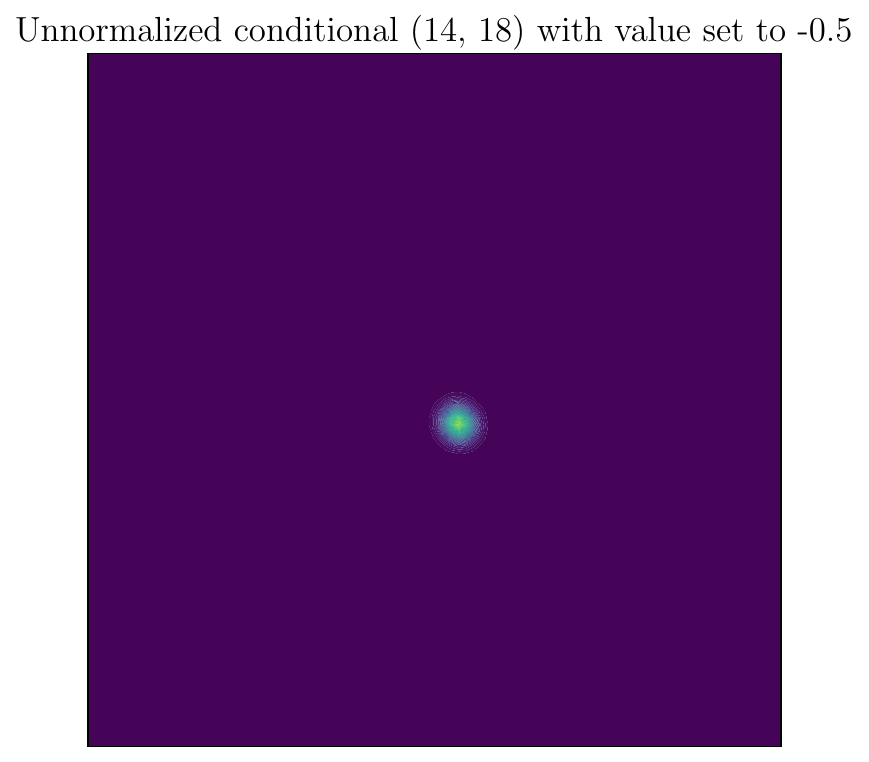} \\
    \includegraphics[width=0.35\linewidth]{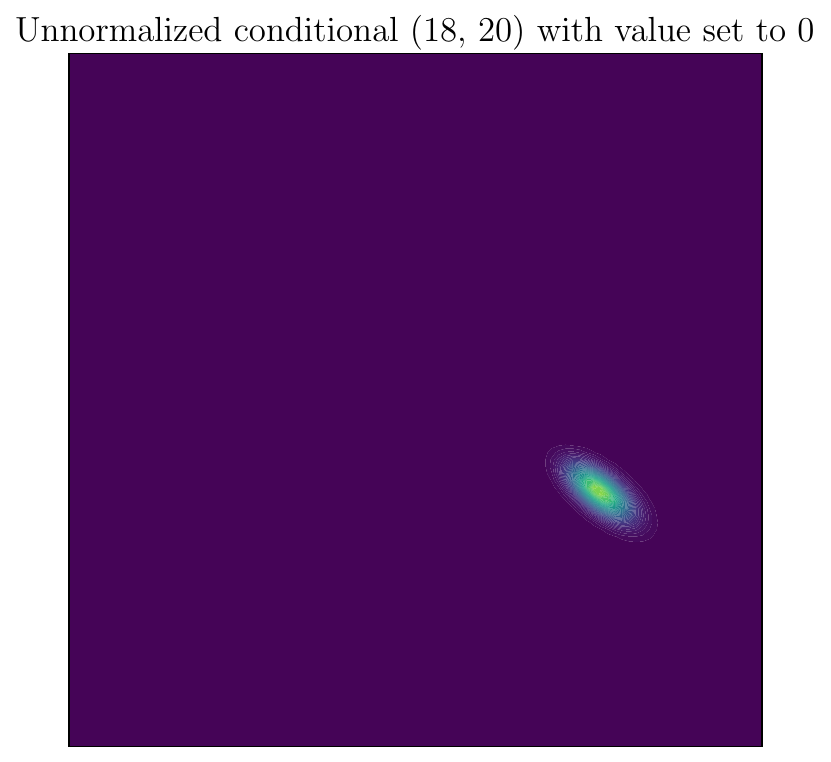} &
    \includegraphics[width=0.35\linewidth]{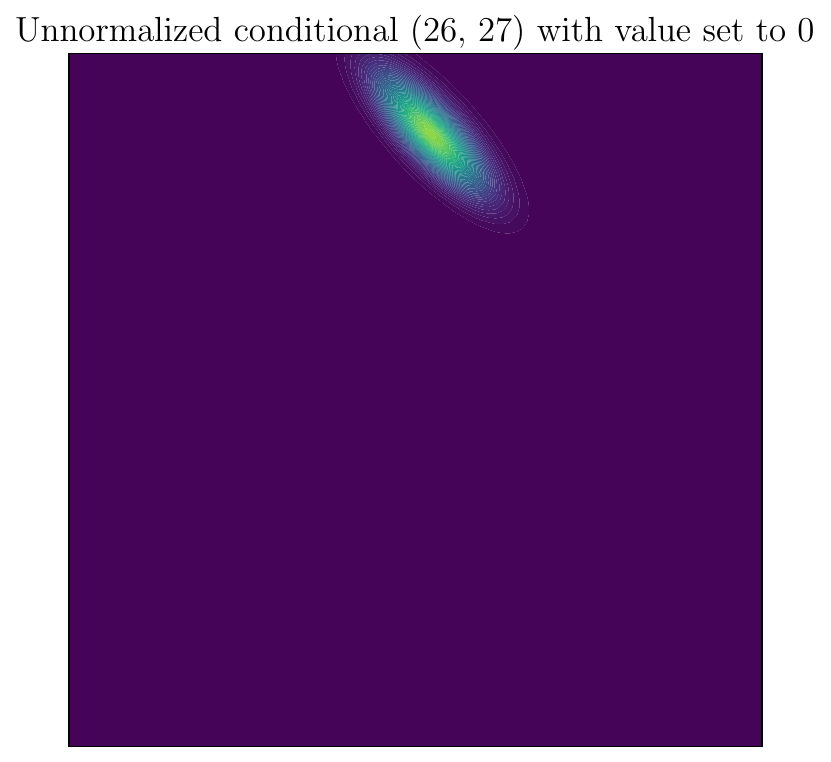} \\
    \includegraphics[width=0.35\linewidth]{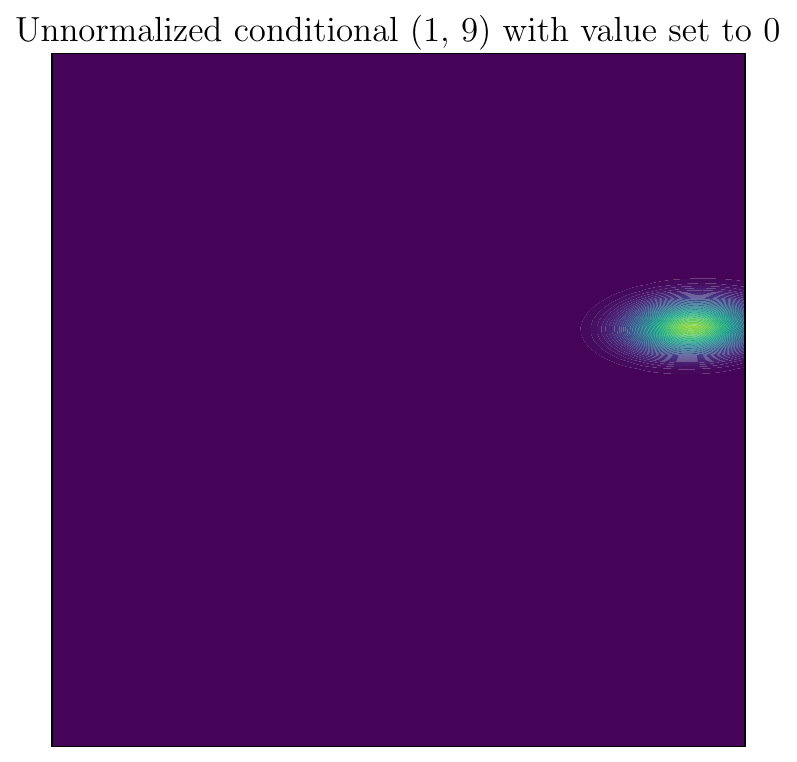} &
    \includegraphics[width=0.35\linewidth]{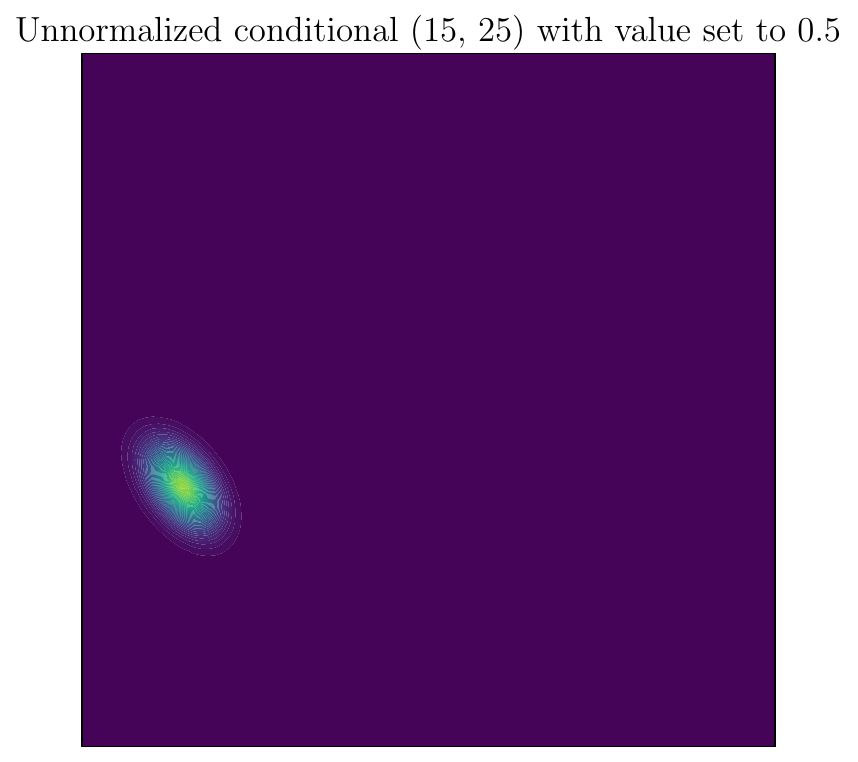}
  \end{tabular}
  \caption{Some unnormalized bivariate conditionals for the GermanCredit (first row), BreastCancer (second row), and Ionosphere (last row) logistic regression targets. Each panel conditions all remaining features on evidence values mentioned in the caption.}
  \label{fig:logreg_conditionals}
\end{figure}

In \cref{fig:logreg_conditionals}, we show plots of some (unnormalized) conditionals for the logistic regression targets. 
Proper marginalization would require sampling with MCMC and or symbolic integration. 
We observe that most plots strongly resemble a simple Gaussian.
As a result, negative parameters might not be beneficial for modeling these targets, which could explain why GMMs and SMMs perform similarly in \cref{tab:logreg_elbo}. The ELBO is calculated using $p(\vx) = \widetilde{p}(\vx)/Z_{\pi}$ where we use a ground truth $Z_{\pi}$ when available and otherwise use $\widetilde{p}(\vx)$. We computed this ground truth for credit and sonar only using the adaptive tempered SMC implementation from the BlackJAX library \citep{cabezas2024blackjax} as for other datasets this proved less reliable - changing SMC hyperparameters led to different results. We used $5 \cdot 10^{5}$ samples (per SMC iteration $t$) and found that using more did not lead to different results. 

\subsection{RQ4: SMMs for Neuro-symbolic Targets}\label{app:rq_4}
We now provide a quantitative comparison between SMMs and GMMs for the neuro-symbolic targets discussed in RQ4). 
We initialize Gaussian components in a grid covering the target and tune their initial standard deviation in $\{1, 2\}$ for \emph{scenario 1} (\cref{fig:pal_fig_1}) and \{2, 4\} for \emph{scenario 2} (\cref{fig:pal_2}). For $K=2$, we set the initial component means to $(3,5)$ and $(6, 5)$ respectively. For \emph{scenario 1} (\cref{fig:pal_fig_1}), we further adjust the bounds for the grid initialization based on $K$ as follows: for $K=4$, we set the limits for $x_1$ as $[2, 4]$ and for $x_2$ as $[3, 7]$; for higher values of $K$, they are set to $[1, 5.5]$ and $[1, 9]$ respectively. For \emph{scenario 2} (\cref{fig:pal_2}), we set the bounds to $[2, 8]$ for $x_1$ and $[1, 7]$ for $x_2$ for $K > 2$.

Mixture weights are initialized as $\frac{1}{K}$ (where is the number of components). 
For SMMs, we also initialize complex weights by sampling from a $\text{Unif}([0,1])$ distribution. 
We train with $3$ random seeds using Adam (lr=$0.01$, weight decay=$0.001$ on mixture weights), a maximum of $15000$ update steps with samples, and a patience of $1000$ on the training loss. 
We select a model as described in \cref{app:rq_1}.

\begin{figure}[ht]
\begin{center}
\resizebox{\textwidth}{!}{
\begin{tabular}{ccccc}
Target & $K=2$ & $K=4$ & $K=8$ & $K=16$\\\midrule
& 
$-1.15 \cdot 10^{0} \pm 4.89 \cdot 10^{-3}$ &
$-6.99 \cdot 10^{-1} \pm 5.92 \cdot 10^{-3}$ &
$-4.86 \cdot 10^{-1} \pm 4.41 \cdot 10^{-3}$ &
$-2.70 \cdot 10^{-1} \pm 2.72 \cdot 10^{-3}$\\
\includegraphics[scale=0.2]{figures/01_fig_1/fig_1_target.pdf}&
\includegraphics[scale=0.2]{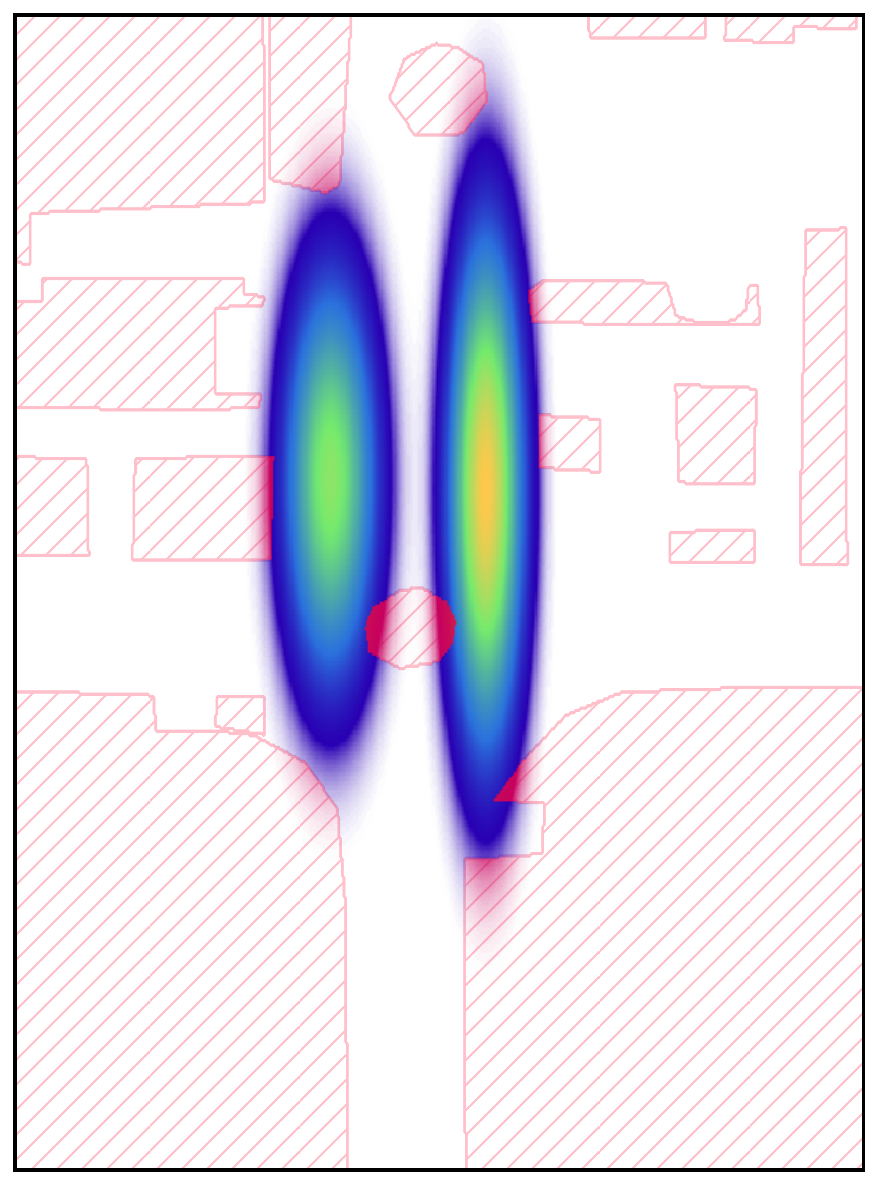}
&
\includegraphics[scale=0.2]{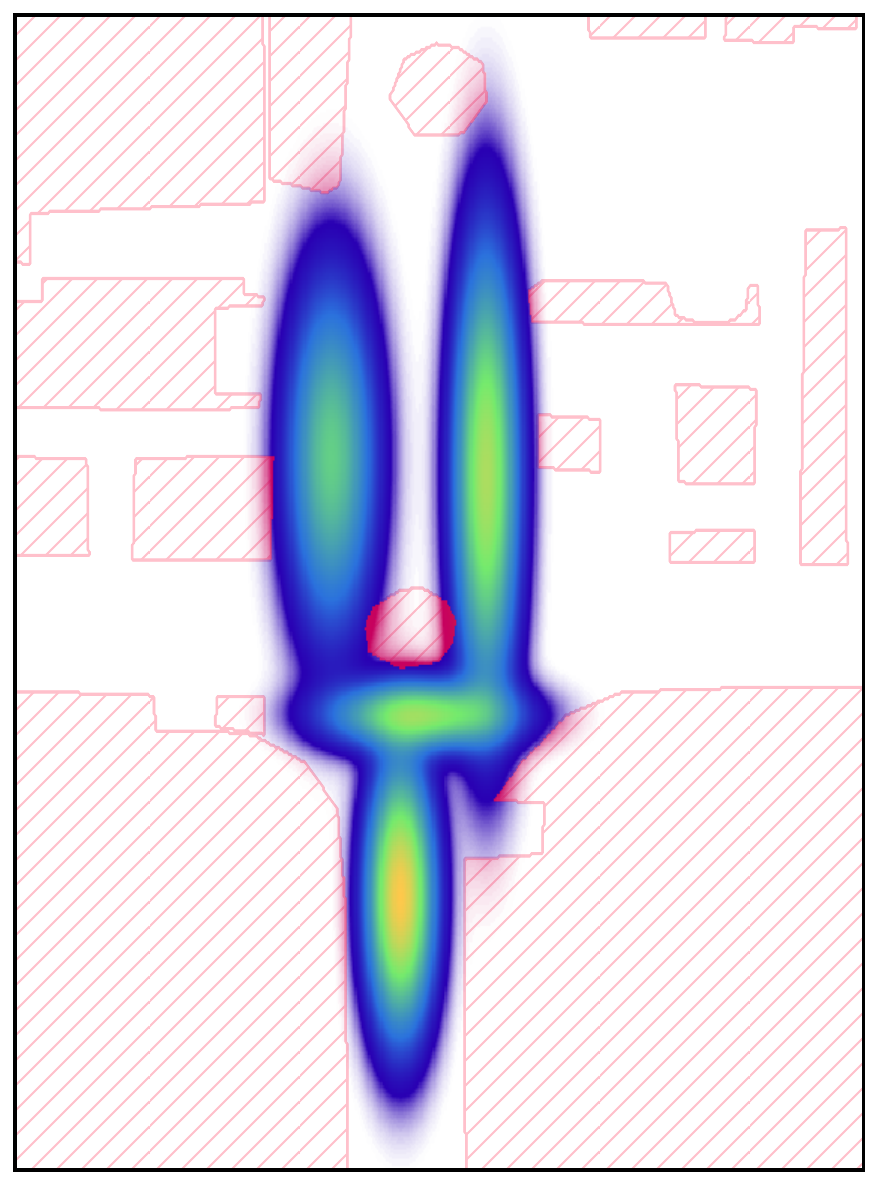}
&
\includegraphics[scale=0.2]{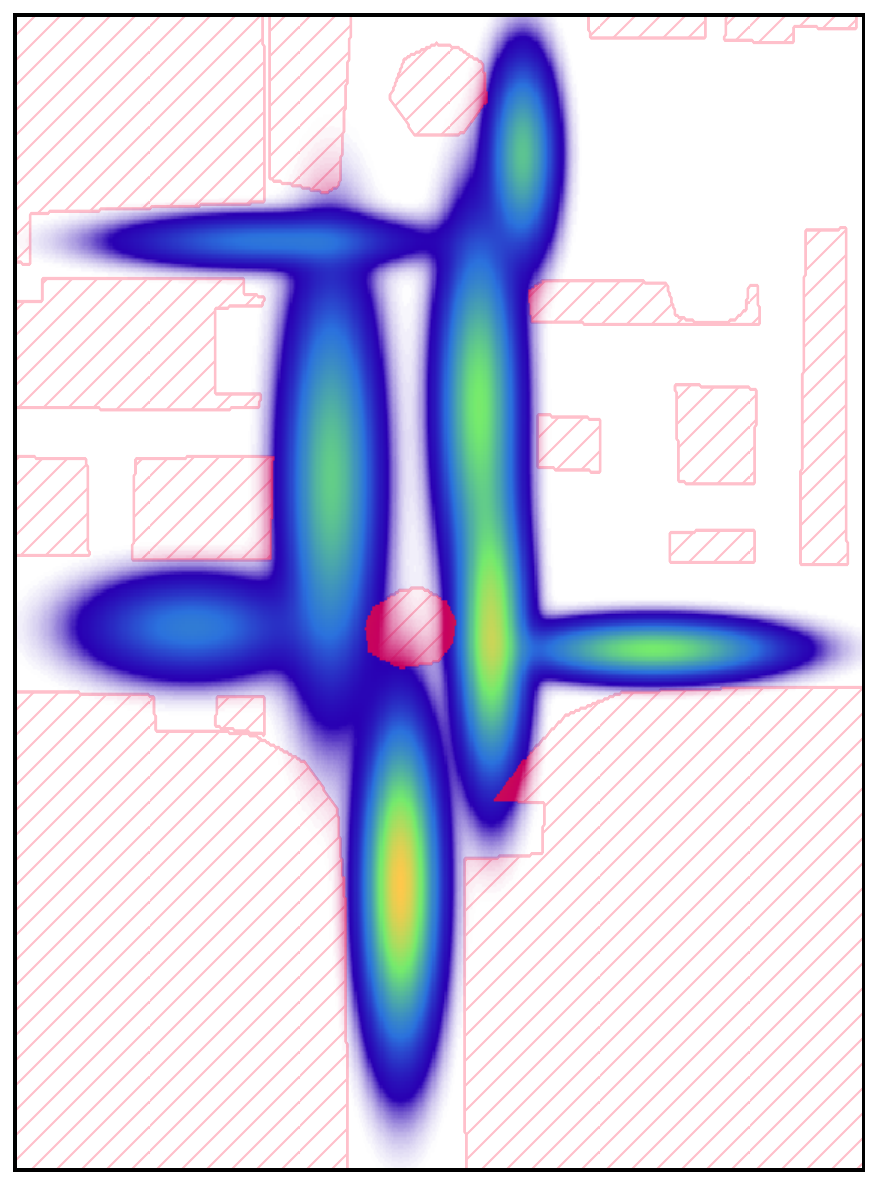}
&
\includegraphics[scale=0.2]{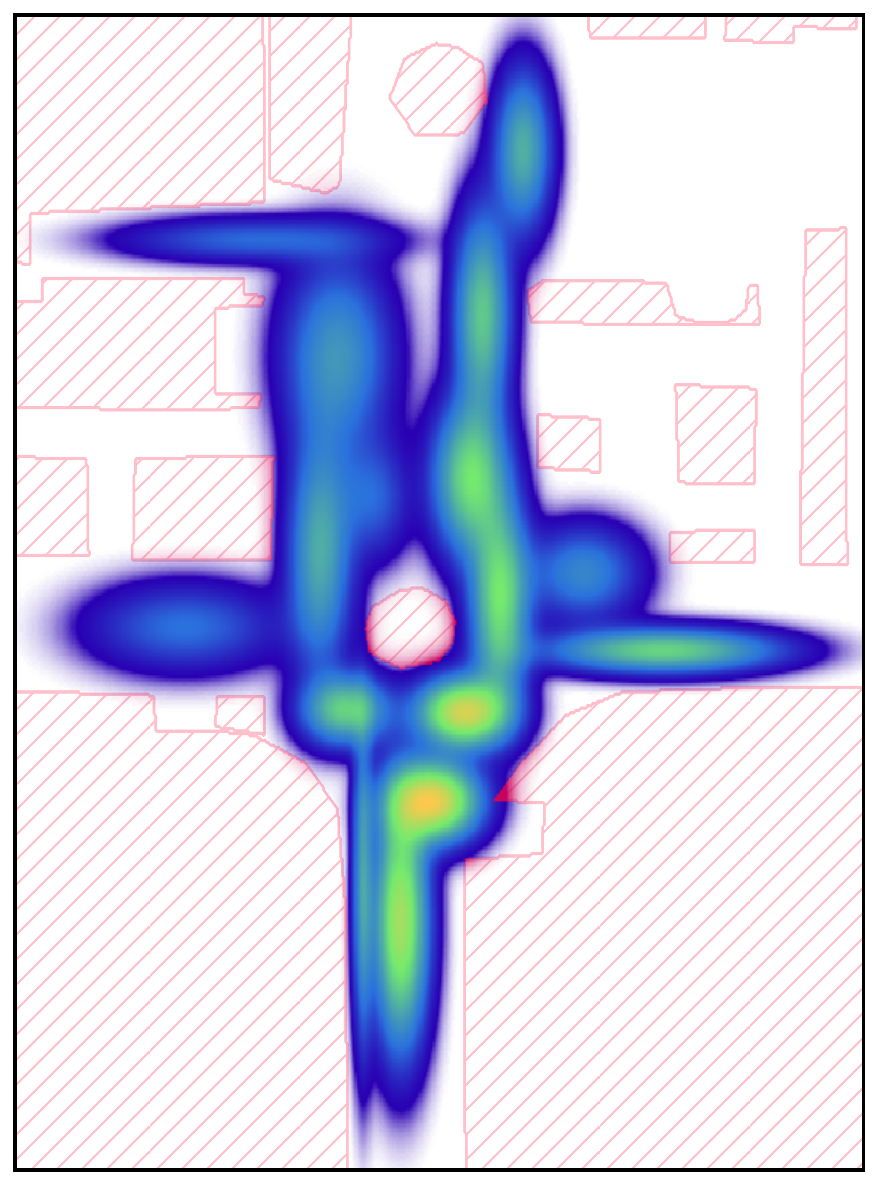}\\
&
$-1.14 \cdot 10^{0} \pm 6.95 \cdot 10^{-3}$ & %
$-6.00 \cdot 10^{-1} \pm 6.13 \cdot 10^{-3}$ & %
$-4.45 \cdot 10^{-1} \pm 5.31 \cdot 10^{-3}$ &
$-2.69 \cdot 10^{-1} \pm 2.66 \cdot 10^{-3}$\\
\includegraphics[scale=0.2]{figures/01_fig_1/fig_1_target.pdf}
& \includegraphics[scale=0.2]{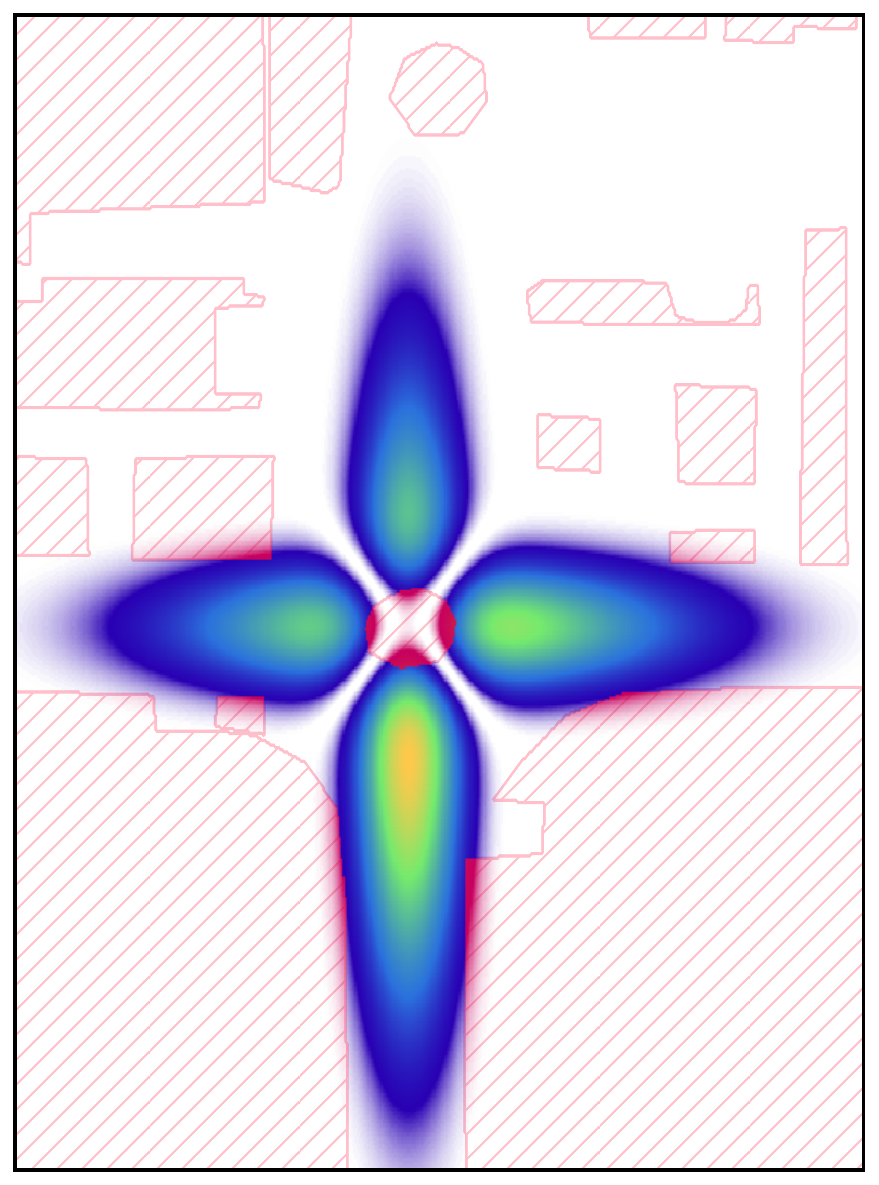}
&
\includegraphics[scale=0.2]{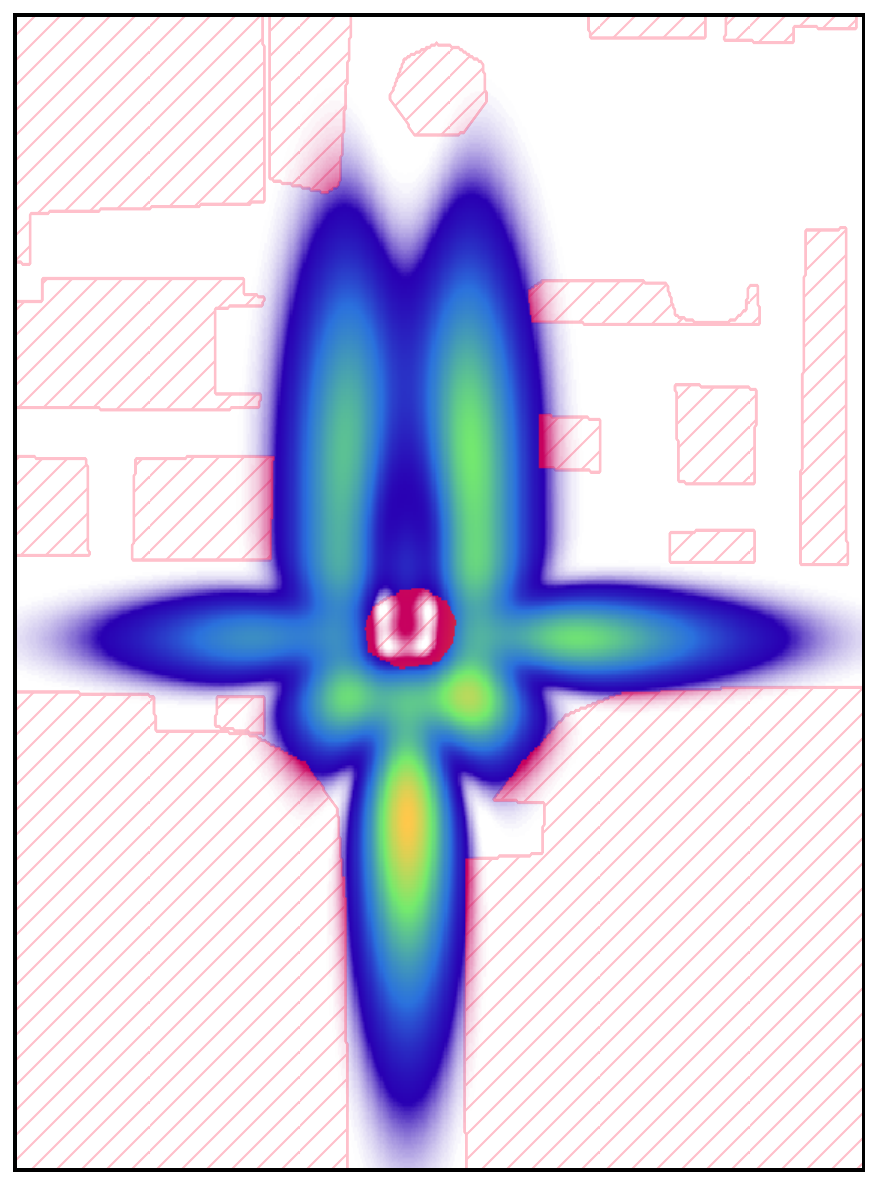}
&
\includegraphics[scale=0.2]{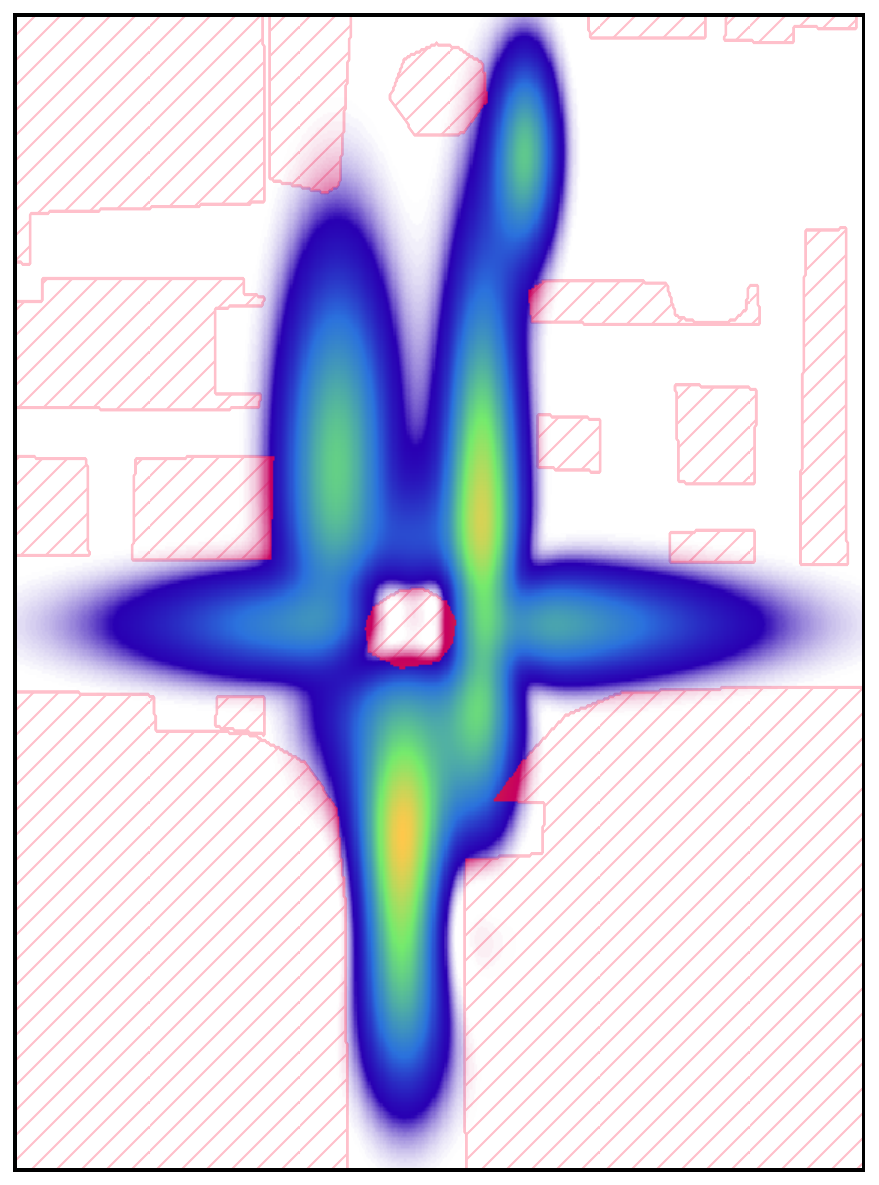}
&
\includegraphics[scale=0.2]{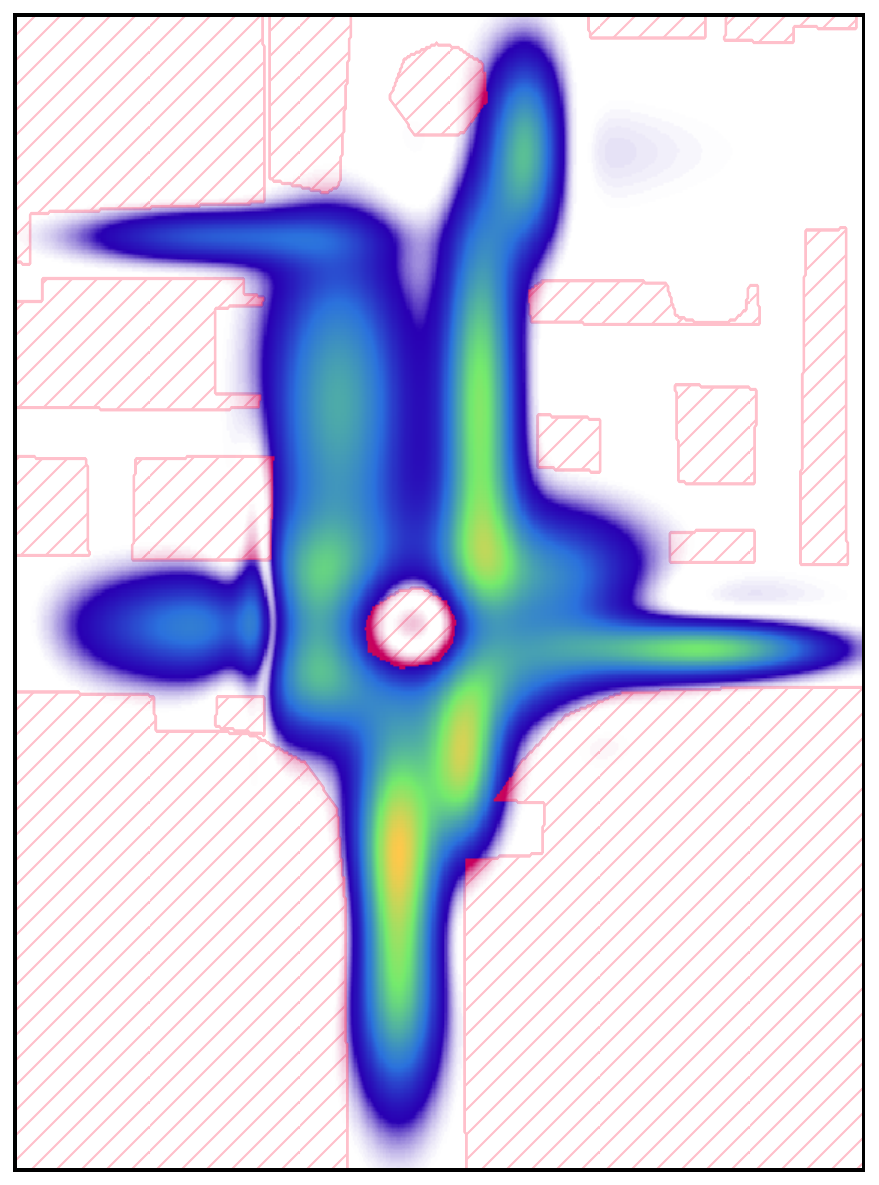}\\
Target & $K=2$ & $K=4$ & $K=8$ & $K=16$\\\midrule
& $-1.16 \cdot 10^{0} \pm 3.70 \cdot 10^{-3}$
& $-8.75 \cdot 10^{-1} \pm 5.49 \cdot 10^{-3}$
& $-5.75 \cdot 10^{-1} \pm 5.14 \cdot 10^{-3}$
& $-3.22 \cdot 10^{-1} \pm 4.29 \cdot 10^{-3}$\\
\includegraphics[scale=0.2]{figures/pal_app/scenario_2/pal_target_2.pdf}
& \includegraphics[scale=0.2]{figures/pal_app/scenario_2/selected_pal_2_gmm_2_polished.pdf}
& \includegraphics[scale=0.2]{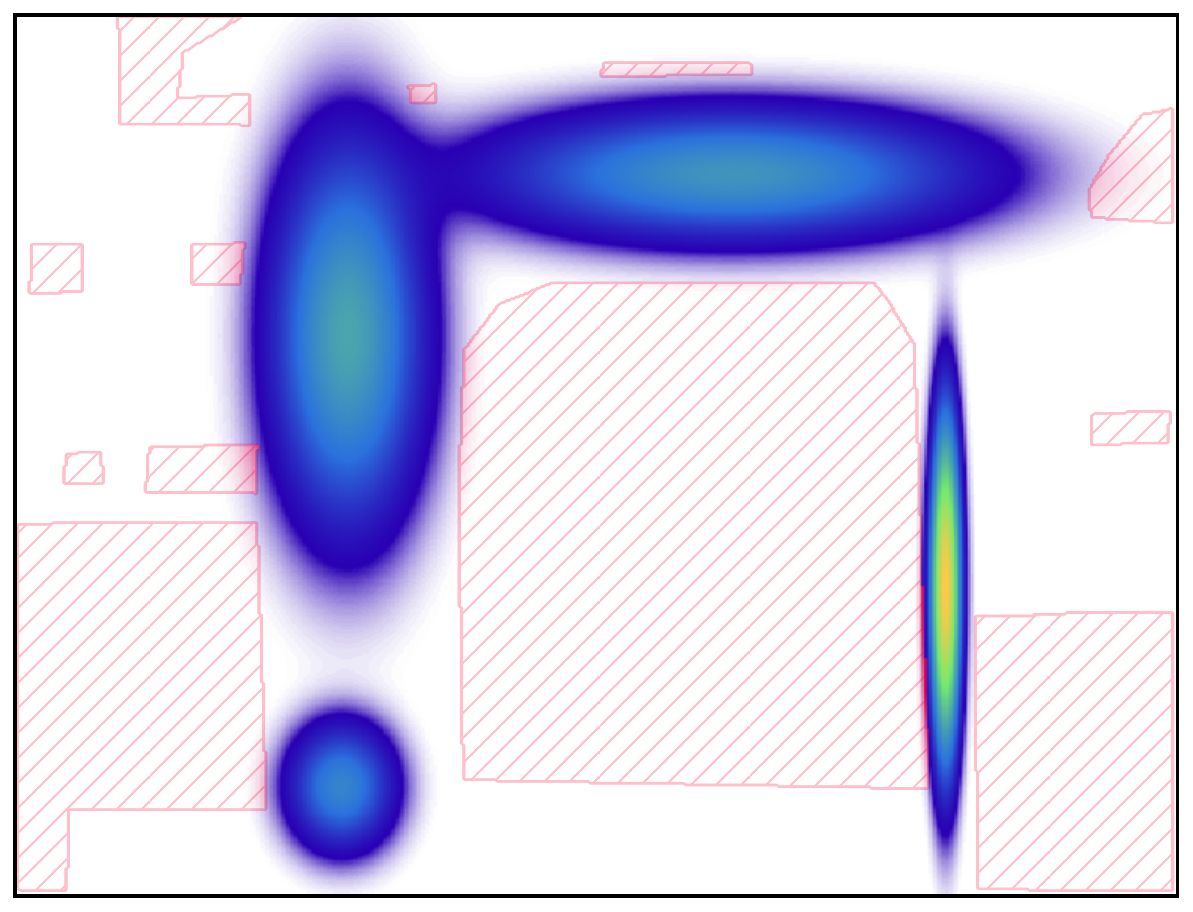}
&
\includegraphics[scale=0.2]{figures/pal_app/scenario_2/selected_pal_2_gmm_8_polished.pdf}
&
\includegraphics[scale=0.2]{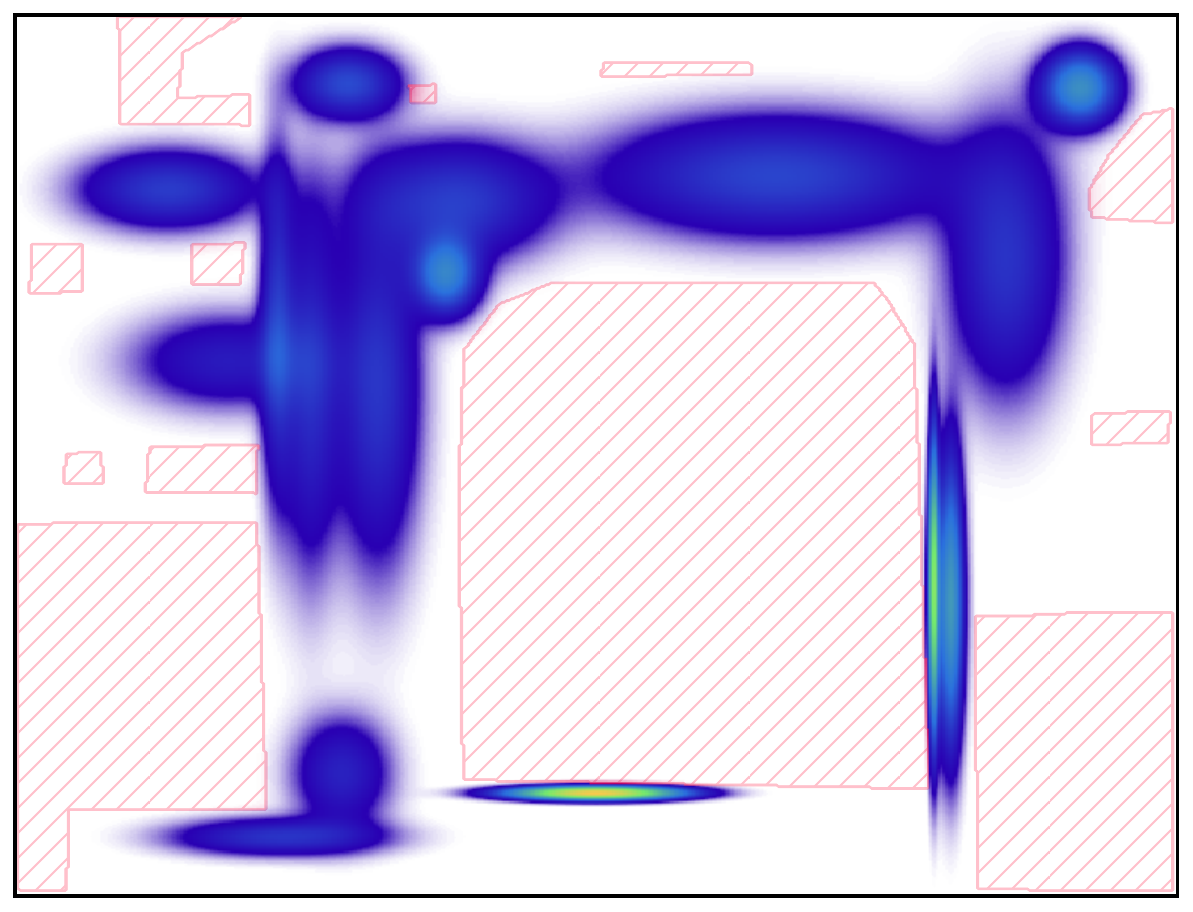}\\
& 
$-1.63 \cdot 10^{0} \pm 7.99 \cdot 10^{-3}$
& $-1.08 \cdot 10^{0} \pm 3.95 \cdot 10^{-3}$ %
& $-7.26 \cdot 10^{-1} \pm 4.68 \cdot 10^{-3}$ %
& $-5.93 \cdot 10^{-1} \pm 3.93 \cdot 10^{-3}$ %
 \\
\includegraphics[scale=0.2]{figures/pal_app/scenario_2/pal_target_2.pdf}
&
\includegraphics[scale=0.2]{figures/pal_app/scenario_2/selected_pal_2_smm_2_polished.pdf}
&
\includegraphics[scale=0.2]{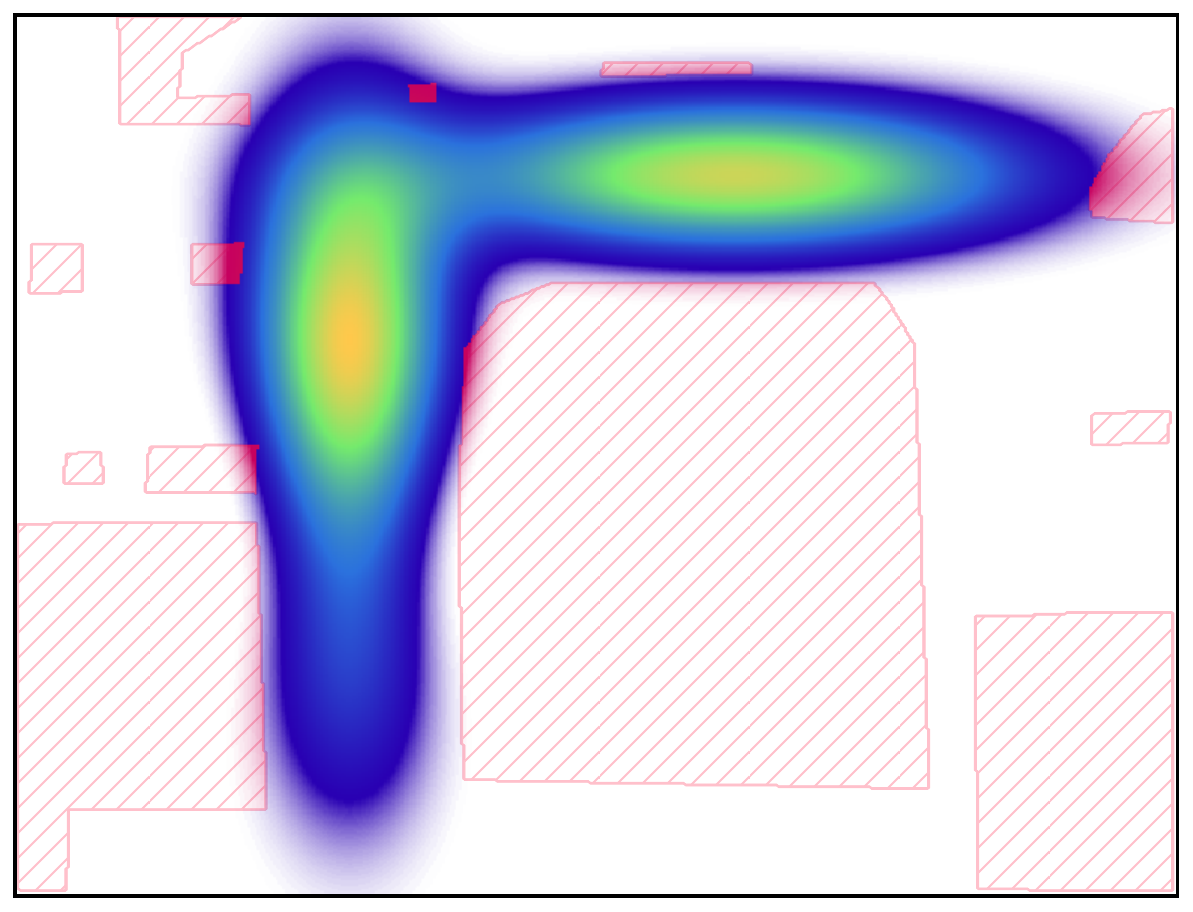}
&
\includegraphics[scale=0.2]{figures/pal_app/scenario_2/selected_pal_2_smm_8_polished.pdf}
&
\includegraphics[scale=0.2]{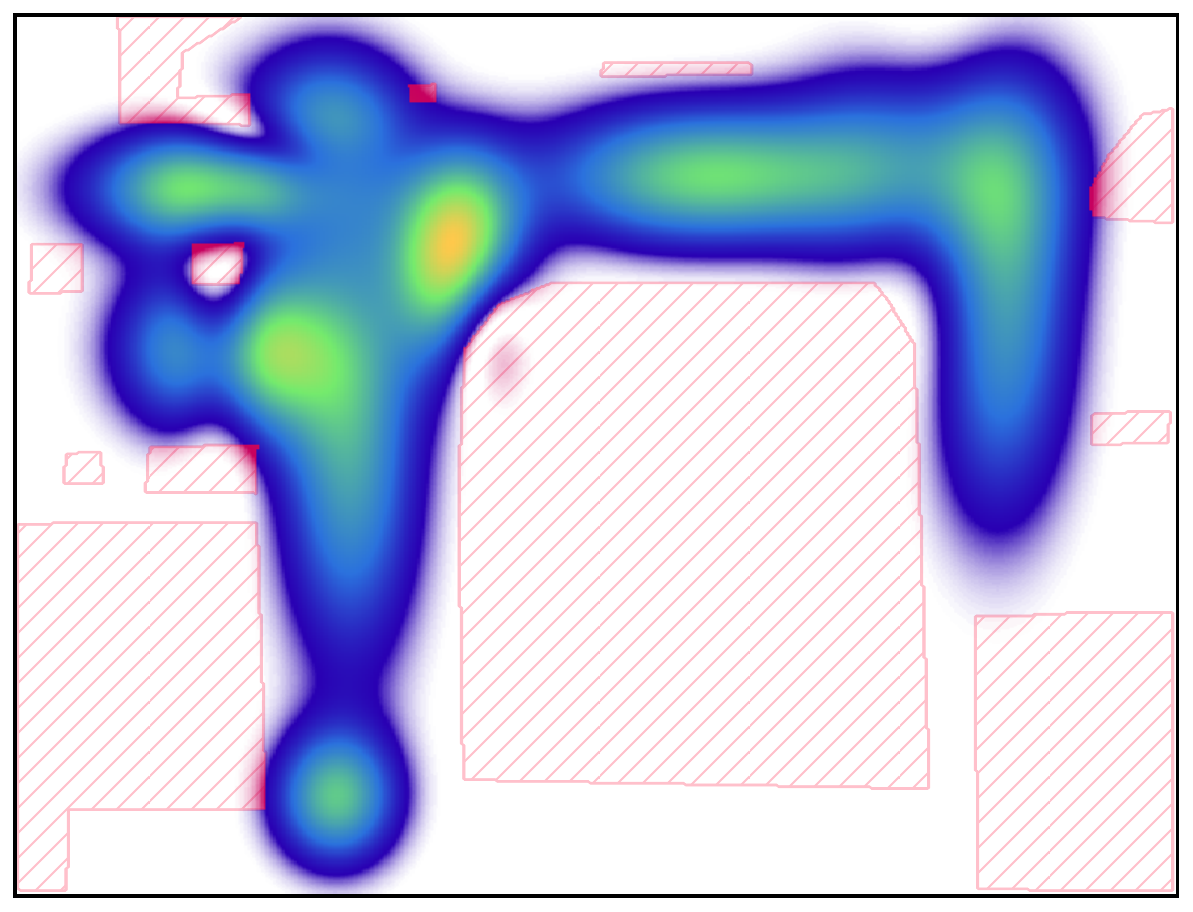}
\end{tabular}
}
\end{center}
\label{fig:pal_1_app}
\caption{Additional results for the SDD targets. GMMs are shown in the top row, SMMs in the bottom row. The ELBO above each image was estimated from $10^5$ samples. We report the mean and standard deviation over $10$ repeated estimations. \emph{Higher is better.}}
\end{figure}

Since the target does not have informative gradients w.r.t. the inputs at constrained areas, we use the RLOO gradient estimator for \emph{both} SMMs and GMMs as opposed to reparameterization. 
We learn the SMMs via rejection sampling as it has shown to provide a good tradeoff between learning efficiency and approximation quality in our synthetic experiments (\cref{tab:high_dim}).
The models were trained and selected based on $10^4$ samples. The ELBO values in \cref{fig:pal_1_app} were estimated from $10^5$ samples; estimation was repeated $10$ times and we report the resulting mean and standard deviation.

\end{document}